\documentclass{article}

\PassOptionsToPackage{numbers, compress}{natbib}
\usepackage{arxiv_style}

\usepackage{microtype}
\usepackage{graphicx}
\usepackage{subfigure}
\usepackage{booktabs} 

\usepackage{placeins} 
\usepackage{siunitx} 
\usepackage{listings}
\usepackage{bm}
\usepackage{rotating}

\usepackage{hyperref}

\usepackage{amsmath}
\usepackage{amssymb}
\usepackage{mathtools}
\usepackage{amsthm}
\usepackage{enumitem}
\usepackage{ulem}

\usepackage[capitalize,noabbrev]{cleveref}

\theoremstyle{plain}

\theoremstyle{definition}

\theoremstyle{remark}

\usepackage[textsize=tiny]{todonotes}
\usepackage{textcomp}
\usepackage[table]{xcolor}
\usepackage{multirow}
\usepackage{scalerel}
\usepackage{makecell}

\usepackage{pifont}
\usepackage{wrapfig}
\usepackage{titletoc}
\usepackage{IEEEtrantools}

\newcommand{\removetodo}[1]{}  

\newlength{\myvspace}
\author{%
  Matthias Blaschke\thanks{Equal contribution.} \\
  University of Augsburg \\ 
  CAAPS\thanks{Centre for Advanced Analytics and Predictive Sciences} \\
  Germany \\
  \And
  Daniel Kienzle\footnotemark[1] \\
  University of Augsburg \\ 
  CAAPS\footnotemark[2] \\
  Germany \\
  \And
  Zsuzsanna Koczor-Benda \\
  University of Warwick \\ 
  United Kingdom \\
  \And
  Julian Lorenz \\
  University of Augsburg \\ 
  CAAPS\footnotemark[2] \\ 
  Germany \\
  \And
  Rainer Lienhart \\
  University of Augsburg \\ 
  CAAPS\footnotemark[2] \\ 
  Germany \\
  \And
  Fabian Pauly \\
  University of Augsburg \\ 
  CAAPS\footnotemark[2] \\ 
  Germany \\
}

\title{Beyond Drug Discovery: The Nanotechnology Molecular Optimization (NMO) Benchmark} 

\begin{document}
\bstctlcite{bstctl:nodash}

\maketitle

\begin{abstract}
Generative molecular design is shaped by simple proxy benchmarks for drug-like properties and models pretrained on large pharmaceutical datasets.
This combination yields strong benchmark metrics but limits transferability to domains structurally distinct from drug discovery.
To overcome this limitation and drive discovery toward real, scientifically grounded targets, we introduce the Nanotechnology Molecular Optimization (NMO) Benchmark, which bridges machine learning (ML) and quantum materials science.
NMO acts simultaneously as a rigorous testbed for the ML community and a discovery engine for nanotechnology research.
The suite replaces proxy oracles with quantum simulations and introduces strict protocols that prioritize scientific utility over leaderboard-oriented overfitting.
The physics-based NMO tasks impose hard structural constraints and rugged fitness landscapes, posing fundamentally new requirements on generative models.
Notably, advanced molecular optimization methods underperform much simpler approaches on the NMO tasks.
We develop a new baseline method identifying the critical components to solve the NMO tasks, including a novel representation for modeling structural constraints and a domain-agnostic pretraining strategy to eliminate pharmaceutical dataset bias. 
Our results surpass state-of-the-art physical properties and reveal previously unknown structural motifs, offering new insights for the nanotechnology community and demonstrating that ML can drive genuine scientific discovery.
\end{abstract}    
\section{Introduction}
\label{sec:intro}
Generative molecular design has achieved strong benchmark performance, yet this progress is increasingly concentrated in pharmaceutical applications and only evaluated against proxy oracles that do not reflect the complexity of real-world objectives.
Many seemingly ``generalist'' models are implicitly specialized \cite{nigam2023tartarus}.
Recent methods like GenMol \citep{lee2025genmol}, InVirtuoGen \citep{kaech2025invirtuogen}, and f-RAG \citep{lee2024molecule} excel at finding optimized candidates for the proxy tasks they are optimized for, but are not designed to discover structures in entirely new spaces.
Simple benchmarks like PMO \cite{gao2022pmo} become increasingly saturated, encouraging models to overfit instead of learning generalizable principles \cite{hutchinson2022evaluation_gaps}. \\[\myvspace]
At the same time, materials science is shifting from the laborious manual discovery of useful molecules to the automated inverse design of materials and molecules for specific applications.
This is especially interesting in the field of nanotechnology, where functional devices are built from the bottom up by tailoring the properties of individual molecules.
Prominent examples are Self-Assembled Monolayers (SAMs) \citep{love2005self} and Metal-Organic Frameworks (MOFs) \citep{li1999design}, which have recently been recognized with the 2025 Nobel Prize in Chemistry.
This demonstrates the immense potential of designing molecular structures for nanotechnology applications. 
However, the high degree of domain-specific knowledge required in both generative modeling and quantum simulations currently creates a significant barrier between the machine learning (ML) and materials science communities. \\[\myvspace]
To bridge this gap, we introduce the \textbf{Nanotechnology Molecular Optimization (NMO)} Benchmark Suite for molecular design targeting challenging quantum physical tasks with real impact.
Importantly, our benchmark is easily accessible to ML researchers without a background in physics, allowing them to tackle real-world problems in materials science and nanotechnology.
At the same time, we provide the infrastructure that allows the nanotechnology community to access the proposed candidates, thereby connecting both domains.
Our suite currently covers 3 distinct fields of nanotechnology, each represented by its own thriving scientific community.
These are chosen because they currently transition from fundamental research to applied device engineering.
Recent milestones confirm that experimental methodologies have matured to provide the resolution necessary for validating and measuring molecular properties directly \cite{yelishala2025phonon, luan2026tuning, gemma2023full, xomalis2021detecting}.
Consequently, the primary bottleneck has shifted from measurement to molecular design: the challenge of identifying optimal candidates within an astronomical search space.
In contrast to previous benchmarks, our tasks require modeling the binding of molecules to gold surfaces, which completely changes their properties compared to free molecules and introduces hard constraints on the molecular structure. \vspace{\myvspace}
\begin{figure}
    \centering
    \includegraphics[width=0.65\linewidth]{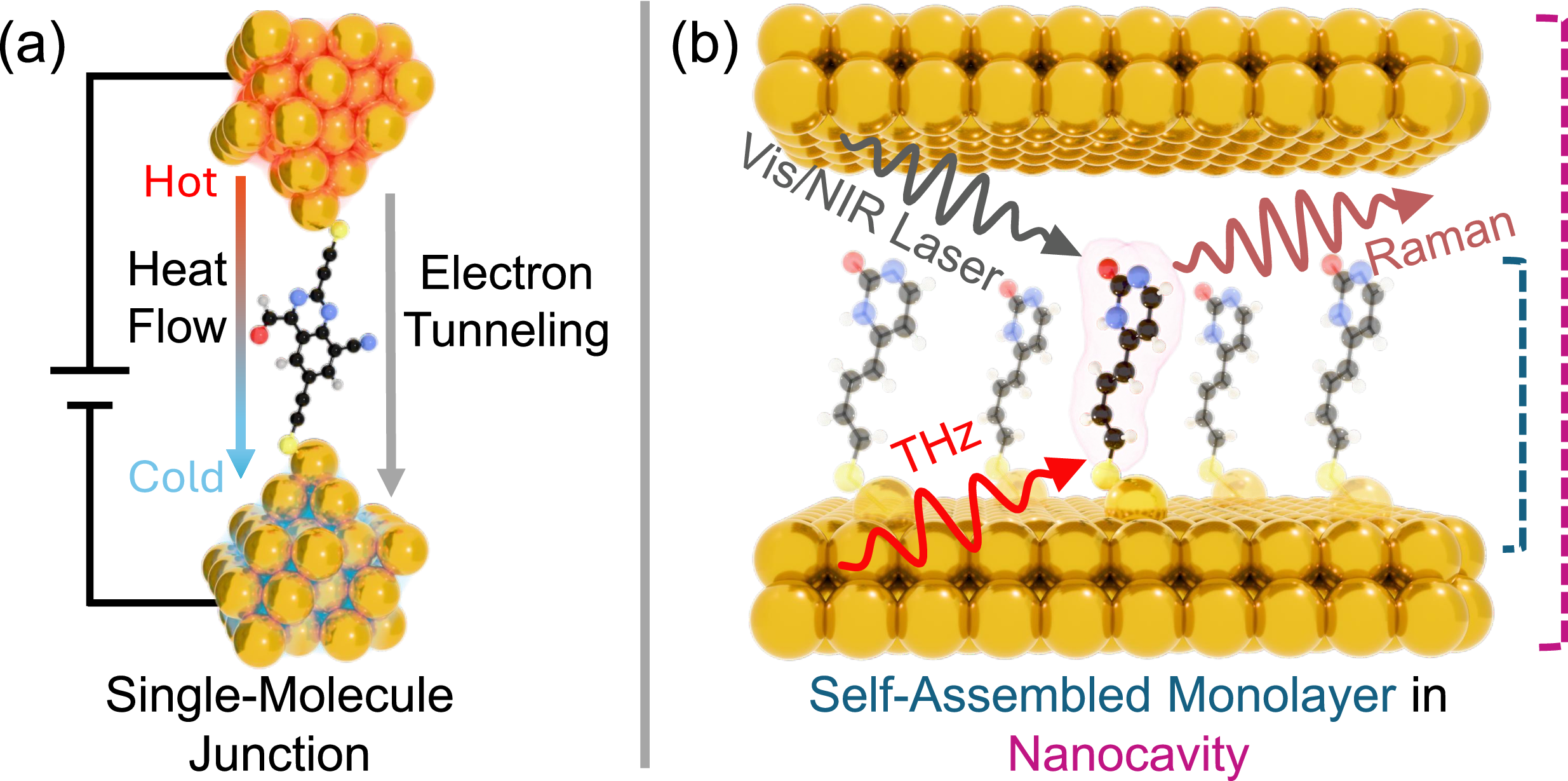}
    \caption{Molecular systems: 
    (a) A molecule contacted by gold surfaces on both sides forms a single-molecule junction for tuning thermal (Phonon Task) or thermoelectric transport (Thermoelectric Task). 
    (b) Molecules anchored on bottom gold surface (SAM) forming a nanocavity for THz detection via Raman scattering (Molecular Optomechanics Task).
    }
    \label{fig:physics_motivation}
\end{figure}
\noindent First, in the \textbf{Phonon Task}, we design single-molecule junctions (MJs, see \cref{fig:physics_motivation}(a)) to precisely control heat flow at the atomic scale. 
This approach enables the inverse design of robust thermal insulators, helping to engineer the next generation of nanoscale devices. \\
Second, in the \textbf{Thermoelectric Task}, we optimize the efficiency of converting heat into electricity in MJs, which can be used for next-generation cooling systems or sustainable waste heat harvesting. \\
Third, in the \textbf{Molecular Optomechanics Task}, we tailor molecules for a SAM (see \cref{fig:physics_motivation}(b)), facilitating room-temperature detection of terahertz (THz) radiation for medical and security applications. \\[\myvspace]
The combination of an extremely rugged fitness landscape created by quantum simulations, hard constraints due to the molecule binding and a strict benchmark protocol designed to prevent overfitting, creates a non-trivial ML challenge.
We show that advanced molecular optimization methods underperform simple genetic algorithms \cite{tripp2023molga, jensen2019graph} on the NMO tasks.
Yet even the strongest methods fail to identify high-performing candidates for all three tasks.
We identify crucial challenges for solving the NMO tasks, present possible solutions to these challenges, and provide a baseline method based on the genetic GFN framework \cite{kim2024geneticgflow} incorporating these solutions.
Our method finds state-of-the-art candidates that surpass current literature performance in all three tasks, demonstrating that our benchmark can drive real scientific discovery in nanotechnology.
Our contributions are as follows:
\begin{itemize}[leftmargin=*,itemsep=0pt,parsep=0.5\myvspace,topsep=0pt,partopsep=0pt]
    \item \textbf{NMO Benchmark Suite:} We introduce the Nanotechnology Molecular Optimization (NMO) Benchmark to evaluate molecular design for complex quantum physics applications (phonon transport, thermoelectric efficiency, and molecular optomechanics).
    NMO enforces strict evaluation protocols to prevent task-specific overfitting and incorporates hard structural constraints typical of real-world nanotechnology.
    The suite is fully accessible to ML researchers without a physics background, providing a reproducible interface between the ML and nanotechnology communities.

    \item \textbf{Solution Properties:} We identify two critical failure modes of existing methods on NMO: inability to natively represent molecule-electrode binding, and distributional mismatch from pharmaceutical pretraining datasets.
    To address these, we introduce \textsl{Graph Group SELFIES (GGS)}, a fragment-based molecular representation that encodes electrode binding natively, guarantees chemical validity by construction and ensures synthesizability through a curated fragment inventory.
    GGS enables the creation of \textsl{unbiased synthetic pretraining} datasets, allowing models to learn fundamental chemical assembly principles without historical pharmaceutical data biases.
    This is particularly important because, for the NMO tasks, there are no datasets with known molecular properties available for pretraining.
    The fragment library can be curated as a "Chemist's Shop" of building blocks that are readily available in a real lab setting.

    \item \textbf{Baseline Method:} We provide a strong baseline method extending the genetic GFN framework \cite{kim2024geneticgflow} with our proposed solution properties and illustrating their importance.
    Our method discovers molecules surpassing physical properties reported in domain-specific literature across all three NMO tasks and reveals previously unknown structural motifs. 
    This exemplarily demonstrates the potential of our benchmark to offer scientific insights for the nanotechnology community.
\end{itemize}
\section{Related Work}

\textbf{Methods in Generative Molecular Design}:\,
Molecular optimization is currently dominated by pharmaceutical benchmarks like the PMO suite \cite{gao2022pmo}. 
These suites evaluate the capacity of models to optimize drug-like properties using simple proxy oracles (e.g. similarity to arbitrary substances). \\
Current popular approaches to solve these benchmarks include \textsl{data-driven models} that remember, interpolate, and combine valid substructures from massive datasets (e.g., SAFE-GPT \cite{noutahi2023safe}, GenMol \cite{lee2025genmol}, f-RAG \cite{lee2024molecule}, and InVirtuoGen \cite{kaech2025invirtuogen}).
These models are heavily optimized for pharmaceutical proxy tasks, exhibiting strong distributional bias toward drug-like chemical space.
f-RAG and GenMol even rely on task-specific vocabularies and adapted hyperparameters across all 23 PMO tasks, heavily limiting the generalizability of the reported performance. \\ 
In contrast, \textsl{search-based methods} navigate chemical space, often utilizing algorithms like Genetic Algorithms (molGA \cite{tripp2023molga}), Reinforcement Learning (REINVENT \cite{olivecrona2017reinvent}), and Generative Flow Networks (GFNs) \cite{kim2024geneticgflow,bengio2021gflow}.
While they are also often initialized with dataset priors (e.g. \cite{jain2022biological}), they are not overly reliant on them by design and instead focus on optimizing target properties from scratch. \\
Both paradigms are mostly evaluated on proxy oracles whose narrow functional form is readily exploited \cite{hutchinson2022evaluation_gaps,nigam2023tartarus}.
The success of a model on these benchmarks is often measured by simple metrics too coarse to separate true optimization capacity from benchmark-specific adaptation and memorization.
For genuine scientific utility, benchmarks must transition toward high-fidelity simulation based scoring, that cannot be exploited that easily, combined with strict benchmark protocols.
To the best of our knowledge, TARTARUS \cite{nigam2023tartarus} represents the only approach, replacing proxy oracles with computational chemistry workflows, yet it uses task-specific datasets and starting points.
The NMO benchmark pairs high-fidelity quantum-mechanical simulations with a strict evaluation protocol anchored in real material science objectives, making it completely dataset-independent.
With this, we enable the ML community to tackle real scientific physics problems.\\[\myvspace]
\textbf{Molecular Optimization in the Physical Sciences}:\,
Generative molecular design approaches in the physical sciences are often designed for isolated specialized applications.
For example, ML has been applied to discover stable inorganic crystals \cite{merchant2023scaling}, generate quantum material lattices \cite{okabe2025scigen}, and inversely design Metal-Organic Frameworks \cite{cleeton2025inverse}.
Most relevant to our NMO tasks, existing work has used Genetic Algorithms to optimize MJs for phonon transport and mechanosensitivity \cite{blaschke2025revealing,blaschke2023designing}.
Similarly, optomechanical THz upconversion performance has been predicted by screening databases \cite{koczor2021molecular} or by biasing 3D generative models like G-SchNet \cite{koczor2025generative, gebauer2019GSchNet}.
However, these approaches remain within their respective scientific communities, underlining the need for a standardized benchmark. \\[\myvspace]
\textbf{Methodological Bottlenecks}:\,
Successfully solving physical problems requires \texttt{representations} and \texttt{vocabularies} that escape pharmaceutical dataset biases and guarantee structural validity.
Standard string representations like SMILES \cite{weininger1988smiles} allow syntactically correct strings that violate chemical rules.
Extensions like SAFE \cite{noutahi2023safe} utilize fragments but still fail to guarantee valid combinations, while SELFIES \cite{krenn2020selfies} and Group SELFIES \cite{cheng2023groupselfies} rely on post-hoc parsers that cause arbitrary structural truncations \cite{reboul2025improving}. 
Recent models (e.g. GenMol \cite{lee2025genmol}) mine their vocabularies from massive pharmaceutical datasets using heuristics like BRICS \cite{degen2008brics}.
This introduces an implicit drug-like dataset bias and does not guarantee that the vocabulary is accessible in a real lab setting. 
Our GGS representation addresses these failure modes by construction. It guarantees validity without truncation, builds from synthesizable fragments, and supports unbiased pretraining. Additionally, it naturally encodes the molecule-electrode binding sites required by the NMO tasks.
\section{The NMO Benchmark Suite}
\label{sec:nmo_suite}
The NMO benchmark suite provides a standardized evaluation environment for generative molecular design in quantum materials science.
The suite evaluates models across three distinct nanotechnology applications: Phonon Transport (PH), Thermoelectrics (TE), and Molecular Optomechanics (MO).
It is specifically designed to be accessible to the ML community without requiring prior domain expertise in quantum physics.
To achieve this, the NMO suite is implemented to operate fully automatically for these tasks.
A generative agent simply outputs a one-dimensional molecular string representation (such as SMILES or GGS).
The suite's backend automatically parses the string, constructs the functional nanostructure, performs 3D geometry optimization, and executes the complex quantum simulations.
The program then returns a scalar fitness score back to the agent and writes extensive metadata.
Usually, such simulations are computationally infeasible for large-scale optimization, but recent advances in semi-empirical methods have made it possible to evaluate these properties with reasonable accuracy and speed.
Therefore, our scoring utilizes the semi-empirical \texttt{xtb} package \cite{bannwarth2021xtb}, and our implementation for calculating the physical properties is strictly validated against literature in \cref{sec:nmo_details}. 
A core challenge of NMO is that anchor positions must be part of the optimization. 
The benchmark supports this through two interfaces: GGS (native) or SMILES with atom indices. 
Methods are free to use either or propose new representations.
We also provide the simple PMO proxy tasks through the same interface, allowing for quick prototyping and debugging before moving to the more expensive NMO tasks.
Furthermore, we provide the infrastructure for ML researchers to easily share their proposed candidates with the nanotechnology community.
Simultaneously, its flexible implementation allows nanotechnology experts to design and integrate new tasks in the future based on their specific research interests by leveraging a broad set of available properties (e.g. relaxed geometry, electronic structure, vibrational properties).

\subsection{Design Tasks and Reward Formulation}

The central objective of each task in the NMO benchmark is to find molecules that optimize the physical properties relevant for the task under a strictly limited budget of \num{10000} oracle evaluations.
This constraint reflects the real-world cost of quantum simulations and experimental validations, therefore, ensuring that success in NMO can translate to practical scientific discovery. \\[\myvspace]
For each task, we define a fitness function that combines the physical property of interest $\mathcal{O}_{\text{task}}(x)$ with penalties $\mathcal{P}_i(x)$, and hard physical constraints $\Theta(x)$:
\begin{equation}
    f_{\text{task}}(x) = c_{\text{task}} \cdot \mathcal{O}_{\text{task}}(x) \cdot \mathcal{P}_\mathrm{SA, task}(x) \cdot \mathcal{P}_\mathrm{rot}(x) \cdot \Theta(x) \, .
    \label{eq:fitness_struc}
\end{equation}
Here, $c_{\text{task}}$ is a heuristic scaling constant ensuring that fitness values are on a comparable scale across tasks.
The penalty $\mathcal{P}_{\mathrm{SA}}(x)$ is based on the synthetic accessibility (SA) score \cite{ertl2009estimation}. 
Due to the different scaling of the physical properties $\mathcal{O}_{\text{task}}(x)$, we use task-specific functional forms.
$\mathcal{P}_{\mathrm{rot}}(x)$ is a sigmoid-shaped penalty for the number of rotatable bonds ($N_\mathrm{rot}$), which prevents massive conformer ensembles that would be difficult to control experimentally.
For all tasks, we use $\mathcal{P}_{\mathrm{rot}}(x) = \frac{1}{1+e^{N_\mathrm{rot} - 3.5}}$.
Finally, $\Theta(x) \in \{0, 1\}$ represent hard physical constraints, immediately setting the fitness to zero if a molecule fails to meet essential criteria.
This happens if a molecule cannot form a valid junction geometry, fails our custom filter set (see \cref{sec:filters}), or if the HOMO-LUMO gap, an indicator of chemical stability, is below a certain threshold.
These extensive safeguards go far beyond what is typically done in generative molecular design, establishing a realistic framework for practical scientific discovery.
In the following, we describe each task in more detail. \\[\myvspace]
\textbf{Phonon Task (PH)}: \,
At the nanoscale, heat transport is governed by quantized lattice vibrations called phonons.
For the \texttt{PH} task, we optimize for molecules that exhibit strongly suppressed thermal conductance $\kappa_\mathrm{ph}$ when contacted by macroscopic electrodes (see \cref{fig:physics_motivation}(a)).
Recent measurements can resolve transport changes down to the single-atom level \cite{luan2026tuning,yelishala2025phonon}.
Despite these experimental capabilities, the search for effective molecular-scale thermal insulators remains an unsolved problem with applications in nanoelectronic devices \cite{cahill2014nanoscale} and quantum technology for effective heat management. \\[0.5\myvspace]
We define $\mathcal{O}_\text{PH} = 1/\kappa_\mathrm{ph}$ reciprocal to the thermal conductance to optimize for low thermal conductance and penalize the synthetic accessibility on the same scale $\mathcal{P}_\mathrm{SA,PH}(x) = \frac{1}{\mathrm{SA}}$. The scaling constant is set to $c_\mathrm{PH} = 0.5$. Details are given in \cref{sec:phonon_details}.  \\[\myvspace]
\textbf{Thermoelectric Task (TE)}: \,
Beyond the transport of heat, MJs also serve as conductive pathways for electrons (see  \cref{fig:physics_motivation}(a)), allowing them to tunnel through the molecule when a voltage bias is applied \cite{ratner2013brief}.
Driven by the thermoelectric effect \cite{reddy2007thermoelectricity}, MJs can convert waste heat back into usable electrical power, which is a critical for sustainable energy technologies.
The efficiency is given by the \textit{figure of merit} $Z=GS^2/\kappa$ (often reported as $ZT$ with $T$ being the temperature), which requires a multi-objective optimization: maximizing the electrical conductance ($G$) and the Seebeck coefficient ($S$) while simultaneously minimizing the total thermal conductance ($\kappa=\kappa_\mathrm{ph} + \kappa_\mathrm{el}$). \\[0.5\myvspace]
We set $\mathcal{O}_\text{TE} = Z$ with $c_{\mathrm{TE}} = 0.01$, and a linear penalty $\mathcal{P}_\mathrm{SA,TE}(x) = \frac{10-\text{SA}}{9}$.
Computational details are provided in \cref{sec:te_details}. \\[\myvspace]
\textbf{Molecular Optomechanics Task (MO)}: \,
Detectors for THz and mid-infrared (MIR) radiation have impactful applications for non-destructive imaging in medicine, security, and quality control \cite{pawar2013terahertz}.  
We consider nanoplasmonic THz detectors \cite{roelli2020molecular,xomalis2021detecting,chen2021continuous}, which work by upconverting THz radiation to visible frequencies utilizing specific vibrational modes of molecules.
In the \texttt{MO} task, we construct such a detector by integrating a molecular SAM into a dual nanoantenna that enhances both incoming and outgoing light, enabling ultrasensitive detection of THz radiation.
This is displayed in \cref{fig:physics_motivation}(b). \\[0.5\myvspace]
Following \citet{koczor2021molecular}, the suite evaluates the intrinsic light upconversion capability ($P$) while considering geometrical influences ($F$), setting $\mathcal{O}_\text{MO} = P+F$.
Moreover, we use $c_\mathrm{MO} = 1/4$ and $\mathcal{P}_\mathrm{SA,MO}(x) = \left(\frac{10-\text{SA}}{9}\right)^2$.
Details are given in \cref{sec:thz_details}.

\subsection{Performance Metrics}
\label{sec:metrics}
The goal of NMO is to maximize fitness under a strict budget of $N=10000$ fitness evaluations similar to \cite{gao2022pmo}.
This is a realistic setting where each oracle call represents a costly quantum simulation, therefore, emphasizing the importance of sample efficiency and practical utility.
We evaluate success with the following metrics:
\begin{itemize}[leftmargin=*,itemsep=0pt,parsep=0pt,topsep=0pt,partopsep=0pt]
    \item \textbf{Top-10 AUC:} The Area Under the Curve of the fitness scores of the top-10 molecules found up to iteration $i$, assessing both sample efficiency and candidate quality \cite{gao2022pmo}.
    \item \textbf{Mean Top-10 Fitness:} The average fitness of the ten best molecules after the optimization process, providing a direct measure of the quality of the best candidates discovered.
    \item \textbf{Mean Top-10 SA:} The average SA score \cite{ertl2009estimation} of the top-10 candidates is a popular measure to estimate if a molecule is synthesizable.
    Following \citet{vorsilak2020syba}, we consider scores above $4.5$ as unlikely to be synthesizable. 
    \item \textbf{Relevance Indicator (RI):} \, A binary indicator of whether a candidate's physical properties surpass the best molecules reported in the current scientific literature for the respective task while maintaining an SA score below 4.5.
    The selection of these thresholds is summarized in \cref{sec:thresholds}.
    We note that these are free of any heuristic parameters. 
    If a method is able to find such relevant molecules for all three tasks, it demonstrates its potential for genuine scientific impact beyond just optimizing a mathematical proxy.
\end{itemize}

\subsection{Handling Structural Constraints: The Graph Group SELFIES representation}
\label{sec:ggs}

\begin{figure} 
    \centering
    \includegraphics[width=0.65\linewidth]{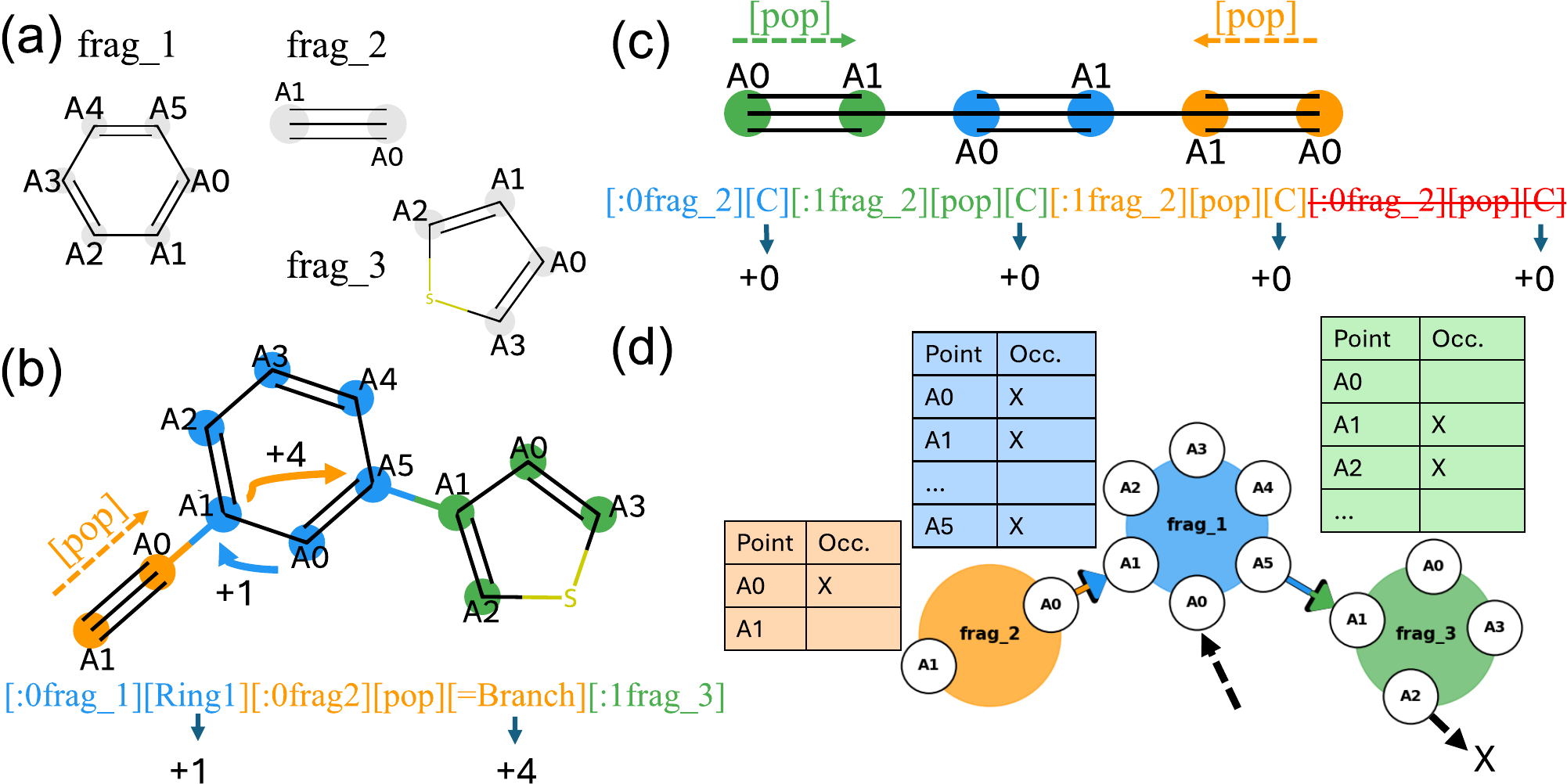}
    \caption{(a) Fragment library. (b) Molecule with GS string representation (see \cref{sec:GS_example_explanation}). (c) Syntactically correct GS string has to be truncated (red) during decoding due to insufficient coupling points on the central fragment (blue). (d) GGS representation of molecule from (b) as a DAG with tracked valence (colored tables), directed edges (arrows), and termination marker (\texttt{X}.)}
    \label{fig:encoding_overview}
\end{figure}
Solving the NMO benchmark requires modeling the anchor positions where the molecule connects to the electrodes. 
NMO accepts standard SMILES with explicit atomic indices, allowing users to freely utilize existing generative models. 
However, standard string representations lack native topological support for these anchors. 
To provide the community with a robust, out-of-the-box infrastructure, we introduce Graph Group SELFIES (GGS) as optional interface for the NMO suite.
GGS is based on Group SELFIES (GS) \cite{cheng2023groupselfies}, that provides a robust foundation by encoding molecules as sequences of predefined fragments.
Each token \texttt{frag\textsubscript{n}} in the string represents a molecular fragment from a predefined vocabulary (see \cref{fig:encoding_overview}(a)). Its connectivity to the previous and next groups is defined via specific attachment points, denoted $S_\mathrm{in}$ and $S_\mathrm{out}$.
A molecule is thus constructed by chaining these tokens, e.g. \fbox{$[\cdots]\left[:S_\mathrm{in}\langle\mathrm{frag}_n\rangle\right]\left[S_\mathrm{out}\right][\cdots]$}.
Branches can be created using a special \texttt{[pop]} token, which ends the current branch and returns to the previous fragment.
An example is shown in \cref{fig:encoding_overview}(b).
While GS guarantees valid molecules via a post-hoc parser, this approach often comes at the cost of structural truncation.
If a generated token requires an attachment point that is not available (e.g., the light blue fragment in \cref{fig:encoding_overview}(c)), the parser cuts the chain, resulting in a molecule different from the intended sequence. 
This creates encoding components that remain functionally decoupled from the resulting molecule and reward aliasing, which can hinder learning. \\[\myvspace]
GGS overcomes these shortcomings, guarantees validity by construction, and adds native anchor modeling.
This is done by handling molecules as a directed acyclic graph (DAG) while using the GS notation as its textual string representation (see \cref{fig:encoding_overview}(d)).
This string-based notation ensures that the new encoding remains accessible to established molecular design methods that frequently use string representations.
During construction, we explicitly track the valence and available coupling sites of every node in the graph.
Allowing only chemically possible connections ensures that every generated graph describes a valid molecule by construction, eliminating the ambiguity of post-hoc sanitization.
Crucially, the graph structure defines source and sink nodes which \textbf{can model electrode connections natively.}
In addition, the graph enables robust genetic operations such as crossover and mutation, guaranteeing offspring are valid molecules by satisfying valence constraints (see \cref{sec:genetic_operations_search}).
In contrast, string-based genetic operators often break syntax and validity.  \\[\myvspace]
The valid-by-construction nature of GGS allows us to create a \textbf{synthetic dataset} of random graphs.
This is critical in regimes where domain-specific data is unavailable, as is the case for the NMO benchmark.
Our synthetic dataset enables a new domain-agnostic pretraining route in which generative models learn fundamental chemical assembly principles directly, without inheriting the uncontrollable bias of historical datasets.
Details are given in \cref{sec:synthetic_datasets}.
When paired with an appropriate model, this facilitates synthetic-to-real generalization that extends beyond our specific NMO application. \\[\myvspace]
In summary, GGS offers three properties relevant to NMO: validity by construction, a synthetic pretraining route that avoids pharmaceutical bias, and native representation of electrode binding.
Further details and limitations are given in \cref{sec:ggs_details}.
Making anchors part of the optimization is a more general principle which our experiments identify as crucial for finding high-performing candidates.
GGS is one way to realize this, but alternative approaches (e.g. anchor positions by an auxiliary task) might be equally compatible.

\subsection{Benchmark Protocol}
\label{sec:benchmark_protocol}
In response to frequent exploitations of previous benchmarks (e.g., task-specific vocabularies or oracle calls before optimization for an informed starting point), we impose a strict evaluation protocol.
Any use of task-specific datasets or task-specific hyperparameter tuning is explicitly forbidden, thus, models must use a single configuration across all three physically distinct tasks.
We provide a fragment library, which is defined by domain experts, as part of the benchmark infrastructure.
While methods are free to choose their own building blocks, they have to use the same fragment library for all three tasks, preventing task-specific over-optimization through the choice of fragments.
Furthermore, we provide a SMILES-translated version of our synthetic pretraining dataset, ensuring that the fundamental domain knowledge captured in our fragment library remains fully accessible to standard literature methods.
For the PH task, methods are permitted to bound the length of generated molecules to avoid a known degenerate regime of the oracle (see \cref{sec:limitations}).
We also provide a set of cheap heuristic filters (see \cref{sec:filters}) that may be applied before the fitness call to remove duplicates and unstable candidates.
Candidates removed by these filters do not count toward the evaluation budget.
Evaluations are strictly limited to a fixed budget of \num{10000} fitness evaluations per seed, and all reported results must average across five consecutive seeds that are not cherry-picked.
Importantly, the use of expensive fitness evaluations before the main optimization loop is prohibited. \\[\myvspace]
Taken together, these constraints turn NMO into a test of generalist optimization capability.
A single method, with a single configuration, must navigate three physically distinct optimization problems without any task-specific assistance.
Success under this protocol is evidence that a method has learned to optimize molecules in a physical setting, not that it has been tuned to a specific one.
While the computational cost of the NMO tasks precludes exhaustive hyperparameter tuning anyway, a genuinely robust method should remain effective without relying on perfectly tuned hyperparameters, since such methods are far more likely to be useful in scientific domains.

\section{Methodology}
\label{sec:methodology}

\textbf{Literature Methods}.\,
We evaluate five popular methods to cover the dominant paradigms in molecular optimization. 
REINVENT~\cite{olivecrona2017reinvent} pre-trains a sequence model on molecules and fine-tunes it via reinforcement learning using oracle scores as rewards. 
MolGA~\cite{tripp2023molga} is a training-free genetic algorithm that generates offspring via crossover and mutation. 
Both are simple established approaches built around the SMILES encoding but can be adapted to novel molecular encodings. 
GenMol~\cite{lee2025genmol} is a masked diffusion model generating fragment-based SAFE sequences non-autoregressively, and f-RAG~\cite{lee2024molecule} augments a SAFE-based language model with a dynamic fragment vocabulary, iteratively retrieving and assembling high-scoring building blocks.
Both are selected for their state-of-the-art performance on the PMO benchmark, but are heavily built around SMILES encoding and depend on massive pretraining, making an adaptation to new encodings difficult. 
Genetic-GFN~\cite{kim2024geneticgflow} combines a GFlowNet policy with genetic operators, learning to sample proportionally to the oracle reward.
The method demonstrated strong sample efficiency and diversity in molecular optimization, making it well-suited to our NMO setting where oracle calls are expensive.
The simple architecture and low computational cost for the model itself make it straightforward to adapt and extend. \\[\myvspace]
\textbf{Our Baseline Method: Extensions to the Genetic GFN Framework}.\,
We extend the Genetic GFN framework \cite{kim2024geneticgflow} to create a domain-agnostic optimization method capable of navigating the rugged landscapes of physical simulations.
We introduce critical extensions to provide a robust baseline capable of solving the NMO tasks.
Details and specific architectural choices are provided in \cref{sec:gentic_gfn_framework}. 
In the following we provide a high-level overview of the key components: \\[0.5\myvspace]
\emph{GGS Integration and Synthetic Pretraining:} We switch from the standard SMILES to GGS encoding to leverage the valid-by-construction property and advanced graph-level genetic operators.
Pretraining on a synthetic dataset of \num{300,000} random graphs establishes a foundational validity-prior and minimizes dataset bias.
This approach enables synthetic-to-real generalization and extends beyond NMO to other domains where specialized data is unavailable.
We note that these changes are most important for success on the NMO tasks. \\[0.5\myvspace]
\emph{Transformer Architecture:} We replace the recurrent backbone with a transformer, adopting a modern standard for deep learning. \\[0.5\myvspace]
\emph{Adaptive Stability Control:} The stability of GFlowNets in rugged energy landscapes is an active research challenge \cite{kim2023learning,lau2024qgfn,deleu2025relative,fan2025adaptive}.
In our setting, the fitness landscapes are highly rugged and we observe two distinct failure modes: \textit{Catastrophic Forgetting} (unlearning the syntax rules and generating sequences that fail to construct a GGS graph) and \textit{Mode Collapse} (high duplicate rate).
To solve this, we introduce two adaptive mechanisms.
Dynamic Cooldown (DCD) stabilizes gradients by re-exposing the model to valid molecular syntax when the invalid rate rises. 
Dynamic Exploration (DEX) prevents the policy from collapsing into narrow, high-reward modes by flattening the sampling distribution.
Details can be found in \cref{sec:adaptive_stability}.  \\[0.5\myvspace]
\emph{Descriptor Injection:} We enhance the agent by injecting chemical intuition through a multitask auxiliary objective during pretraining and optimization.
The model predicts a set of cheaply computable molecular descriptors directly from its latent representation.
This objective provides the model with a basic understanding of structure-property relationships without requiring any domain-specific pretraining data. 
During the optimization phase, this acts as a regularizer to prevent latent space collapse.
Details are provided in \cref{sec:descriptors_method}.

\section{Experiments}
\label{subsec:ablation}
The performance metrics for all tested methods are summarized in \cref{tab:auc_table}. \\[\myvspace]
\textbf{Literature Methods:}\,
\emph{f-RAG and GenMol} rely on SMILES and massive pharmaceutical pretraining.
With default vocabularies and anchors placed on the first and last non-hydrogen atoms (implicit), both fail across all three oracles, with near-zero performance.
f-RAG and GenMol seem to be too tightly coupled to their pharmaceutical priors to adapt.
\emph{molGA} as training-free genetic algorithm provides the cleanest test of the encoding interface.
With SMILES and implicit anchor modeling, molGA is competitive on MO and achieves non-zero performance on TE and PH, demonstrating that this setting allows optimization capacity.
Switching to GGS and adopting our advanced genetic operators (see \cref{sec:genetic_operations_search}) yields large gains on TE and improves PH (especially bringing down SA), confirming that native anchor specification is decisive for two-sided binding.
On MO, GGS reduces AUC while bringing SA below the synthesizability threshold.
The drop suggests that the more flexible SMILES encoding is beneficial for the MO task.
For \emph{REINVENT}, we ablate the pretraining-data and encoding axes separately.
ZINC + SMILES fails across all tasks.
Replacing ZINC with our synthetic dataset (translated to SMILES) yields only marginal gains for TE and PH, ruling out a pure data-swap explanation.
Switching to GGS then produces substantial improvement on TE and brings SA below threshold across tasks.
Even with GGS, REINVENT does not find relevant PH candidates.
\emph{Genetic-GFN} is the strongest SMILES method on MO, where the bias from pharmaceutical pretraining appears to help, but it is constrained on the TE and PH tasks and produces above-threshold SA scores throughout.
\begin{table}
\centering
\setlength{\extrarowheight}{0pt}
\addtolength{\extrarowheight}{\aboverulesep}
\addtolength{\extrarowheight}{\belowrulesep}
\setlength{\aboverulesep}{0pt}
\setlength{\belowrulesep}{0pt}
\caption{\textbf{Evaluation of Literature Methods and Ablation Study on the NMO benchmark.} SA scores exceeding the threshold of 4.5 are highlighted in red. Zero values are due to rounding.}
\vspace{-0.18cm}
\label{tab:auc_table}
\resizebox{0.99\linewidth}{!}{%
\begin{tabular}{l|cccc|cccc|cccc} 
\toprule
\multirow{2}{*}{\textbf{Variant}}                                                     & \multicolumn{4}{c|}{\textbf{TE Task}}                                                                                                                                                                                                                                                                                                                                                                                                                                        & \multicolumn{4}{c|}{\textbf{PH Task}}                                                                                                                                                                                                                                                                                                                                                                                                                                                & \multicolumn{4}{c}{\textbf{MO Task}}                                                                                                                                                                                                                                                                                                                                                                                                                                   \\ 
\cmidrule(l){2-13}
                                                                                      & \textbf{AUC} $\left.\kern-\nulldelimiterspace\right\uparrow$ & \begin{tabular}[c]{@{}c@{}}\textbf{Mean $f_\mathrm{TE}$}\\\textbf{Top 10} $\left.\kern-\nulldelimiterspace\right\uparrow$\end{tabular} & \begin{tabular}[c]{@{}c@{}}\textbf{Mean SA}\\\textbf{Top 10} $\left.\kern-\nulldelimiterspace\right\downarrow$\end{tabular}  & \textbf{RI}         & \textbf{AUC} $\left.\kern-\nulldelimiterspace\right\uparrow$ & \begin{tabular}[c]{@{}c@{}}\textbf{Mean $f_\mathrm{PH}$}\\\textbf{Top 10} $\left.\kern-\nulldelimiterspace\right\uparrow$\end{tabular} & \begin{tabular}[c]{@{}c@{}}\textbf{Mean SA}\\\textbf{Top 10} $\left.\kern-\nulldelimiterspace\right\downarrow$\end{tabular} &  \textbf{RI}        & \textbf{AUC} $\left.\kern-\nulldelimiterspace\right\uparrow$ & \begin{tabular}[c]{@{}c@{}}\textbf{Mean $f_\mathrm{MO}$}\\\textbf{Top 10} $\left.\kern-\nulldelimiterspace\right\uparrow$\end{tabular} & \begin{tabular}[c]{@{}c@{}}\textbf{Mean SA}\\\textbf{Top 10} $\left.\kern-\nulldelimiterspace\right\downarrow$\end{tabular} &  \textbf{RI}          \\ 
\midrule
f-RAG                                                                                 & \multicolumn{1}{l}{$0.00\pm0.00$}                            & \multicolumn{1}{l}{$0.00\pm0.00$}                                                                                                     & \multicolumn{1}{l}{$4.08\pm0.83$}                                                                                                                                                                                                & \multicolumn{1}{l|}{\ding{55}(0/5)}  & \multicolumn{1}{l}{$0.02\pm0.01$}                            & \multicolumn{1}{l}{$0.04\pm0.02$}                                                                                                      & \multicolumn{1}{l}{$\textcolor{red}{4.58}\pm0.89$}                                                                                                                                                                                                & \multicolumn{1}{l|}{\ding{55}(0/5)} & \multicolumn{1}{l}{$0.10\pm0.06$}                            & \multicolumn{1}{l}{$0.22\pm0.18$}                                                                                                       & \multicolumn{1}{l}{$\textcolor{red}{4.81}\pm0.90$}                                                                                                                                                                                                        & \multicolumn{1}{l}{\ding{55}(0/5)}    \\
GENMOL                                                                                & \multicolumn{1}{l}{$0.00\pm0.00$}                            & \multicolumn{1}{l}{$0.00\pm0.00$}                                                                                                     & \multicolumn{1}{l}{$4.23\pm0.05$}                                                                                                                                                                                                & \multicolumn{1}{l|}{\ding{55}(0/5)}  & \multicolumn{1}{l}{$0.02\pm0.00$}                            & \multicolumn{1}{l}{$0.02\pm0.02$}                                                                                                      & \multicolumn{1}{l}{$3.85\pm0.07$}                                                                                                                                                                                                & \multicolumn{1}{l|}{\ding{55}(0/5)} & \multicolumn{1}{l}{$0.00\pm0.00$}                            & \multicolumn{1}{l}{$0.01\pm0.01$}                                                                                                       & \multicolumn{1}{l}{$4.10\pm0.16$}                                                                                                                                                                                                         & \multicolumn{1}{l}{\ding{55}(0/5)}    \\ 
\hline
\begin{tabular}[c]{@{}l@{}}molGA (SMILES)\end{tabular}                             & \multicolumn{1}{l}{$0.13\pm0.10$}                            & \multicolumn{1}{l}{$0.26\pm0.13$}                                                                                                     & \multicolumn{1}{l}{$\textcolor{red}{4.71}\pm0.19$}                                                                                                                                                                                               & \multicolumn{1}{l|}{\ding{51}(1/5)}  & \multicolumn{1}{l}{$0.07\pm0.01$}                            & \multicolumn{1}{l}{$0.10\pm0.02$}                                                                                                      & \multicolumn{1}{l}{$\textcolor{red}{4.54}\pm0.24$}                                                                                                                                                                                               & \multicolumn{1}{l|}{\ding{55}(0/5)} & \multicolumn{1}{l}{$0.57\pm0.14$}                             & \multicolumn{1}{l}{$1.28\pm0.57$}                                                                                                       & \multicolumn{1}{l}{$\textcolor{red}{4.58}\pm0.34$}                                                                                                                                                                                                        & \multicolumn{1}{l}{\ding{51}(5/5)}  \\
\begin{tabular}[c]{@{}l@{}} + Switch to GGS  \\ \phantom{+ } \& our GA \end{tabular}                                                                     & \multicolumn{1}{l}{$0.68\pm0.05$}                            & \multicolumn{1}{l}{$1.11\pm0.26$}                               & \multicolumn{1}{l}{$3.76\pm0.39$}                                                                                                                                                                                               & \multicolumn{1}{l|}{\ding{51}(5/5)} & \multicolumn{1}{l}{0.35$\pm$0.13}                                 & \multicolumn{1}{l}{0.52$\pm$0.13}                                                                                                           & \multicolumn{1}{l}{3.52$\pm$0.15}                                                                                                                                                                                                     & \multicolumn{1}{l|}{\ding{55}(0/5)} & \multicolumn{1}{l}{$0.36\pm0.04$}                            & \multicolumn{1}{l}{$0.43\pm0.04$}                                                                                                       & \multicolumn{1}{l}{$4.49\pm0.19$}                                                                                                                                                                                                         & \multicolumn{1}{l}{\ding{51}(5/5)}    \\ 
\hline
\begin{tabular}[c]{@{}l@{}}REINVENT \\ (SMILES \& ZINC)\end{tabular}                     & \multicolumn{1}{l}{$0.00\pm0.00$}                            & \multicolumn{1}{l}{$0.01\pm0.02$}                                                                                                     & \multicolumn{1}{l}{$3.81\pm0.30$}                                                                                                                                                                                               & \multicolumn{1}{l|}{\ding{55}(0/5)}   & \multicolumn{1}{l}{$0.01\pm0.00$}                            & \multicolumn{1}{l}{$0.02\pm0.00$}                                                                                                      & \multicolumn{1}{l}{$3.72\pm0.22$}                                                                                                                                                                                               & \multicolumn{1}{l|}{\ding{55}(0/5)} & \multicolumn{1}{l}{$0.01\pm0.01$}                            & \multicolumn{1}{l}{$0.02\pm0.02$}                                                                                                       & \multicolumn{1}{l}{$3.11\pm0.16$}                                                                                                                                                                                                         & \multicolumn{1}{l}{\ding{51}(1/5)}    \\
+ Synth. Dataset                                                                      & \multicolumn{1}{l}{$0.06\pm0.09$}                            & \multicolumn{1}{l}{$0.11\pm0.13$}                                                                                                     & \multicolumn{1}{l}{$\textcolor{red}{4.56}\pm0.25$}                                                                                                                                                                                               & \multicolumn{1}{l|}{\ding{55}(0/5)}  & \multicolumn{1}{l}{$0.04\pm0.00$}                            & \multicolumn{1}{l}{$0.06\pm0.01$}                                                                                                      & \multicolumn{1}{l}{$4.18\pm0.25$}                                                                                                                                                                                                & \multicolumn{1}{l|}{\ding{55}(0/5)} & \multicolumn{1}{l}{$0.14\pm0.02$}                            & \multicolumn{1}{l}{$0.26\pm0.02$}                                                                                                       & \multicolumn{1}{l}{$\textcolor{red}{4.54}\pm0.18$}                                                                                                                                                                                                         & \multicolumn{1}{l}{\ding{51}(5/5)}    \\
+ Switch to GGS                                                                       & \multicolumn{1}{l}{$0.44\pm0.02$}                            & \multicolumn{1}{l}{$0.62\pm0.03$}                                                                                                     & \multicolumn{1}{l}{$3.28\pm0.25$}                                                                                                                                                                                               & \multicolumn{1}{l|}{\ding{51}(2/5)}  & \multicolumn{1}{l}{$0.10\pm0.01$}                            & \multicolumn{1}{l}{$0.20\pm0.05$}                                                                                                      & \multicolumn{1}{l}{$3.29\pm0.22$}                                                                                                                                                                                             & \multicolumn{1}{l|}{\ding{55}(0/5)} & \multicolumn{1}{l}{$0.11\pm0.04$}                            & \multicolumn{1}{l}{$0.28\pm0.14$}                                                                                                       & \multicolumn{1}{l}{$3.86\pm0.39$}                                                                                                                                                                                                     & \multicolumn{1}{l}{\ding{51}(3/5)}    \\ 
\hline
\begin{tabular}[c]{@{}l@{}}Genetic GFN \\ (SMILES \& ZINC)\end{tabular}         & 0.14 $\pm$ 0.06                                              & 0.23 $\pm$ 0.10                                                                                                                       & \textcolor{red}{4.98}$\pm$ 0.56                                                                                                                                                                                                          & \ding{55}(0/5)                       & 0.08 $\pm$ 0.03                                              & 0.10 $\pm$ 0.05                                                                                                                        & \textcolor{red}{4.85} $\pm$ 0.31                                                                                                                                                                                                      & \ding{51}(1/5)                      & 1.29 $\pm$ 0.32                                              & 1.71 $\pm$ 0.42                                                                                                                         & \textcolor{red}{4.55} $\pm$ 0.39                                                                                                                                                                                                    & \ding{51}(5/5)                     \\
+ Synth. Dataset                                                                 & 0.22 $\pm$ 0.07                                              & 0.36 $\pm$ 0.09                                                                                                                       & \textcolor{red}{4.89} $\pm$ 0.23                                                                                                                                                                                                     & \ding{55}(0/5)                       & 0.10 $\pm$ 0.04                                              & 0.20 $\pm$ 0.10                                                                                                                        & \textcolor{red}{5.16} $\pm$ 0.13                                                                                                                                                                                                       & \ding{51}(1/5)                      & 0.62 $\pm$ 0.31                                              & 0.85 $\pm$ 0.32                                                                                                                         & \textcolor{red}{5.15} $\pm$ 0.39                                                                                                                                                                                                                & \ding{51}(5/5)                        \\
+ Switch to GGS                                                                  & 0.63 $\pm$ 0.22                                              & 0.91 $\pm$ 0.18                                                                                                                       & 3.59 $\pm$ 0.33                                                                                                                                                                                                                      & \ding{51}(5/5)                      & 0.52 $\pm$ 0.22                                              & 0.75 $\pm$ 0.25                                                                                                                        & 3.35 $\pm$ 0.23                                                                                                                                                                                                                        & \ding{51}(4/5)                      & 0.55 $\pm$ 0.10                                              & 0.79 $\pm$ 0.13                                                                                                                         & 4.17 $\pm$ 0.30                                                                                                                                                                                                                                & \ding{51}(5/5)                    \\
\rowcolor[rgb]{0.949,0.949,0.949}Full method (ours)                                   & 0.78 $\pm$ 0.23                                              & 1.19 $\pm$ 0.42                                                                                                                       & 4.06 $\pm$ 0.30                                                                                                                                                                                                                        & \ding{51}(5/5)                    & 0.33 $\pm$ 0.16                                              & 0.75 $\pm$ 0.48                                                                                                                        & 3.37 $\pm$ 0.42                                                                                                                                                                                                                      & \ding{51}(4/5)                      & 0.42 $\pm$ 0.08                                              & 0.59 $\pm$ 0.16                                                                                                                         & 4.40 $\pm$ 0.37                                                                                                                                                                                                                                 & \ding{51}(5/5)                       \\
\bottomrule
\end{tabular}
}
\vspace{-0.50cm}
\end{table} \\[\myvspace]
\textbf{Baseline Method}\,
We provide a comprehensive ablation study of our genetic GFN framework extensions, systematically evaluating the impact of each component introduced in \cref{sec:methodology}.
\cref{tab:auc_table} shows an excerpt with the most informative configurations.
The full ablation is reported in \cref{tab:auc_table_suppl}. \\[0.5\myvspace]
\uline{TE:} \,
The original SMILES variants fail to reliably find high-performing candidates, as this encoding cannot natively model the two-sided gold binding required for MJs. 
Switching to GGS resolves this topology mismatch and substantially increases AUC and fitness, leading to robust RI discovery (5/5). 
The full method, adding transformer, stability mechanisms and descriptors, achieves the best mean AUC and fitness. 
The transformer leaves fitness unchanged but enables a richer latent space that the descriptors then exploit, providing structure-property intuition. 
As shown in the appendix, the stability measures are decisive to prevent training collapse on this rugged landscape.
Finally, our full method identifies the candidate with the highest $ZT$ value.\\[0.5\myvspace]
\uline{PH:} \,
SMILES-based variants again struggle with the dual binding topology, finding relevant candidates at most sporadically.
GGS is the decisive change here, achieving the highest AUC and fitness and robust RI discovery (4/5).
The full method works robustly with small cost to AUC. 
Notably, the full method still finds the candidate with the lowest thermal conductance.
\\[0.5\myvspace]
\uline{MO:} \,
MO has only one-sided binding and is the only task where the distributional bias from pharmaceutical pretraining appears to help.
The original setup (SMILES + ZINC) performs competitively on AUC and fitness, but its mean SA score lies slightly above the synthesizability threshold.
Removing the ZINC pretraining drops performance and worsens SA.
Switching to GGS brings SA below threshold, which is crucial for practical applications, at the cost of some AUC and fitness.
In contrast to the other tasks, auxiliary descriptors in the full method measurably degrade MO performance, suggesting the auxiliary objective conflicts with this task's optimization.  \\[\myvspace]
\textbf{Summary.}\,
SOTA methods (f-RAG, GenMol) fail outright on NMO, unable to adapt to physical objectives.
Strikingly, the simple training-free molGA robustly finds relevant candidates on MO under the same SMILES interface and default anchor placement available to f-RAG and GenMol, outperforming methods with advanced architectures.
This suggests that recent gains on pharmaceutical benchmarks have come primarily from task-specific priors rather than from better optimization, leaving the field's most sophisticated methods with little to contribute once the prior is removed. \\[0.5\myvspace]
Our ablations reveal that anchor modeling via GGS is a crucial factor for the dual-binding topologies of TE and PH, significantly improving molGA, REINVENT and Genetic-GFN on these tasks. \\[0.5\myvspace]
Our baseline method shows that NMO is solvable and provides an exemplary blueprint combining GGS and synthetic pretraining. 
Our full method finds relevant candidates across all three oracles and delivers the best physical properties on TE and PH, with potential for real scientific impact.
\Cref{sec:solving_NMO} summarizes our key findings on developing effective methods for the NMO benchmark.
To ensure that we do not overfit to NMO ourselves, we evaluate our baseline method on PMO and find it remains competitive (\cref{sec:PMO_benchmark}).

\subsection{Scientific Impact of Top-Performing Molecules}
\label{sec:impact_of_mols}
\begin{figure}      
    \centering
    \includegraphics[width=0.65\linewidth, trim=0mm 1mm 0mm 2mm, clip]{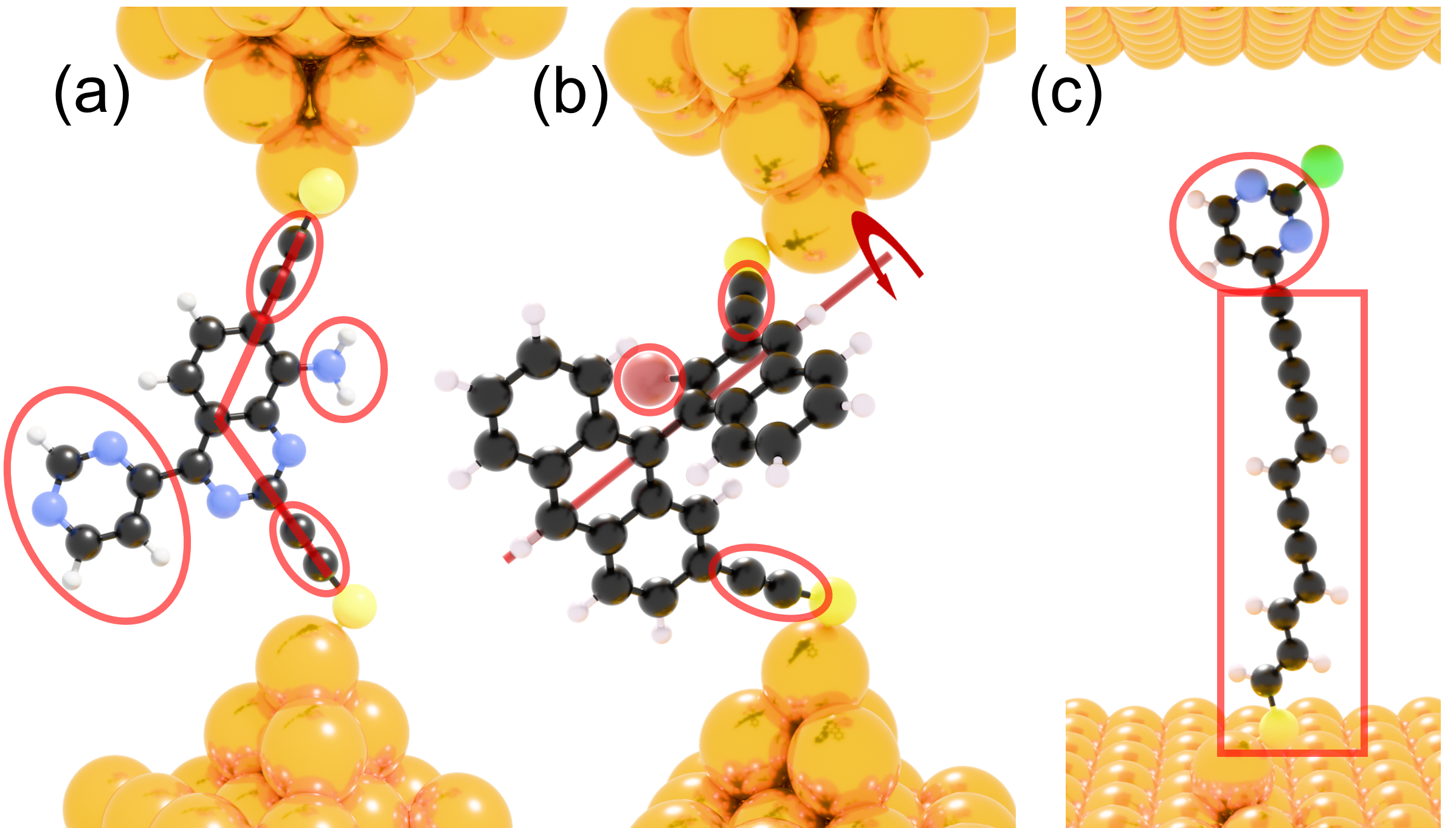}
    \caption{Top performing molecules surpassing previous literature results for (a) TE ($ZT=8.5$; $\mathrm{SA}=3.9$), (b) PH ($\kappa_\mathrm{ph}= 0.1\,\frac{\mathrm{pW}}{\mathrm{K}}$; $\mathrm{SA}=3.18$), and (c) MO tasks ($P=9.9$; $\mathrm{SA}=4.35$). Decisive features identified in \cref{sec:extended_physics_ana} are marked in red.}
    \label{fig:top_molecules}
\end{figure}

Our baseline method finds top-performing candidates across all three oracles, each surpassing previous literature results in their respective fields (dataset provided, see \cref{sec:code_and_data_avail}).
Selected examples are shown in Figure~\ref{fig:top_molecules} and detailed physical analysis is provided in \cref{sec:nmo_details}. In the following, we illustrate the scientific insights and impact that can be drawn from these candidates. 
%
For the TE oracle, the threshold for technological relevance $ZT>3$ from \citet{gemma2021roadmap} is exceeded.
Notably, our approach rediscovers and combines beneficial molecular motifs reported across specialized literature, such as the overall bent structure, side groups, and phonon-mode filtering blocks near the anchors.
For the phonon oracle, the molecule has remarkably low thermal conductance caused by the same end groups and a strategically placed bromine atom that induces twisting.
This candidate improves upon recent literature results \cite{blaschke2025revealing} while achieving substantially better SA scores. 
For the optomechanics oracle, the selected molecule has exceptional upconversion capability (on xtb and DFT level), surpassing the previous best candidate ($P=7.88$) proposed by \citet{koczor2025generative}. 
Importantly, our approach finds a novel molecular design structure: combining a thiol anchor, a conjugated chain and an aromatic end group results in highly Raman and IR-active modes and likely promotes SAM formation.
This structure is distinct from previously reported molecules, highlighting the potential of our framework to explore candidates beyond existing biases.

\subsection{Limitations}
\label{sec:limitations}
Despite careful validation (\cref{sec:nmo_details,sec:extended_physics_ana}), xTB calculations involve a trade-off between accuracy and tractability.
The MO task can overestimate upconversion intensities in the high-fitness tail ($P>15$), and candidates in this regime should be interpreted with care.
An explicit check still indicates strong performance, despite considerably smaller $P$ values at the DFT level (see \cref{sec:thz_results}).
A recently released xTB method \cite{froitzheim2025g} may address this in future work.
Our safeguards cannot catch every undesirable pattern either.
The PH task, for example, has a degenerate regime in which arbitrarily long molecules achieve artificially low thermal conductance.
This is addressed at the benchmark protocol level by permitting length bounds, and is further self-limiting through the computational cost of this regime.
Proposed molecules should therefore be seen as promising candidates, with downstream verification as the natural next step in the scientific process.

Like all fragment-based approaches, our library introduces a degree of bias, which can be controlled explicitly.
In \cref{sec:library_ablation}, we identify the enabling components for the broader community.
High computational cost limits evaluations to five seeds, reducing statistical power.
However, this matches or exceeds standard practice even for considerably cheaper benchmarks \cite{gao2022pmo, kaech2025invirtuogen}.
\section{Conclusion}
\label{sec:conclusion}
We introduced the NMO Benchmark for generative molecular design in nanotechnology.
NMO stands apart from existing benchmarks in two ways: quantum simulations replace proxy oracles, and a strict protocol rewards generalist methods over per-task overfitting.
Our work grounds its tasks in relevant scientific problems while remaining accessible to ML researchers without a physics background. \\
Its tasks pose distinct challenges, notably rugged fitness landscapes and hard structural constraints from molecule-electrode binding.
We show that state-of-the-art generative models struggle on NMO while a simple genetic algorithm stays competitive, echoing TARTARUS~\cite{nigam2023tartarus} in showing that simple methods generalize where sophisticated ones do not, and strongly indicating that the field needs generalist methods rather than benchmark-tuned specialists. \\
NMO surfaces concrete challenges for which we propose solutions: a new encoding ensuring practical synthesizability while natively modeling electrode binding, and a domain-agnostic pretraining strategy that is free of implicit dataset bias.
Our baseline method combines these solutions and surpasses published literature results on all three oracles, identifying candidates with scientific value. \\
NMO is therefore both a rigorous testbed for the ML community and a discovery engine for nanotechnology research, building a sustained interface between two previously disjoint communities to catalyze future work in both.


\begin{ack}
The authors gratefully acknowledge the scientific support and HPC resources provided by the Erlangen National High Performance Computing Center (NHR@FAU) of the Friedrich-Alexander-Universität Erlangen-Nürnberg (FAU) under the NHR project b296ee. NHR funding is provided by federal and Bavarian state authorities. M.B. and F.P acknowledge funding by the German Research Foundation (Deutsche Forschungsgemeinschaft) within the Collaborative Research Center (Sonderforschungsbereich) 1585 (project number 492723217), subproject C02 and acknowledge use of the LiCCA high-performance computing cluster of the University of Augsburg, co-funded by the German Research Foundation (project number 499211671). We thank A. Görz for proofreading the manuscript. Z.KB. acknowledges funding by a UKRI Future Leaders Fellowship (UKRI3089).
\end{ack}

\newpage
\FloatBarrier

\bibliography{main}
\bibliographystyle{IEEEtranN}

\FloatBarrier

\newpage
\appendix
\onecolumn

\FloatBarrier
\begin{center}
    \LARGE \textbf{Technical appendices and supplementary material}
\end{center}

\section*{Table of Contents}
\startcontents[supplementary]
\printcontents[supplementary]{l}{1}{\setcounter{tocdepth}{3}}
\newpage

\section{NMO Benchmark}
\label{sec:nmo_suppl}
\subsection{NMO Task Details}
\label{sec:nmo_details}
In this section, we explain the physical background for the three tasks of the NMO benchmark.
We explain the physical theory, give details on the practical implementation, show validations for the implemented oracles, and place the results in context with the existing literature.
Additionally, we also discuss the limitations of our approach.

\subsubsection{Heat Transport at the Single Molecule Level}
\label{sec:phonon_details}

\begin{figure}[h]
    \centering
    \includegraphics[width=0.8\textwidth]{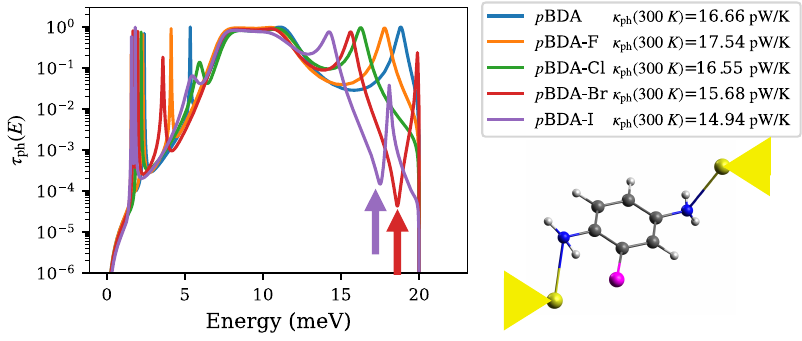}
    \caption{Validation of the phonon transport implementation. Left: Phonon transmission for a set of molecules with the corresponding thermal conductance given in the legend. The purple and red arrows mark destructive quantum interference features of the corresponding molecule. The structure of the molecules is shown on the right. The yellow triangles indicate how the molecule is embedded in a junction. The pink-colored atom in the molecular structures indicates the position of the substituents fluorine ($p$BDA-F), chlorine ($p$BDA-Cl), bromine ($p$BDA-Br), and iodine ($p$BDA-I). }
    \label{fig:phonon_validation}
\end{figure}

The ability to engineer molecules that can efficiently conduct or insulate heat when placed between two macroscopic electrodes opens up new avenues for thermal management in nanoelectronics or novel heat-based devices. 
The thermal conductance is measured in Watt per Kelvin (W/K), which is the power transmitted through the junction per temperature difference between the two electrodes.
Recent advancements in experimental techniques have pushed measurement resolution into the Picowatt per Kelvin (pW/K = $10^{-12}$ W/K) regime \cite{yelishala2025phonon,luan2026tuning, cui2019thermal, mosso:NanoLett2019}, which is the typical scale of thermal conductance through single molecules at room temperature.
Designing molecules with desired transport properties is a complex task due to the interplay between molecular structure and quantum effects \cite{klockner2017tuning, blaschke2025revealing, connectivity5, markussen2013phonon}. 

\subsubsection*{Details on underlying theory}
The phonon-mediated thermal transport can be modeled using the Landauer-Büttiker formalism \cite{buttiker1988absence, Cuevas2017, wang2014nonequilibrium}.
The evaluation of the phononic thermal conductance for the molecular candidates is challenging, as it requires a careful balance between high numerical accuracy and minimal computational time. 
To this end, various numerical schemes have been established, offering different trade-offs between physical rigour and computational demand \cite{burkle2015first, markussen2013phonon}. We adapt the formalism used in \cite{blaschke2025revealing,markussen2013phonon} with a reimplementation optimized for high-throughput calculations.

The thermal conductance $\kappa_{\mathrm{ph}}$ is calculated by:
\begin{equation}
        \kappa_{\mathrm{ph}}(T)=\frac{1}{h}\int_0^{\infty} \mathrm{d}E E \tau_{\mathrm{ph}}(E)\frac{\partial n(E,T)}{\partial T}.
    \label{eq:conductance}
\end{equation}
The decisive part is the energy-dependent phonon transmission function $\tau_{\mathrm{ph}}(E)$, which describes the probability for phonons (vibrations with a given frequency determined by the energy $E$) to be transmitted from one electrode to the other through the molecule.
The weight function ${\partial n(E,T)}/{\partial T}$ is the derivative of the Bose-Einstein distribution $n(E,T)$ with respect to temperature $T$ reflecting the quantum nature of phonon transport. 
Structural changes in the molecule mainly have a local influence on transmission e.g. interferences due to the quantum nature \cite{blaschke2025revealing, connectivity5, markussen2013phonon}. 
Since the thermal conductance is determined as an integral over transmission, it is particularly difficult to achieve a large variation in conductance. 

The phonon transmission $\tau_{\mathrm{ph}}$ of a molecular junction is determined by \cite{mingo2006anharmonic}:
\begin{equation}
    \tau_{\mathrm{ph}}(E) = \mathrm{Tr}\left[ \bm{D}^\mathrm{r}(E) \bm{\Lambda}_\mathrm{L}(E)\bm{D}^\mathrm{a}(E) \bm{\Lambda}_\mathrm{R}(E)\right].
    \label{eq:transmission_ph}
\end{equation}
In this equation, $E$ is the energy, $\bm{D}^\text{r}(E)$ ($\bm{D}^\text{a}(E)$) denotes the so-called Green's function of the molecule and $\bm{\Lambda}_{X}(E)$ are linewidth broadening matrices.
These $\bm{\Lambda}_{X}(E)$ describe the coupling to the electrodes.
The Green's function contains all relevant information about the scattering processes in the molecule. 
The Green's function is calculated from the dynamical matrix $\bm{K}$, which is the mass-weighted Hessian ($K_{ij} = (1/\sqrt{M_i M_j}) \partial^2 E/\partial u_i \partial u_j$) and additional terms called self-energies $\bm{\Pi}_X(E)$ that describe the coupling to the electrodes:
\begin{equation}
    \bm{D}^\text{r}(E) = \left[ (E/ \hbar)^2\bm{1}-\bm{K}-\bm{\Pi}^\text{r}_\text{L}(E)-\bm{\Pi}^\text{r}_\text{R}(E) \right]^{-1}.
    \label{eq:propagator}
\end{equation}
The linewidth broadening matrices are calculated from the self-energies via 
\begin{equation}
\bm{\Lambda}_{X}(E)=-2 \mathrm{Im}\left[ \bm{\Pi}_{X}^{\text{r}}\right]~.
\end{equation}
For further details, we refer the reader to the literature \cite{burkle2015first}.

The self-energies depend on the surface Green's function (SGF) of the electrodes. Similar to the Green's function of the molecule, the SGF function describes the propagation of phonons in the electrodes. 
We choose the SGF according to a Debye model which is a minimalistic model for phonons in metals \cite{mingo2006anharmonic,markussen2013phonon}. 
Within this model, the SGF and the self-energies can be calculated without including parts of the electrodes explicitly in the simulation \cite{burkle2015first}, which significantly reduces the computational cost. 
Previous work has shown that attaching just one gold atom to the anchor groups of the molecule and coupling the Debye model to these gold atoms is sufficient to describe the phonon transport and reproduce important trends from the literature \cite{blaschke2025revealing}. 

\subsubsection*{Details on practical implementation and validation}
The practical implementation includes several steps:
\begin{itemize}
    \item \textbf{GGS Encoding}: Attach the anchor groups (gold-thiol group) to the source and sink fragments of the molecule. 
    We use gold-thiol anchors as they are well-established both experimentally and theoretically \cite{rubio2001mechanical, frisenda2015electrical, van2024mechanoelectric}. 
    For \textbf{SMILES} no anchor positions are defined by the encoding. 
    We choose the first and last (non-hydrogen) atom in the SMILES string as anchor points and attach the gold-thiol groups there.
    \item Optimize the molecular geometry to find a minimum energy configuration. 
    We employ xtb with the GFN1-xTB method \cite{bannwarth2021xtb} for this purpose.
    \item Align molecule anchors along z-axis (gold-gold axis): 
    \item Set fitness of molecules with a HOMO-LUMO gap below $0.2~$eV to zero (hard constraints in Equation \eqref{eq:fitness_struc}).
    \item Set fitness of molecules that cannot form a reasonable junction with the electrodes to zero (hard constraints in \eqref{eq:fitness_struc}).
    This can happen if parts of the relaxed molecule are above or below the anchor groups in z-direction.
    \item Calculate the Hessian matrix using xtb and construct the dynamical matrix.
    \item Calculate the phonon transmission with the described formalism, employing a GPU-accelerated and batch processing optimized implementation of presented formulas. 
    The code outputs the calculated thermal conductance in units of pW/K at a temperature of $300$~K.
\end{itemize}
\textbf{We provide the complete program code, which fully automates the described computational workflow. 
To utilize the implementation, the user only needs to supply the GGS or SMILES string for a molecule.
Within this setting, a non-expert user can employ our code as a benchmark in the future.}

A reference calculation for validation is shown in \cref{fig:phonon_validation}. The figure displays the phonon transmission for a well-established set of molecules where interference effects are induced by specific substitutions \cite{luan2026tuning, klockner2017tuning}.
The first molecule in this set is a so-called $p$BDA (1,4-benzenediamine) junction.
The nitrogen atoms are connected to one gold atom on each side modeling the electrodes as described. 
The set of molecules is obtained by substituting one hydrogen atom in the benzene ring with halogens (fluorine, chlorine, bromine, iodine). 
These are atoms of increasing mass, all heavier than hydrogen.
The position of these substituents is indicated by the colored atom in the right panel of \cref{fig:phonon_validation}. 
Heavy atoms induce destructive quantum interference effects in the phonon transmission above a certain substituent mass.
These interferences can be seen by the dips in the transmission for $p$BDA-Br and $p$BDA-I around $17~$meV (marked with arrows in \cref{fig:phonon_validation}). 
The energy at which these interference dips occur is an important molecular internal feature of the phononic transmission.
The correct energy positions of these dips are reproduced by our implementation \cite{klockner2017tuning}.
Apart from the interference effects, the thermal conductance (given in the caption of \cref{fig:phonon_validation}) is also in good agreement with literature values \cite{blaschke2025revealing}. 
In summary, the presented implementation is able to reproduce important trends from the literature.

The lowest thermal conductance values in literature specially designed to optimize for low thermal conductance with the same modeling are around $\kappa_\mathrm{ph}=0.1-0.4~\mathrm{pW/K}$~\cite{blaschke2025revealing}. 
Those candidates have SA scores around $4$, with a significant portion exceeding our SA score threshold of $4.5$~\cite{vorsilak2020syba}. 
\textbf{We therefore define highly relevant molecules for the NMO task as those with thermal conductance below $0.25~\mathrm{pW/K}$ and an SA score below $4.5$, with particular emphasis on synthetic accessibility.}

\subsubsection{Thermoelectric Efficiency at the Molecular Scale}
\label{sec:te_details}
High thermoelectric efficiency can be achieved by materials with high electrical conductance, high Seebeck coefficient, and low thermal conductance.
This is notoriously difficult because transport coefficients are intrinsically coupled.
High electrical conductance typically implies high thermal conductance, severely limiting performance \cite{majumdar1998lower}.
Single-molecule junctions, however, offer a unique pathway to break these correlations by exploiting discrete molecular orbitals to act as sharp energy filters, thereby decoupling charge and heat transport \cite{gemma2021roadmap}.
Consequently, finding high-performance thermoelectric molecules constitutes a high-dimensional optimization problem: one must navigate a combinatorially vast chemical space to identify candidates that suppress phonon transmission while maintaining favorable electronic properties.

\subsubsection*{Details on underlying theory}
Similar to the phonon transport, the electronic transport is modeled using the Landauer-Büttiker formalism \cite{buttiker1988absence, Cuevas2017}. 
Analogously to the phononic thermal conductance, the electronic conductance is defined by an integral over the corresponding energy-dependent transmission function $\tau_\mathrm{el}(E)$:
\begin{equation}
    G = \frac{2e^2}{h} K_0;~K_n = \int \mathrm{d}E \tau_{\mathrm{el}}(E)\left(-\frac{\partial f(E,T)}{\partial E}\right)(E-\mu)^n~.
    \label{eq:electrical_conductance}
\end{equation}
The integrand is weighted by the derivative of the Fermi--Dirac distribution $f(E,T)$, as the charge carriers (electrons) follow Fermi statistics.
This derivative, $-\partial f / \partial E$, is energetically sharply peaked in the vicinity of the chemical potential $\mu$, which coincides with the Fermi energy $\mu\approx E_\mathrm{F}$ of the electrodes at low temperature. 
Consequently, the electronic conductance is determined by the transmission $\tau_\mathrm{el}(E)$ within this narrow energy window around $E_\mathrm{F}$.
Additional quantities derived from the electronic transport are the electronic contribution to the thermal conductance $\kappa_\mathrm{el}$ (electrons also carry heat) and the Seebeck coefficient $S$. 
These are defined via the integrals $K_n$ introduced in Eq.~\eqref{eq:electrical_conductance}~\cite{burkle2015first}:
\begin{equation}
    \kappa_\mathrm{el} = \frac{2}{hT}\left(K_2 - \frac{K_1^2}{K_0}\right)
\end{equation}
\begin{equation}
    S = -\frac{1}{eT}\frac{K_1}{K_0}
\end{equation}
The Seebeck coefficient $S$ quantifies the thermoelectric voltage induced by a temperature difference across a molecular junction. 
It is measured in Volts per Kelvin (V/K).
Typical values for single-molecule junctions are in the range of several tens of microvolts per Kelvin ($\mu$V/K) \cite{reddy2007thermoelectricity,van2024mechanoelectric}.

The electronic transmission function is calculated by a trace over Green's functions and linewidth broadening matrices, similar to the phononic case (see Eq.~\eqref{eq:transmission_ph}):
\begin{equation}
    \tau_{\mathrm{el}}(E) = \mathrm{Tr}\left[ \bm{G}^\mathrm{r}(E) \bm{\Gamma}_\mathrm{L}(E)\bm{G}^\mathrm{a}(E) \bm{\Gamma}_\mathrm{R}(E)\right].
    \label{eq:transmission_el}
\end{equation}
The electronic Green's functions $\bm{G}^\mathrm{r}(E)$ are derived from the Hamiltonian $\bm{H}$ and overlap matrix $\bm{S}$ of the molecule:
\begin{equation}
    \bm{G}^\text{r}(E) = \left[ E\bm{S}-\bm{H}-\bm{\Sigma}^\text{r}_\text{L}(E)-\bm{\Sigma}^\text{r}_\text{R}(E) \right]^{-1}.
    \label{eq:electronic_propagator}
\end{equation}
We use the xtb program package and its pytorch implementation to calculate $\bm{H}$ and $\bm{S}$ \cite{bannwarth2021xtb,friede2024dxtb}. The self-energies $\bm{\Sigma}_X(E)$ describe the coupling to the electrodes and are connected to the linewidth broadening matrices via $\bm{\Gamma}_{X}(E)=-2 \mathrm{Im}\left[ \bm{\Sigma}_{X}^{\text{r}}\right]$. Further details can be found in the literature \cite{pauly2008cluster}.

An accurate description of the electrodes is essential for modeling electronic transport. 
While approximate methods analogous to the model used for phononic transport can operate without an explicit calculation of the electrode structure (e.g., the wide-band limit \cite{verzijl2013applicability,blaschke2023designing}), we incorporate parts of the electrodes to ensure a proper level alignment and a well-defined Fermi energy \cite{pauly2008cluster}. 
This is crucial for the accurate determination of electronic transport properties, most notably the Seebeck coefficient \cite{burkle2015first}. The calculated molecule is therefore extended by gold clusters at both ends, which is depicted in \cref{fig:thermoelectric_junction}(b),(c).

Within this setting the SGF needed for the self-energies $\bm{\Sigma}^r_X(E)$ can be calculated via a recursive approach \cite{sancho_highly_1985,pauly2008cluster}. 
We follow the ideas presented in \cite{Hierarchical}, with our own reimplementation optimized for high-throughput calculations on GPUs and CPUs. 
The SGF is calculated once on an energy grid and is reused for all transport calculations. 
The starting point is a periodic DFTB calculation \cite{aradi2007dftb} employing the xtb Hamiltonian for the system depicted in \cref{fig:thermoelectric_junction}(a). 
This so-called supercell can be decomposed into principal layers (PL) along the transport direction indicated by the red boxes.
The Hamiltonian of each PL and the coupling between adjacent PLs are converted to real space from the periodic calculation via an inverse Fourier transform.
The resulting matrices are then used to calculate the SGF via the recursive Green's function approach \cite{sancho_highly_1985}.
By including the structure of one PL explicitly in the calculation of the geometry of the junction (see \cref{fig:thermoelectric_junction}(b)), the precalculated SGF can be directly used to calculate the self-energies $\bm{\Sigma}^r_X(E)$. 
More details on this procedure can be found in \cite{pauly2008cluster,Hierarchical}.

\subsubsection*{Limitations}
Predicting the Seebeck coefficient accurately is particularly challenging, as it depends on the slope of the transmission function at the Fermi energy \cite{burkle2015first} and is therefore highly sensitive to the precise alignment of molecular energy levels with respect to $E_\mathrm{F}$ \cite{yan2024substituents,gemma2023full,burkle2015first}.
This could, for example, be fine-tuned through gating in an experimental setting \cite{chen2024quantum}.
However, explicit electrode inclusion establishes a solid foundation \cite{pauly2008cluster, Hierarchical}.
In addition, DFT-based methods are known to overestimate the conductance (transmission in the gap) due to fundamental limitations of the method \cite{perdew_density_1985} and even dynamic effects due to thermal fluctuations have to be considered for a good agreement with experiments \cite{van2024mechanoelectric}. 

Despite these shortcomings, the employed approach has recently demonstrated considerable predictive power and remains the standard for modeling electronic transport properties in single-molecule junctions.

\subsubsection*{Details on practical implementation and validation}
\begin{figure}[h]
    \centering
    \includegraphics[width=0.8\textwidth]{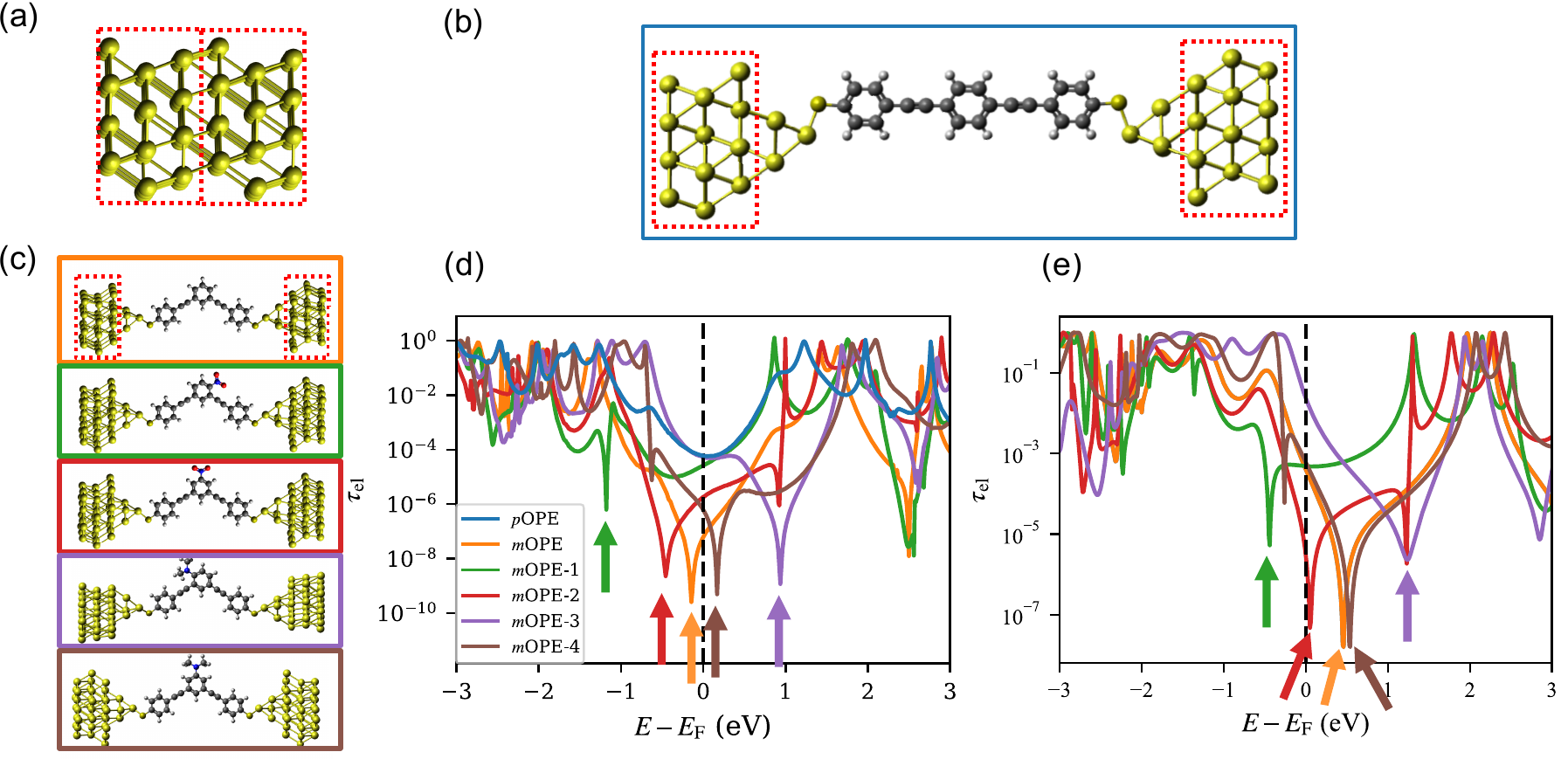}
    \caption{(a) Supercell containing two principal layers (red boxes) for periodic calculation. (b) Junction geometry used for the electronic transport calculation of the $p$OPE molecule. Gold electrodes containing one principal layer (red dashed box) from (a) are attached to each side together with additional gold atoms for a proper junction geometry. (c) All $m$OPE molecule derivatives used to benchmark the transport calculation. (d) Electronic transmission $\tau_\mathrm{el}$ as function of energy. The black dashed line indicates the Fermi energy. Colors of the curves are chosen according to the box colors in (b) and (c). Arrows indicate the destructive quantum interference features. (e) Transmission  for the $m$OPE3 molecules calculated in a recent publication \cite{yan2024substituents}. Arrows indicate the destructive quantum interference features.}
    \label{fig:thermoelectric_junction}
\end{figure}

We provide the complete program code along with all input scripts for the quantum chemistry calculations, needed to reproduce the results and run the NMO benchmark suite.
Similar to the phonon transport, our implementation fully automates the computational workflow starting from a GGS or SMILES string.
In addition to the steps outlined for phonon transport, the workflow automatically attaches gold clusters to the molecule to model the electrodes (see \cref{fig:thermoelectric_junction}(b),(c)) and calculates all electronic transport properties. 
The electrical conductance is computed in units of the conductance quantum $G_0 = 2e^2/h$, the Seebeck coefficient in $\mu$V/K, and the electronic thermal conductance in pW/K at a temperature of $300$~K. 
In literature, the thermoelectric figure of merit is often reported as $ZT$, which is the product of $Z$ and the absolute temperature $T$, making it dimensionless.
\textbf{Alignment, geometry optimization, attachment of electrodes, transport calculation, filtering for reasonable junction geometries and extraction of transport coefficients are performed automatically.
This allows the non-expert user to easily evaluate the thermoelectric properties of arbitrary molecules.}

We validate our implementation by a sensible test set of molecules shown in \cref{fig:thermoelectric_junction} featuring very characteristic electronic transport properties \cite{yan2024substituents}. 
The first molecule is a so-called $p$OPE junction displayed in \cref{fig:thermoelectric_junction}(b). 
It has a transmission valley around the Fermi energy in \cref{fig:thermoelectric_junction}(d), typical for single-molecule junctions.
We remind the reader that the transmission around the Fermi energy (black dashed line in \cref{fig:thermoelectric_junction}(d,e)) directly determines the conductance via Equation \eqref{eq:electrical_conductance}.
Changing from the so-called para configuration to the meta configuration $m$OPE induces a destructive quantum interference dip visible below the Fermi energy in \cref{fig:thermoelectric_junction}(d) indicated by the orange arrow.
This showcases the complexity of the optimization problem as a small structural change (changing the position of one bond) has a profound influence on the electronic structure and therefore on the transport properties.
Especially destructive quantum interferences strongly influence the Seebeck coefficient \cite{van2024mechanoelectric}.
Therefore, a reasonable modeling of these molecular features is essential, which we test using the set of $m$OPE molecules shown in \cref{fig:thermoelectric_junction}(c).

The molecules in the test set contain side groups attached to the central rings with decisive influence on the electronic structure.
The energy positions of the interference features shift depending on the side group \cite{yan2024substituents}. 
Literature calculations from \cite{yan2024substituents} are shown in \cref{fig:thermoelectric_junction}(e).
The relative energetic ordering of these interference features is correctly reproduced by our implementation. 
The energy ordering in \cref{fig:thermoelectric_junction}(d,e)  is as follows (from low to high energy): $m$OPE-1, $m$OPE-2, $m$OPE, $m$OPE-4, and $m$OPE-3. 
This validates the electronic part of our implementation.
The reference publication \cite{yan2024substituents} does not report thermoelectric efficiency values, preventing direct comparison.
We list all the transport properties calculated with our implementation for the $m$OPE molecules in Table \ref{tab:full_transp_properties} for $T=300~\mathrm{K}$.
This shows that the calculated range of the thermoelectric figure of merit $ZT$ spans two orders of magnitude for this set of molecules.
All candidates are far from the limit of $ZT>3$ needed for practical applications \cite{gemma2021roadmap}, highlighting the challenge of finding high-performance thermoelectric molecules.
We note that our implementation employs a more computationally efficient method (GFN-xTB, \cite{friede2024dxtb}) compared to the reference (DFT with PBE functional \cite{perdew1996generalized}) being at least two orders of magnitude faster.

Quantitative validation is challenging due to the variety of theoretical approaches and approximation levels in the literature. 
We therefore additionally compute transport properties for the HBT-OPE3-A molecule from the first direct measurement of thermoelectric efficiency at room temperature reported in \citep{gemma2023full}. 
In that work, $ZT \approx 1.3 \times 10^{-5}$ was measured with transport coefficients $G = 1.82 \times 10^{-4}~G_0$, $S = -8.7~\mu\text{V/K}$, and $\kappa = \kappa_{\mathrm{el}} + \kappa_{\mathrm{ph}} = 24~\text{pW/K}$.
Our method yields $ZT = 1.52 \times 10^{-3}$ with $G = 2.3 \times 10^{-3}~G_0$, $S = 17.99~\mu\text{V/K}$, and $\kappa = \kappa_{\mathrm{el}} = 11.43~\text{pW/K}$. 
Considering the known limitations of the approach listed above, the agreement is reasonable. 
The corresponding molecular geometry is included in the validation set of the provided code for interested readers.
Theoretical values for $ZT$ in single-molecule junctions with gold electrodes using comparable levels of theoretical modeling range from $ZT=1.4$~\cite{OligoyneZT} to $ZT=2.4$~\cite{ZT2}.

Achieving quantitative agreement with experiments remains an open challenge in the field \cite{gemma2023full,van2024mechanoelectric} (see the limitations discussed above). However, proposing potential high-performance candidates may catalyze the field and drive further experimental and theoretical advancements \cite{UnresolvedQuestions}. \textbf{We therefore define a threshold for promising candidates as $ZT > 3$ at room temperature, in line with~\cite{gemma2021roadmap}, combined with an SA score below $4.5$~\cite{vorsilak2020syba} to ensure synthetic accessibility.}

\begin{table}[htbp]
    \centering
    \caption{Quantum Transport Properties of Molecular Candidates in \cref{fig:thermoelectric_junction}.}
    \label{tab:full_transp_properties}
    \resizebox{0.5\linewidth}{!}{%
    \begin{tabular}{l | cccc | c}
    \toprule
    \textbf{} & \makecell{\textbf{G} \\ ($G_0$)} & \makecell{\textbf{S} \\ ($\mu$V/K)} & \makecell{\textbf{$\kappa_{el}$} \\ (pW/K)} & \makecell{\textbf{$\kappa_{ph}$} \\ (pW/K)} & \makecell{\textbf{ZT} \\ ($T=300~\mathrm{K}$)} \\ 
    \midrule
    $m$OPE & 6.94E-08 & -90.87 & 3.99E-05 & 11.07 & 1.20E-06 \\ \hline
    $m$OPE-1 & 2.81E-05 & -28.36 & 1.64E-02 & 13.87 & 3.78E-05 \\ \hline
    $m$OPE-2 & 1.82E-06 & -37.51 & 1.02E-03 & 11.26 & 5.29E-06     \\ \hline
    $m$OPE-3 & 5.10E-05 & 15.66 & 3.15E-02 & 8.83 & 3.28E-05 \\ \hline
    $m$OPE-4 & 6.73E-07 & 80.90 & 3.76E-04 & 10.48 & 9.77E-06 \\ \hline
    \bottomrule
    \end{tabular}%
    }
\end{table}

\subsubsection{Molecular Optomechanics: Terahertz Detection Based on Self-Assembled Monolayers in Nanoplasmonic Cavities}
\label{sec:thz_details}

Detecting THz radiation through molecular optomechanics relies upon the  unique capabilities of specific molecular vibrational modes to absorb THz/MIR radiation and simultaneously to scatter light inelastically from a Vis/NIR laser source. 
Nanoscale detection is achievable with the help of a dual nanoantenna construct that provides substantial electromagnetic field enhancements in both the THz/MIR (incoming) and the Vis/NIR (incoming and outgoing fields) range.

\subsubsection*{Details on underlying theory}
In this technique, THz radiation is detected indirectly as an increase in the Raman anti-Stokes scattering signal ($\Delta I^\mathrm{aS}_m$) at the frequency of the active vibrational mode $m$ \cite{koczor2021molecular}. 
\begin{equation}
\Delta I^\mathrm{aS}_m = \frac{I^\mathrm{A}_\mathrm{L} I^\mathrm{aS}_\mathrm{L}}{h\nu^\mathrm{A}}N\tau_m\langle \sigma^\mathrm{A}_m \sigma^\mathrm{aS}_m \rangle \sim I_m^\mathrm{c} \cdot g^{-2} \cdot S^{-1}~.
\end{equation}
This quantity depends on the power density ($I^\mathrm{A}_\mathrm{L}$) and energy ($h\nu^\mathrm{A}$) of the THz/MIR radiation to be detected, and the power density of the Vis/NIR laser ($I^\mathrm{aS}_\mathrm{L}$) used for inducing the Raman anti-Stokes effect.
Crucially, the signal is influenced by a range of physical properties of the molecule: the number of molecules that are present in the nanocavity ($N$), the vibrational lifetime ($\tau_m$), the absorption cross section ($\sigma^\mathrm{A}_m$) and Raman anti-Stokes cross section ($\sigma^\mathrm{aS}_m$) of the normal mode in question. 
Following  the approximations detailed in \citet{koczor2021molecular}, the performance of a molecular vibrational mode for THz detection depends on (1) the frequency up-conversion capability of the mode ($I_m^\mathrm{c} $), (2) the length of the molecule ($g$) perpendicular to the surface, which influences the field enhancement in the nanocavity approximately as $g^{-2}$ \cite{baumberg2019extreme}, and (3) the surface area covered by the molecule ($S$) which determines how many molecules fit within the nanocavity to contribute to the signal ( $N \sim S^{-1}$).
The intensity for up-conversion is calculated by:
\begin{equation}
I_m^\mathrm{c}= c \frac{(\bar{\nu}^\mathrm{aS}+\bar{\nu}_m)^4}{\bar \nu_m} \langle |{\underline{e}}{\underline{\mu}}'_m|^2 |{\underline{e}}{\underline{\underline{\alpha}}}'_m {\underline{e}}|^2 \rangle~.
\end{equation}
The relation considers the well-known wavenumber dependent scaling term of Raman anti-Stokes scattering \cite{le2008principles}, where the wavenumber of the Vis/NIR laser is given by $\bar{\nu}^\mathrm{aS}$ and the wavenumber of the vibrational mode by $\bar{\nu}_m$. 
In the current work, we consider a NIR laser with a wavelength of 785nm.   
As $I_m^\mathrm{c} $ is used to quantify an increase in anti-Stokes intensity due to the presence of the THz radiation, it does not contain the usual exponential scaling factor considering the thermal (Bose Einstein) occupancy of vibrational levels in Raman spectroscopy. 
The constant scaling factor is detailed in \citet{koczor2021molecular}.
${\underline{\mu}}'_m$ and ${\underline{\underline{\alpha}}}'_m$ denote the dipole derivative vector and the polarizability derivative tensor with respect to normal mode $m$, and ${\underline{e}}$ denotes the THz/MIR and Vis/NIR field polarization vectors which are all aligned in the surface normal direction in the device. 
The angle brackets correspond to an averaging over relative orientations of the molecule and ${\underline{e}}$. 
Here, we consider all possible orientations using the analytical formula derived in \citet{koczor2021molecular}, but future work can explore more realistic sampling of possible molecular orientations within the nanocavity.

To arrive at a molecular-level property, \citet{koczor2021molecular} introduced the dimensionless quantity $P$ that collects all modes with vibrational frequencies within the experimentally relevant THz-MIR range ($M$), considered as 30-1000 cm$^{-1}$:
\begin{equation}
P = \frac{1}{\sigma} \left( \log_{10}\left(\sum_{m \in M} I_{m}^{c}\right) -\mu \right)~,
\label{eq:P_value_def}
\end{equation}
where the logarithm and standardization were applied to facilitate the training of ML-based predictors for $P$. 
We use the transformation parameters $\sigma$ (standard deviation) and $\mu$ (mean) determined for a set of randomly selected commercial molecules in \citet{koczor2021molecular} to enable direct comparison with previous works \cite{koczor2021molecular,koczor2022molecular,koczor2025generative}.  
To keep the relative importance of up-conversion capability and geometrical constraints unchanged, the physical property contribution ($P+F$) to the fitness function applies similar logarithmic and scaling transformations in $F$:
\begin{equation}
F = - \frac{2}{\sigma} \log_{10}(g) - \frac{1}{\sigma} \log_{10}(S)
\label{eq:G_up}
\end{equation}

\subsubsection*{Details on practical implementation and validation}
\begin{figure}
    \centering
    \includegraphics[width=0.35\textwidth]{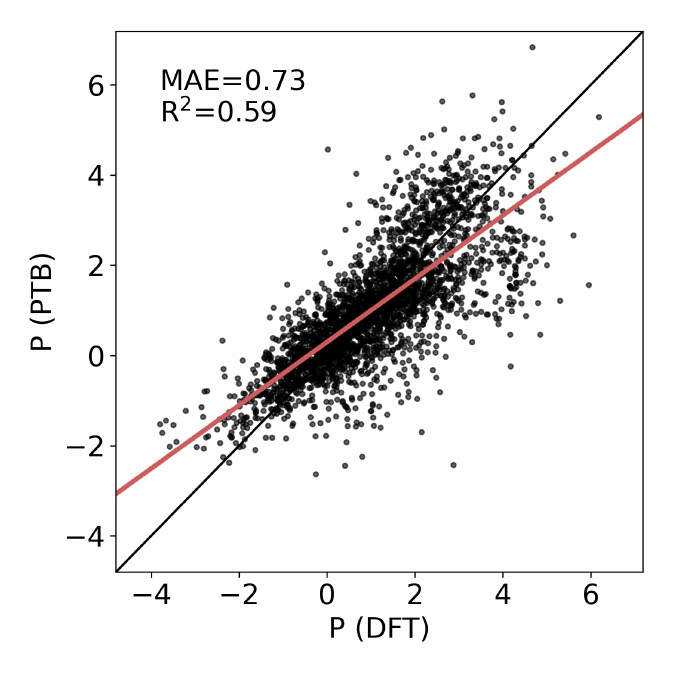}
    \caption{Comparison of PTB and DFT calculated $P$ values for the \textit{Gold} database of Molecular Vibration Explorer \cite{koczor2022molecular} (about 2800 molecules), showing the mean absolute error (MAE) and coefficient of determination (R$^2$). The 1:1 line is shown in black while the linear fit to data points is shown in red. }
    \label{fig:ptb_validation}
\end{figure}

We have implemented the calculation of $I_m^\mathrm{c}$ and $P$ within the semi-empirical PTB module \cite{grimme2023non} of the xtb package \cite{bannwarth2021xtb, friede2024dxtb}. 
This methodology combines vibrational frequencies calculated with GFN2-xTB and intensities (from dipole moment and polarizability derivatives) calculated with the PTB method. 
We validate the implementation against the publicly available \textit{Gold} database of Molecular Vibration Explorer \cite{koczor2022molecular} that contains DFT calculations for about 2800 molecules, performed with the B3LYP functional \cite{becke1988density,lee1988development}, D3 dispersion correction \cite{grimme2010consistent} and def2-SVP basis set \cite{weigend2005balanced}.
The molecules are modeled as gold-thiolates (where the hydrogen atom of a thiol group is replaced by a single gold atom), which has been shown in previous studies \cite{griffiths2021resolving,boehmke2024uncovering, wright2021mechanistic, koczor2021molecular, xomalis2021detecting} to give a good agreement with surface-enhanced Raman scattering and frequency up-conversion measurements in nanocavities. 
For a more detailed discussion on the reliability of this level of modeling see \citet{koczor2025generative}.
The PTB calculated $P$ values show satisfactory correlation with DFT calculated values (\cref{fig:ptb_validation}), however, with larger errors for high $P$ (DFT) values.
Its mean absolute error is 0.73, which is higher than that of ML-based predictors \cite{koczor2021molecular,koczor2025generative}. 
However, the PTB method has the advantage of not relying on predefined training sets, which makes it ideal for the current approach.
Additionally, PTB is expected to perform better for a wide range of generated molecules, whereas ML predictors tend to struggle with out-of-distribution generated molecules \cite{koczor2025generative}.
In summary, the PTB method is found adequate for quickly estimating the up-conversion capability of molecules within the design workflow, but computationally more demanding DFT calculations are required for verifying the PTB predictions on the proposed top molecules. 
In particular, PTB-calculated $P$ values above $\sim$15 were always found to be severe overestimations according to our DFT reference calculations (see Section~\ref{sec:thz_results}), thus for such molecules we strongly advise DFT validation of the spectroscopic properties.

For the calculation of the fitness function, the shape of the molecule must also be taken into account, since it influences $g$ and $S$ (see Eq. \eqref{eq:G_up}).
This was done by defining the start and end points of the molecule. 
The start point is where the gold–thiol group is attached, by which the molecule is bonded to the substrate, while the end point is used to orient the molecule.
In the GGS case, the start and end points naturally defined in the encoding are used for this purpose. 
The gold–thiol group is attached at the start point (source node), and the axis defined by the start and end points is aligned along the z-axis (which is the direction perpendicular to the gold surfaces). 
In the SMILES case, analogous to the treatment used for phononic and electronic transport, the first and last non-hydrogen atoms are used to define the start and end points, since these are not canonically defined in the SMILES string.
The length of the molecule along the z-axis gives $g$, and the surface area covered per molecule $S$ is calculated as the base of the smallest cylinder that encloses the molecule.

\textbf{For identifying promising candidates, we define a threshold of $P > 7.88$, the highest value reported in \citet{koczor2025generative}, combined with an SA score below $4.5$~\cite{vorsilak2020syba} to ensure synthetic accessibility.}

\textbf{The full workflow for calculating $P$ from a GGS or SMILES string is fully automated in our implementation including all needed steps like geometry relaxation and alignment.
This allows non-expert users to readily employ our benchmark.}

\subsubsection{Thresholds for Relevant Molecules}
\label{sec:thresholds}
In this section we summarize the thresholds defined in \cref{sec:phonon_details,sec:te_details,sec:thz_details} for defining highly relevant molecules for the three tasks, which are based on a combination of literature values and synthetic accessibility considerations.
Our provided analysis scripts automatically check these thresholds for the generated molecules to identify promising candidates for each task.

 \begin{itemize}
        \item \textbf{Phonon task (PH)}:\,  $k_\mathrm{ph}<0.25\mathrm~{pW/K}$ \& $\mathrm{SA}<4.5$. Rivals the literature \cite{blaschke2025revealing} while strictly ensuring synthesizability by SA threshold \cite{vorsilak2020syba}, overcoming the poor SA score of theoretical baselines. 
        \item \textbf{Thermoelectric task (TE)}:\,  $ZT>3$ \& $\mathrm{SA}<4.5$. Exceeds threshold for technological relevance ($ZT>3$) \cite{gemma2021roadmap}. Same SA threshold as PH to ensure synthesizability.
        \item \textbf{Molecular Optomechanics task (MO)}:\,  $P>7.88$ (xtb Level) \& $\mathrm{SA}<4.5$. Exceeds the highest value reported in \citet{koczor2025generative} (7.88) while ensuring synthesizability with the same SA threshold as PH and TE.
\end{itemize}

\subsubsection{Compute Resources}
\label{sec:compute_ress_benchmark}
Evaluating the phonon transport workflow for the molecule set in \cref{fig:phonon_validation} (calculating each molecule 5 times to build a considerably large batch) takes $34.0~\mathrm{s}$ on 16 cores of an AMD EPYC 7452 processor. 
The transport calculation can be accelerated using the GPU implementation, which reduces the time to $26.4~\mathrm{s}$ on an NVIDIA Tesla V100S GPU with the same CPU hardware.

For the larger molecules in \cref{fig:thermoelectric_junction} (again calculating each molecule 5 times), the workflow for calculating $ZT$ takes $350~\mathrm{s}$ on 16 cores of an AMD EPYC 7452 processor. The GPU implementation (same CPU setting) reduces this to approximately $200~\mathrm{s}$ on an NVIDIA Tesla V100S GPU.

Calculating the $P$ value for the molecule shown in \cref{fig:top_molecules}(c) using xtb takes approximately $20~\mathrm{s}$ on 16 cores of an AMD EPYC 7452 processor (without structure relaxation). 
Since we implemented this functionality directly in the standalone CPU version of xtb, we report single-molecule timings rather than batch timings.

All methods require a prior structure relaxation, which is performed at the xtb level. Extensive timings and scaling behavior for xtb are provided in \citet{bannwarth2021xtb}.

\subsection{Molecular Filters}
\label{sec:filters}
To enhance the chemical stability of the generated molecules, we iteratively developed a suite of heuristic filters utilizing SMARTS (SMILES Arbitrary Target Specification) based substructure searches \cite{daylight2019smarts}. 
SMARTS is a line-notation language based on SMILES that enables the precise description of molecular substructures and patterns using logical operators and wildcards. 
Our filter set is designed to identify and exclude reactive or inherently unstable moieties within the molecular graphs, which can be created by combining building blocks in \cref{fig:building_blocks} or during the SMILES-based reference runs.
Specifically, the SMARTS patterns detailed below were selected for exclusion because the corresponding substructures are known to induce instability—for instance, through susceptibility to hydrolysis or degradation under thermal and photochemical conditions. 
To maximize sampling efficiency, these filters are applied as a preprocessing step prior to the oracle call. 
Consequently, they do not contribute to the total computational budget.
The patterns are listed below, with each entry naming the substructure or chemical group followed by its SMARTS representation:
\begin{lstlisting}[basicstyle=\ttfamily\scriptsize, breaklines=true, frame=single, language=Python]
    Polyyne: "C#CC#CC#C",
    Cumulene: "C=C=C=C",
    Peroxide: "[OX2,OX1]-[OX2,OX1]",
    Hydrazine: "[N&!R]-[N&!R]",
    Anhydride_Instability: "[#6](=O)-[#8]-[#6](=O)",
    Acyl_Urea_Instability: "[#6](=O)-[#7]-[#6](=O)",
    Heteroatom_Alkyne: "[#8,#7]-[#6]#[#6]",
    Acid_Halide: "[CX3](=[OX1])[F,Cl,Br,I]",
    Geminal_Heteroatoms: "[CX4](-[O,N,S,F,Cl,Br,I])(-[O,N,S,F,Cl,Br,I])",
    N_or_O_Halogen: "[#7,#8]-[F,Cl,Br,I]",
    Nitro_Instability: "[#7,#8]-[#7+](=O)[O-]",
    Phosphorus_Halogen: "[#15]-[F,Cl,Br,I]",
    Phospho_Anhydride: "[#15]-[#8]-[#15]",
    Acyl_Cyanide: "[CX3](=O)-[#6]#[#7]",
    Imidoyl_Halide: "[F,Cl,Br,I]-[#6]=[#7]",
    Alpha_Halo_Carbonyl: "[CX3](=O)-[CX4]-[Cl,Br,I]",
    Diazonium: "[N+]#[N-]",
    Nitroso: "[!c]-N=O",
    Ketene: "[#6]=[#6]=[#8]",
    Ketenimine: "[#6]=[#6]=[#7]",
    Carbodiimide: "[#7]=[#6]=[#7]",
    Azo: "[!c]-[#7]=[#7]",
    Pentalene: "[C&R2&r5&!r6&!r7&^2]@[C&R2&r5&!r6&!r7&^2]",
    Enamine_H: "[#7;H1,H2]-[#6;!a]=[#6]",
    Strained_3_Ring: "[r3]",
    Strained_4_Ring: "[r4]",
    Cyclic_Alkyne: "[#6]#[#6;R&r3,r4,r5,r6,r7]",
    Unstable_N_S: "[#7]-[#16]",
    Terminal_Alkene: "[#6;R]=[#6;D1]",
    Quinoid: "[#6]=[c;R]([c;R])"
\end{lstlisting}

This limited set of SMARTS filters cannot account for all undesirable structural motifs possible. 
Thus, we incorporate an additional criterion based on the HOMO-LUMO gap into the oracle call as described in the main text (hard constraints $\Theta$ in Equation \eqref{eq:fitness_struc}). 
Furthermore, our trained sampler does not merely output a single optimized molecule. Instead, it draws from the learned distribution, which significantly increases the probability of identifying successful candidates.

\subsection{Encoding Details}

In this section we give further details on our novel Graph Group SELFIES (GGS) encoding and explain an example of a Group SELFIES (GS) string in detail.
Moreover, we provide a description of the Genetic Algorithm (GA) and how it is designed to work with GGS.

\subsubsection{Details on Graph Group SELFIES}
\label{sec:ggs_details}
The GGS encoding is specialized for the optimization of molecular properties.
GGS represents molecules as directed acyclic graphs (DAGs), while using the regular GS string format as its textual notation.
We refer to the string representation of such a DAG as a GGS string.
For example, the random generation of valid molecules for the synthetic dataset is performed at the graph level and subsequently translated into GGS strings. 
The agent generates GGS strings (see \cref{sec:action_space}), which are then translated into graphs for internal handling.
If such a string can be translated into a GGS graph, the resulting molecule is always valid, without truncations or ambiguities as encountered in regular GS. 
However, the agent may still produce invalid strings, for example strings that do not conform to the scheme defined in \cref{sec:ggs}.

A current limitation of our implementation is a maximum branching depth (pop depth) of one. 
This means that, for example, in \cref{fig:mutation}, no fragment can follow $\mathrm{frag\_2}$. 
This can be lifted in future work, but it currently provides a favorable trade-off between representational power and complexity for many applications. 
Furthermore, we do not include special tokens to handle chirality or stereochemistry in the current implementation as these features are not relevant for the physics-based tasks considered here in combination with the geometry optimization step.

To keep in line with the original GS and SELFIES works \cite{krenn2020selfies,cheng2023groupselfies}, we use an overloaded token table where the outgoing couplings from a fragment $S_\mathrm{out}$ are encoded according to Table \ref{tab:token_index_mapping}.
In this way our GGS strings can be directly decoded using existing GS and SELFIES decoders.
We also adopt modulo arithmetic for $S_\mathrm{in}$ and $S_\mathrm{out}$ present in regular GS. If $S_\mathrm{in}$ or $S_\mathrm{out}$ exceed the number of available anchor points, the values are taken modulo the number of anchor points.

\begin{table}[htbp]
    \centering
    \caption{Mapping between textual $S_\mathrm{out}$ tokens in GS encoding and their corresponding numeric values. The overload table is used in GS and SELFIES so that the same tokens can act both as structural symbols and as numeric values (e.g. for ring closures or branch lengths). This avoids introducing a separate numeric vocabulary, keeping the representation compact, unambiguous, and easier to decode.}
    \label{tab:token_index_mapping}
    \begin{tabular}{|l|l!{\vrule width 1.8pt}l|l|}
        \hline
        token & index & token & index \\ \hline
        {[}C{]}        & 0  & {[}=N{]}        & 8  \\ \hline
        {[}Ring1{]}    & 1  & {[}=C{]}        & 9  \\ \hline
        {[}Ring2{]}    & 2  & {[}\#C{]}       & 10  \\ \hline
        {[}Branch{]}   & 3  & {[}S{]}       & 11  \\ \hline
        {[}=Branch{]}  & 4  & {[}P{]}      & 12 \\ \hline
        {[}\#Branch{]} & 5  &         &  \\ \hline
        {[}O{]}        & 6  &         &  \\ \hline
        {[}N{]}        & 7  &                &    \\ \hline
    \end{tabular}
\end{table}

\subsubsection{Detailed Explanation of a Group SELFIES Example}
\label{sec:GS_example_explanation}
As the Group SELFIES (GS) encoding might not be widely known, we provide a detailed explanation of how it works for the example shown in \cref{fig:encoding_overview}(b).
The Group SELFIES string is: 

\texttt{[:0frag\_1][Ring1][:0frag\_2][pop][=Branch][:1frag\_3]}.

We go through the string part by part:
\begin{itemize}
    \item \texttt{[:0frag\_1]}: Take attachment point $0$ of \texttt{frag\_1} as starting point. Mark attachment point $0$ as the current attachment point for this fragment.
    \item \texttt{[Ring1]}: According to \cref{tab:token_index_mapping}, this translates to a relative shift of $+1$. Add $1$ to the current attachment point. Now the current attachment point is $1$ in \texttt{frag\_1}.
    \item \texttt{[:0frag\_2]}: Add \texttt{frag\_2}, connecting its attachment point $0$ to the current attachment point ($1$) of \texttt{frag\_1}.
    \item \texttt{[pop]}: Move back to the parent fragment, which is \texttt{frag\_1}. The current attachment point in \texttt{frag\_1} is still $1$.
    \item \texttt{[=Branch]}: According to \cref{tab:token_index_mapping}, this translates to a relative shift of $+4$. Add $4$ to the current attachment point. Now the current attachment point is $5$ in \texttt{frag\_1}.
    \item \texttt{[:1frag\_3]}: Add \texttt{frag\_3}, connecting its attachment point $1$ to the current attachment point ($5$) of \texttt{frag\_1}.
\end{itemize}
This simple example does not cover all features of the GS encoding. 
For more details, we refer the reader to the original GS publication \cite{cheng2023groupselfies}.

\subsubsection{Genetic Operations and Genetic Search}\label{sec:genetic_operations_search}
For the genetic search, an initial population is formed by sampling candidates from the replay buffer. 
The selection process for the mating pool uses a rank-based sampling method.
Within each generation, pairs of parents are uniformly selected from this mating pool to undergo a crossover operation and produce offspring, followed by a mutation step.
The mutations are chosen uniformly from the set of available mutation operations described below. 
If crossover is not applied to a pair, mutation is still performed to maintain population diversity. 
The (mutated) offspring are added back into both the population for the next generation and the replay buffer.
This procedure is repeated for a predefined number of generations.

\begin{figure}[h]
    \centering
    \includegraphics[width=0.35\textwidth]{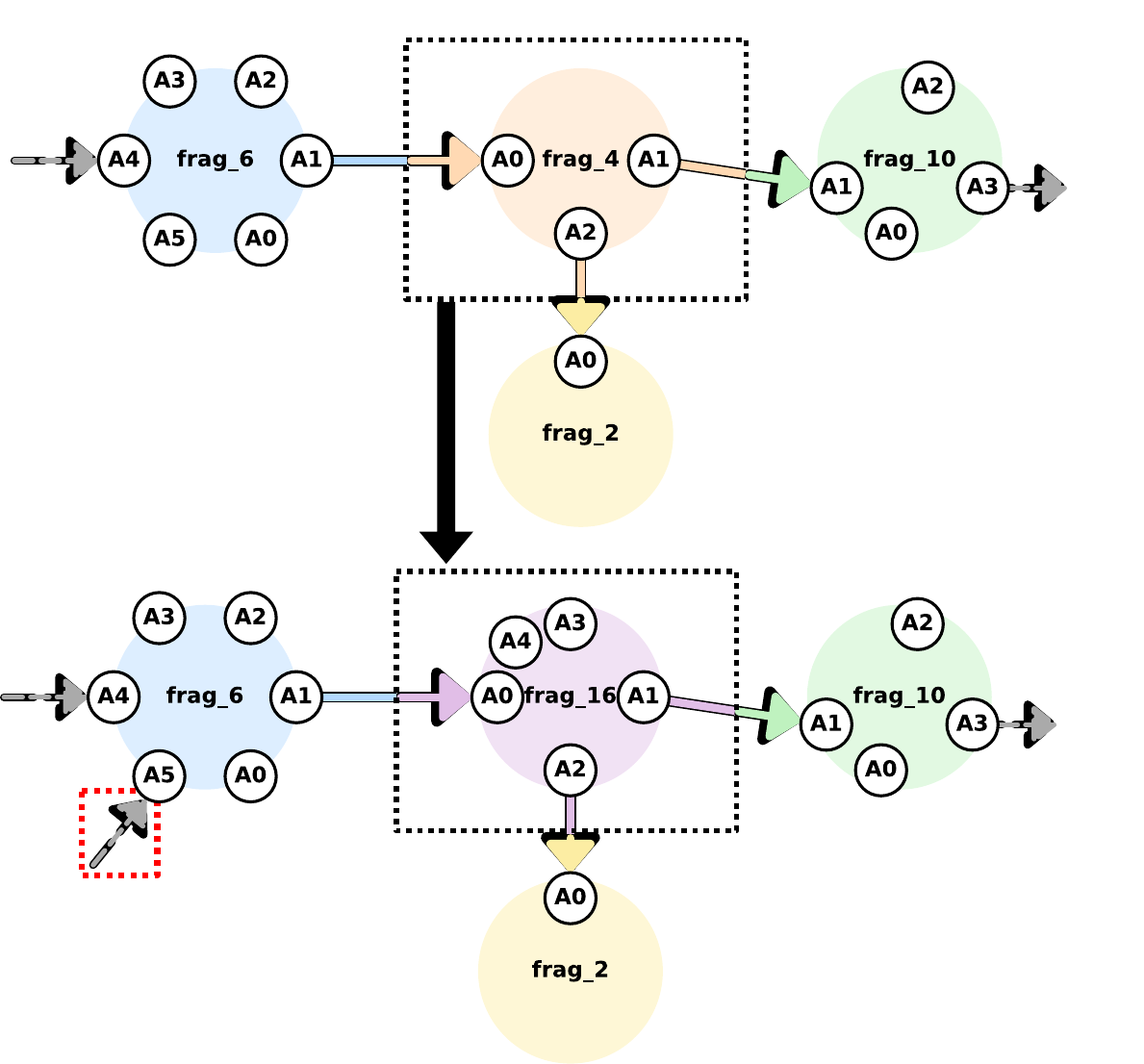}
    \caption{Illustration of mutation operations on the GGS graph structure. Each node represents a molecular fragment and possesses a specific number of attachment points (white circles). Arrows indicate the directed edges of the GGS graph, connecting available attachment points between fragments. In the top panel, the fragment in the dashed box is targeted for a group mutation. A new group is selected from the pool of fragments possessing a sufficient number of attachment points. All incoming and outgoing edges are reconnected to the new group. The bottom panel also shows the anchor position mutation with the dashed red box, where the coupling point of the incoming bond in the source node is changed. This can also be analogously applied to the sink node.}
    \label{fig:mutation}
\end{figure}

The following mutation operations are implemented:
\begin{itemize}

\item \textbf{Group Mutation (1):} Replaces a randomly selected fragment within the graph with a new fragment from the grammar. The operator ensures that the replacement possesses a sufficient number of attachment points to satisfy requirements of existing connections. The process is depicted in \cref{fig:mutation}.

\item \textbf{Bond Mutation (2):} Reconfigures the connectivity between fragments by shifting an existing bond to a different available attachment point. Changing the connectivity is critical for modulating quantum interference effects \cite{reznikova2021substitution, stefani2018large}.

\item \textbf{Anchor Position Mutation (3):} Modifies the specific attachment point used for the electrode interface at either the source (left) or sink (right) fragment. This directly changes the molecule-electrode coupling geometry, which is crucial for transport properties \cite{reznikova2021substitution}. This is also illustrated in the bottom panel of \cref{fig:mutation}. In the figure, the incoming bond to the source node is shifted from attachment point \texttt{A4} to attachment point \texttt{A5}.

\item \textbf{Insert Group Mutation (4):} Breaks an existing internal bond between two fragments and inserts a new fragment in between, effectively expanding the length of the molecular chain.

\item \textbf{Insert Start/End Group Mutation (5):} Appends a new fragment to the molecular termini. This can occur by prepending a new source fragment before the current start node or appending a new sink fragment after the current end node. Adding terminal groups can significantly influence transport properties \cite{blaschke2025revealing,blaschke2023designing}.

\item \textbf{Truncate Mutation (6):} Removes a randomly selected fragment from the graph, provided its deletion does not violate structural integrity and any disconnected parts after deletion can be properly reconnected. Eligible fragments include side branches, regular internal nodes, or terminal fragments (source or sink).

\item \textbf{Anchor Group Mutation (7):} Changes the fragment used at either the source or sink node to a different fragment that can satisfy the attachment point requirements. This mutation allows for exploration of different electrode coupling chemistries.

\item \textbf{Insert Branch Mutation (8):} Adds a new side-chain fragment to a randomly selected node that possesses at least one free, unoccupied attachment point. Side groups can have decisive influence on transport properties and optical properties \cite{yan2024substituents,klockner2017tuning,koczor2025generative}.
    
\end{itemize}

\begin{figure}[h]
    \centering
    \includegraphics[width=0.6\textwidth]{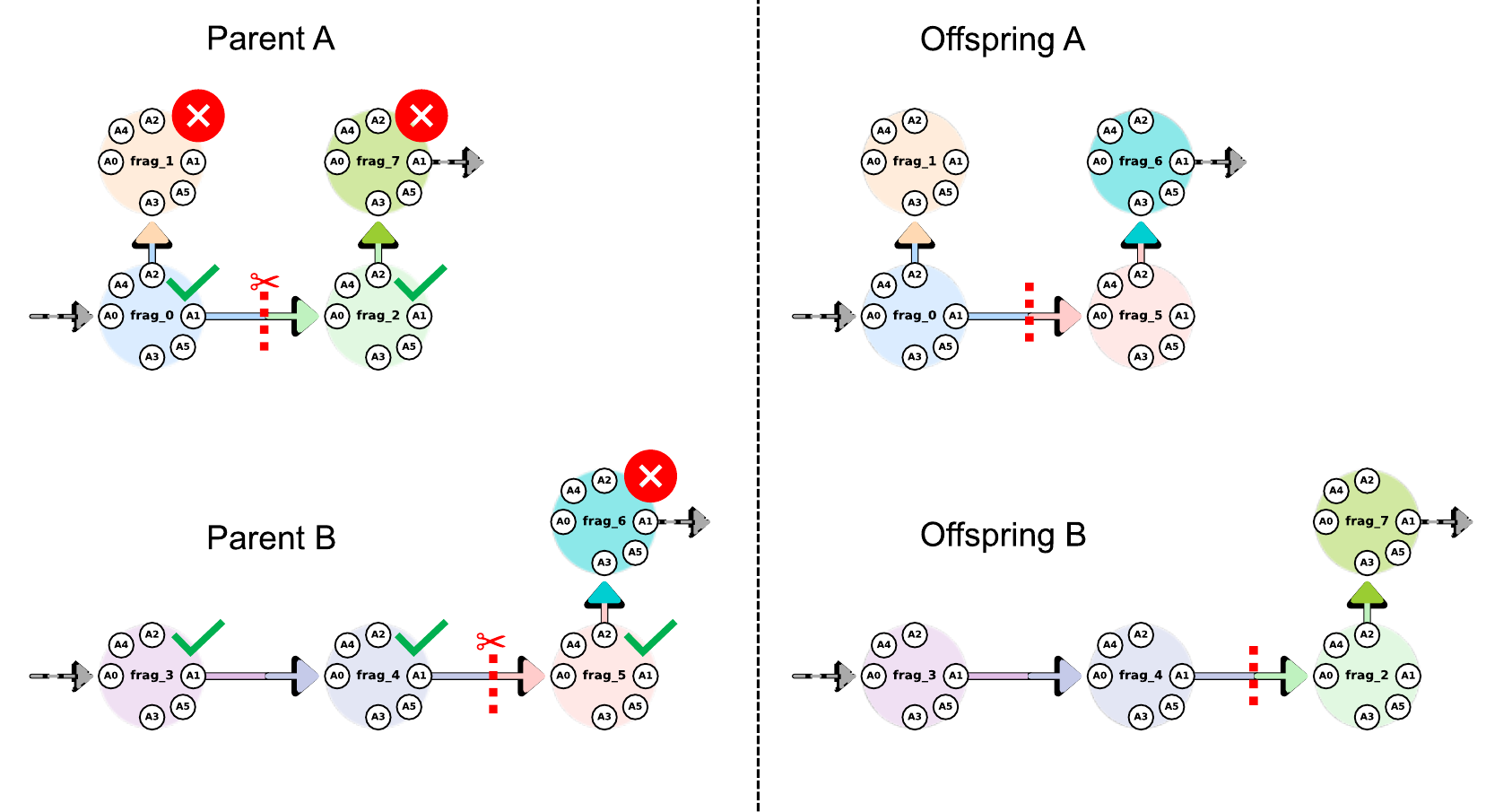}
    \caption{Single-Point Crossover. Two parent DAGs (left) are partitioned at randomly selected edges originating from allowed nodes (green checkmarks). Head and tail subgraphs are exchanged to produce two offspring (right).}
    \label{fig:crossover}
\end{figure}

The crossover procedure is depicted in \cref{fig:crossover}. 
Two parents are chosen to produce two offspring by exchanging head and tail subgraphs at randomly selected crossover points. 
These crossover points are chosen from the set of allowed nodes. 
Allowed nodes include all nodes that are not the sink node or a side-branch node. These categories are indicated by the red crosses and green checkmarks shown in \cref{fig:crossover}.

\subsubsection{Selection of Building Blocks}
\label{sec:building_blocks_selection}
\begin{figure}[h]
    \centering
    \includegraphics[width=0.95\textwidth]{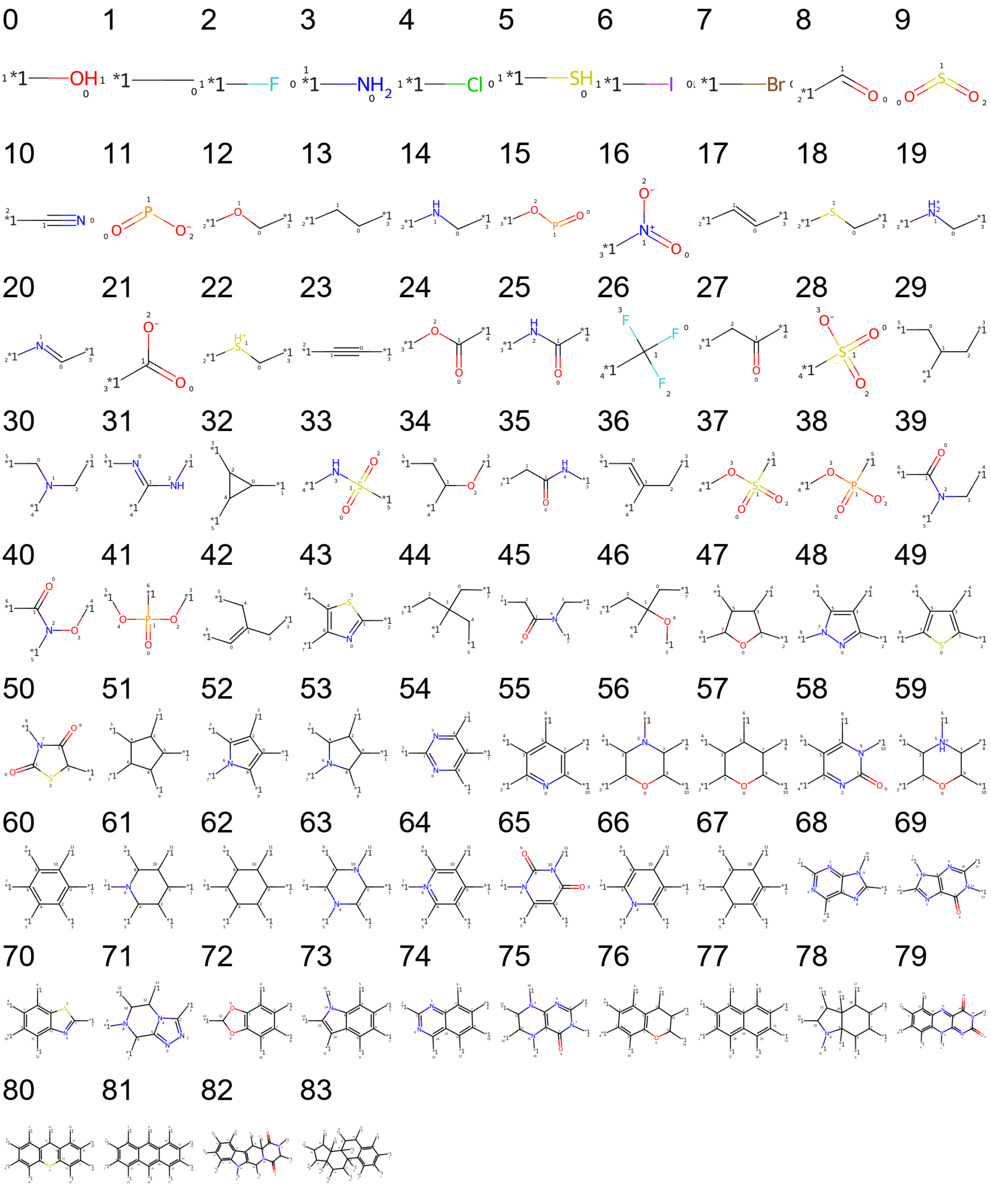}
    \caption{Building blocks, sorted by atom count, used for the construction of molecular candidates. Available attachment points are marked with an asterisk (*). Note that for physics-based tasks, fragments containing sulfur are intentionally excluded.}
    \label{fig:building_blocks}
\end{figure}

The fragments used for the GGS encoding are shown in ~\cref{fig:building_blocks}. 
The selection is built upon the fragments used in previous work \cite{bengio2021gflow}. 
In addition, we added fragments to ensure that every oracle in the PMO benchmark can, in principle, be fulfilled, as a fair comparison with other methods would not be possible if the benchmark targets are not in the search space. The blocks added for this purpose are:
\begin{itemize}
    \item 70: \texttt{scaffold\_hop}
    \item 74: \texttt{deco\_hop}
    \item 72, 82: \texttt{median2}
    \item 78: \texttt{perindopril\_mpo}
    \item 50, 76: \texttt{troglitazone\_rediscovery}
    \item 71: \texttt{sitagliptin\_mpo}
    \item 80: \texttt{thiothixene\_rediscovery}
    \item 83: \texttt{mestranol\_similarity}
\end{itemize}
We also added halogens (blocks 2, 4, 6, 7) and for example anthracene (block 81) and acetylene (block 23) as these blocks are commonly used in molecular phononics \cite{klockner2017tuning,blaschke2025revealing}.
For all physics-based tasks, we excluded all fragments containing sulfur, as this element is reserved for the anchor groups due to its favorable binding properties to gold electrodes \cite{rubio2001mechanical, frisenda2015electrical}. 
Additional sulfur in the molecular backbone would lead to uncontrollable alternative binding sites to the gold electrodes.

The search space within our setting is huge, as the number of possible combinations grows exponentially with molecular length. 
The base of this exponential growth is determined by the number of available fragments and their diverse bonding configurations. 
The size of the resulting search space is estimated to be $\gg 10^{30}$ \cite{blaschke2025revealing}.

The choice of building blocks naturally limits the types of molecules that can be generated, though this enables more focused exploration of chemically relevant regions.
This is exemplified in the final candidates shown in Figure \ref{fig:top_molecules}, where the best-performing molecules identified by the thermoelectric and phonon oracles contain the specially added acetylene fragment. 
In contrast to fragmentation of large datasets, selecting building blocks based on domain knowledge allows the search to be focused on chemically relevant motifs. 
The selection of building blocks could be further coordinated with experimental collaborators to ensure that specific laboratory insights and synthetic accessibility constraints are incorporated into the molecular design space.

\subsection{Creation of Synthetic Datasets}
\label{sec:synthetic_datasets}

The synthetic pretraining dataset is generated through a pipeline consisting of the following stages:

\begin{itemize}
    \item \textbf{Stochastic Graph Assembly:} The generator requires two primary constraints: a maximum number of molecular fragments and a maximum number of \texttt{[pop]} operations. 
    Molecules are constructed stochastically as directed acyclic graphs. 
    Starting from a seed fragment, at each step during the construction, an allowed operation is chosen randomly from a pre-loaded grammar that categorizes building blocks based on their available attachment points. 
    The algorithm iteratively adds fragments and edges, strictly tracking available attachment sites and managing molecular branching. 
    Once the graph is complete, it is translated directly into a valid GGS string encoding (see \cref{sec:ggs_details}).
    
    \item \textbf{Controllable Bias and the ``Chemist's Shop'':} For applications beyond the NMO benchmark, the fragment vocabulary can be chosen freely. 
    This provides a robust method to avoid the implicit, hidden biases found in massive pharmaceutical datasets. 
    While any curated list of fragments naturally introduces some bias, this bias is explicit and completely controllable. 
    For instance, the fragment selection can be tailored in direct consultation with experimental partners to match the specific synthesis capabilities and available reagents of their laboratory, effectively acting as a custom ``chemist's shop'' for the generative model.
    
    \item \textbf{Optional Chemical Stability Filtering:} While the GGS encoding guarantees syntactical validity, the molecular filters can be used to ensure the generated molecules represent chemically stable and plausible structures \cref{sec:filters}. 
    
    \item \textbf{Dual-Format Output for Data-Scarce Regimes:} To support a wide variety of baseline models and frameworks, the pipeline natively generates datasets in the GGS encoding and features an optional translation to the SMILES format. 
    This capability to procedurally generate massive, valid datasets from scratch in both SMILES and GGS is highly critical for training generative models in novel physical domains where data is completely unavailable.
\end{itemize}

\subsection{Code and Data Availability}
\label{sec:code_and_data_avail}
The code is available at \url{https://github.com/blaschma/TheNanotechnologyMolecularOptimizationBenchmark} under an MPL-2.0 license. 
The codebase includes the implementation of the benchmark, the GGS encoding, the baseline models, and all scripts for training and evaluation.
All parts are fully documented, contain extensive READMEs, and are designed for easy use and extension by the community.
Models adapted to work with NMO (and GGS) are also included, with linking the correct license for the respective models.
Submissions to the benchmark will be accepted through pull requests to the GitHub repository, and we will provide clear guidelines for how to submit new models and results.
We provide the relevant molecules found using our baseline method for the nanotechnology community and ML researchers under \url{https://huggingface.co/datasets/blaschma/NMO_Baseline_Relevant_Candidates} (CC BY 4.0 license).

\subsection{Solving NMO: Key Findings}
\label{sec:solving_NMO}
In this section we summarize our findings on how to develop a model capable of solving the NMO benchmark.
The key points are:
\begin{itemize}
    \item \textbf{Anchor modeling is decisive.} The physical properties of NMO tasks are defined by the full molecule together with its electrode-binding anchors. The anchor positions are not auxiliary metadata, they are a decisive factor for the resulting properties and should be part of the learning process. We provide GGS as one native solution, but anchor positions can also be explicitly defined in a SMILES setting, and alternative approaches are equally welcome.
    
    \item \textbf{Chemical space.} The design of the chemical space has influence on optimization performance. We provide a curated set of building blocks. The effect of removing critical fragments such as acetylene is shown in \cref{tab:auc_table_fragment_ablation}. At the same time, the MO task benefits from a pharmaceutical bias, while TE and PH do not, indicating that no single prior fits all tasks.
    
    \item \textbf{Model robustness over per-task tuning.} The NMO tasks are computationally expensive, and classical hyperparameter search across tasks is infeasible. The benchmark protocol enforces this discipline. A single configuration must perform well on three physically distinct tasks. Methods should therefore be designed for robustness rather than fine-tuning.
    
    \item \textbf{Rugged fitness landscapes.} The NMO fitness landscapes, especially for TE, are highly rugged. This can destabilize generative models through catastrophic forgetting or mode collapse. As shown in \cref{tab:auc_table_suppl}, our dynamic stabilization mechanisms (DCD and DEX) address this without requiring KL anchoring to a prior.
    
    \item \textbf{Use the provided filter set.} The benchmark ships with a set of cheap heuristic filters that exclude known unstable motifs. These filters can be applied before fitness evaluation and do not count towards the oracle budget, increasing optimization efficiency.
    
    \item \textbf{Pitfalls.} For PH, arbitrarily long molecules trivially achieve arbitrarily low thermal conductance. Restrict the molecular length to a reasonable range, e.g.\ comparable to the top-performing candidate in \cref{fig:top_molecules}(b). In practice, this regime is self-limiting: the xTB calculations underlying the phonon oracle scale superlinearly with system size, so generating very long molecules rapidly inflates the per-evaluation cost. For MO, candidates with $P > 15$ should be treated with caution, as $P$ is likely overestimated in this regime.
\end{itemize}

\subsection{Impact Statement}
\label{sec:impact_statement}
Since our work targets molecular design without dataset bias, it has the potential to impact a wide range of scientific domains, including nanotechnology, materials science, physics, chemistry, and life sciences.
Our framework broadens access to molecular engineering by enabling the use of a custom fragment-based vocabulary that can be adapted to the specific needs and constraints of different research areas and laboratories.
We note that while our method efficiently optimizes physical properties within the benchmarks scope, final candidates should be validated before deployment through higher-level theory and experiment.
Moreover, the broader impact of the molecules has to be carefully evaluated on a case-by-case basis by domain experts.
Especially features like toxicity and environmental impact need to be considered when deploying molecules in real-world scenarios.

\section{Our Baseline Method: Genetic GFN framework} 
\label{sec:gentic_gfn_framework}
\begin{figure}[h]
    \centering
    \includegraphics[width=0.95\textwidth]{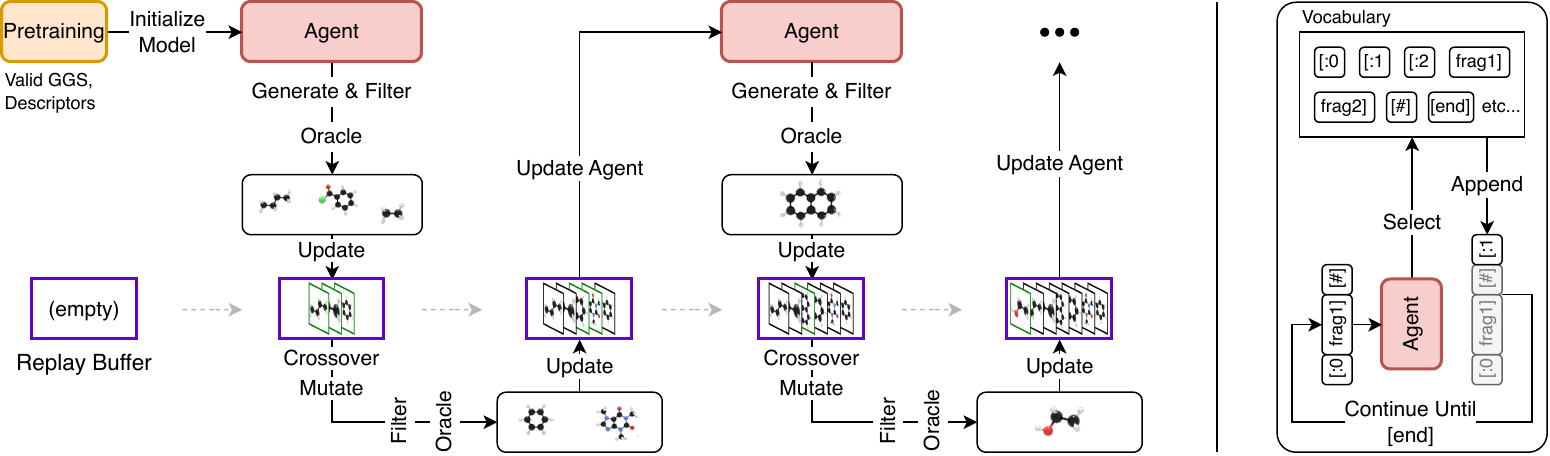}
    \caption{
        \textbf{Overview of the framework.} 
        First, the agent is initialized with the pretraining weights. 
        During optimization, the model first generates a batch of molecules, which are filtered and evaluated by the oracle. 
        High-reward molecules are stored in a replay buffer, from which a GA proposes refined candidates, which are added to the buffer if they are sufficiently good. 
        Finally, a training step is performed on the agent using the molecules from the buffer.
        The right panel illustrates the iterative generation process of a molecule by the agent.
    }
    \label{fig:overview}
    \vspace{-0.4cm}
\end{figure}
We extend Genetic GFNs \citep{kim2024geneticgflow} with novel techniques for dataset-free optimization, creating a domain-agnostic framework for both the NMO and the established PMO benchmark (see \cref{sec:PMO_benchmark}).
An overview of our full pipeline is shown in \cref{fig:overview}.
Specifically, we replace the standard SMILES molecular representation with GGS encoding and substitute the pharmaceutical pretraining dataset with a procedurally generated random pretraining dataset.

\subsection{Principles of the Genetic GFN Framework}
\label{sec:principles_GFN}
Genetic GFNs combine the sampling capability of Generative Flow Networks (GFNs) \cite{bengio2021gflow,bengio2023gflownet} with the exploration efficiency of GAs.
The objective is to learn a policy $P_{\theta}(x)$ that samples molecules $x$ proportional to their reward, $P_{\theta}(x) \propto R(x)$, creating a diverse portfolio of high-performing candidates rather than a single optimum.
Formally, the generation process is modeled as a sequential decision process.
An agent constructs a molecule of length $T$ by iteratively selecting actions from the action space (see \cref{sec:action_space}).
A selected action transitions the current molecular state $s_t$ at step $t$ to the next state $s_{t+1}$ until the terminal state is reached.
The training process proceeds in two phases: \\[0.33ex]
\textbf{Phase 1: Pretraining.} 
First, a prior model $\theta_\text{prior}$ is trained to learn the syntax of chemically valid molecules. 
Minimizing the negative log-likelihood (NLL) of fragment sequences from a pretraining dataset consisting of valid molecules effectively establishes a "validity prior":
{\small
\begin{equation}
    \label{eq:pretraining}
    \mathcal{L}_\text{Prior} = - \sum_{t=0}^{T-1} \log P_{\theta_\text{prior}}(s_{t+1}|s_t).
\end{equation}
}
\textbf{Phase 2: Task Optimization.}
The agent is initialized with the weights of the prior $\theta_\text{prior}$. The training loop then iterates through three steps:
\begin{enumerate}[topsep=0pt,wide=0pt,itemsep=0pt,parsep=0pt]
    \item \textsc{Sampling:} The agent generates a batch of candidate molecules, which are evaluated by the oracle.
    \item \textsc{Genetic Refinement:} High-reward candidates are stored in a replay buffer. A GA performs crossover and mutation on these candidates to discover higher-scoring offspring. These refined samples are also evaluated by the oracle and added to the buffer.
    \item \textsc{Policy Update:} The GFN is trained on high-reward trajectories from the buffer using a simplified Trajectory Balance loss \cite{kim2024geneticgflow}
    {\small
    \begin{equation}
        \mathcal{L}_\text{TB} = \left( \log Z_\mathrm{p} + \sum\limits_{t=0}^{T-1} \log P_{\theta_\text{agent}}(s_{t+1}|s_t) - \log R(x) \right)^2
    \label{eq:tb_loss}
    \end{equation}
    }
    with partition function $Z_\mathrm{p}$, trajectory length $T$, and reward $R(x) = \exp(\beta f(x))$ derived from the fitness $f(x)$.
    The original implementation adds a Kullback-Leibler divergence (KL) regularization term to anchor the agent to a prior.
    We remove the KL term because our prior is initialized from a procedurally generated random dataset, teaching only chemical syntax.
    This allows the policy to be driven exclusively by the physical oracle.
\end{enumerate}

\subsection{Pretraining Without Dataset Bias}
A core limitation of current generative models is their reliance on large-scale datasets for pretraining. 
This imprints a strong distributional bias, restricting the agent's ability to explore novel functional domains.
To overcome this, we introduce a pretraining strategy that relies solely on the constructive logic of our GGS encoding, eliminating the need for any domain-specific prior dataset. \\[0.33ex]
\textbf{Synthetic Pretraining Dataset:} \,
Instead of learning from existing data, we generate a synthetic pretraining dataset $\mathcal{D}_{\text{rnd}}$ by randomly sampling GGS graphs.
Formally, we create a dataset of $N=\num{300000}$ random molecules $\mathcal{D}_{\text{rnd}} = \{x_1, \dots, x_N\}$.  
Each GGS graph $x_i$ is generated through the stochastic assembly of molecular building blocks. 
Starting from a seed fragment, we iteratively append blocks, couplings, or \texttt{[pop]} tokens until the \texttt{[end]} token is sampled or a maximum token limit is reached, while tracking valence and attachment points to ensure chemical validity.
The final graph is then translated into a sequence of action space indices (see \cref{sec:action_space}). \\[0.33ex]
We train the model by minimizing $\mathcal{L}_\text{Prior}$ from \cref{eq:pretraining}, effectively teaching the agent the fundamental \textit{syntax} of the molecular representation.
The network learns which action sequences yield valid chemical encodings without inducing any bias from historical datasets. \\[0.33ex]
\textbf{Molecular Filters:} \,
While GGS guarantees validity, it does not ensure chemical stability.
To prevent the agent from learning to generate unstable or explosive structures (e.g., polyynes), we apply a set of lightweight heuristic filters during the dataset generation and training.
While filtering molecules to eliminate false hits is standard practice in drug discovery \cite{kralj2023molecular}, established filters from that domain are inapplicable in our context, as they tend to discard functional units that are critical for our applications.
Therefore, we develop a custom set of inexpensive filters that are explained in more detail in \cref{sec:filters}.
These filters ground the agent's prior in a chemically plausible space without consuming the oracle budget.

\subsection{Injecting Chemical Intuition via Descriptors}
\label{sec:descriptors_method}
To enhance the model's chemical understanding beyond mere syntax, we augment the GFN architecture with an auxiliary non-interfering regression head that predicts $K=17$ cheaply computable molecular descriptors $\left\{ \mathbf{y}_{\text{d},k} \in \mathbb{R} \right\}_{k=1}^K$ (see \cref{sec:descriptors}) directly from the latent representation.
We define the descriptor loss $\mathcal{L}_\text{desc}$ as the mean squared error between the predicted normalized descriptors $\hat{y}$ and the ground-truth descriptors $y$, which are standardized by the statistics $(\mu, \sigma)$ of the random pretraining dataset:\vspace{-1ex}
{\small
\begin{equation}
    \label{eq:descriptorloss}
    \mathcal{L}_\text{desc} = \frac{1}{K} \sum_{k=1}^K \left( \hat{y}_{\text{d},k} - \frac{y_{\text{d},k} - \mu_k}{\sigma_k} \right)^2.
\end{equation}
}
\noindent We integrate this auxiliary loss into both training phases:\vspace{-0.66ex}
\begin{itemize}[leftmargin=*,itemsep=0pt,parsep=0pt,topsep=0pt,partopsep=0pt]
    \item \textbf{Phase 1 (Pretraining):} We minimize $\frac{1}{2}\cdot(\mathcal{L}_{\text{Prior}} + \mathcal{L}_{\text{desc}})$, enabling the model to get a foundational structure-property understanding from the randomly generated molecules before encountering a physics task.
    \item \textbf{Phase 2 (Optimization):} We minimize $\mathcal{L}_{\text{TB}} + 0.1 \cdot \mathcal{L}_{\text{desc}}$. Retaining this loss acts as a regularizer, preventing the agent's latent space from collapsing and forgetting core chemical concepts.
\end{itemize}

\subsection{Molecular Descriptors}
\label{sec:descriptors}

All descriptors are computed for the fully constructed molecule using RDKit, requiring the agent to learn which fragment combinations lead to which descriptor values:
\begin{enumerate}
    \item \verb|tpsa|: Topological polar surface area (TPSA) of a molecule.
    \item \verb|bertzCT| : Topological index meant to quantify “complexity” of molecules \cite{bertz1981first}.
    \item \verb|num_rings|: Number of rings for a molecule.
    \item \verb|aromatic_rings|: Number of aromatic rings for a molecule.
    \item \verb|mol_wt|: Exact molecular weight for a molecule.
    \item \verb|hba|: Number of Lipinski H-bond acceptors in a molecule.
    \item \verb|hbd|: Number of Lipinski H-bond donors in a molecule.
    \item \verb|rot_bonds|: Number of rotatable bonds for a molecule.
    \item \verb|fsp3|: Fraction of C atoms that are $\text{sp}^3$ hybridized.
    \item \verb|heavy_atoms|: Number of heavy atoms for a molecule.
    \item \verb|heteroatoms|:  Number of heteroatoms for a molecule.
    \item \verb|aliph_carbocycles|: Number of aliphatic (containing at least one non-aromatic bond) carbocycles for a molecule.
    \item \verb|aliph_heterocycles|: Number of aliphatic (containing at least one non-aromatic bond) heterocycles for a molecule.
    \item \verb|aromatic_carbocycles|:  Number of aromatic carbocycles for a molecule.
    \item \verb|aromatic_heterocycles|: Number of aromatic heterocycles for a molecule.
    \item \verb|crippen_logp|: Wildman-Crippen LogP value \cite{wildman1999prediction}. 
    \item \verb|crippen_mr|: Wildman-Crippen MR value \cite{wildman1999prediction}.
\end{enumerate}
The descriptors are stored in a vector containing the numerical values in the order listed above.
The numerical values of different descriptors are on vastly different scales, with values ranging from small fractions (e.g. \verb|fsp3|) to several hundreds (e.g. \verb|bertzCT|). 
To prevent features with larger magnitudes from disproportionately influencing the model's gradients, we apply a normalization.
Each descriptor is individually scaled to zero mean and unit variance, with the underlying statistics computed across the generated pretraining dataset (see Equation \eqref{eq:descriptorloss}).
During the training, the auxiliary non-interfering regression head of the transformer model predicts these normalized values $\hat{y}_{\text{d},k}$.

\subsection{Adaptive Stability Mechanisms}
\label{sec:adaptive_stability}
Training GFNs on rough energy landscapes can be notoriously unstable \cite{kim2023learning,lau2024qgfn,deleu2025relative,fan2025adaptive}.
When solving the NMO benchmark, sometimes outliers are discovered that are significantly better than the current policy's distribution.
Attempting to maximize the likelihood of these outliers too aggressively causes the agent to become unstable, leading to two distinct failure modes: \textit{Catastrophic Forgetting} (unlearning the syntax rules and generating sequences that fail to construct a GGS graph) and \textit{Mode Collapse} (high duplicate rate).
As described in \cref{sec:principles_GFN}, the original Genetic GFN uses a KL penalty to mitigate this problem.
However, we omit this term as we solely rely on a randomly generated pretraining dataset.
Instead, we introduce adaptive strategies that monitor the agent's sampling statistics:\vspace{-1ex}
\begin{itemize}[leftmargin=*,itemsep=0pt,parsep=0pt,topsep=0pt,partopsep=0pt]
    \item \textbf{Dynamic Cooldown (DCD):} 
    If the rate of invalid sequences $r_\mathrm{i}$ rises above 35\%, we infer that the agent's weights are eroding (Catastrophic Forgetting) due to high-variance gradients from high reward outliers.
    For the training of the agent, we then temporarily sample from the replay buffer using \textit{uniform} instead of \textit{rank-based} sampling and quadruple the batch size.
    This stabilizes the gradients by re-exposing the agent to a broader set of valid molecules, allowing it to recover the syntax rules.
    \item \textbf{Dynamic Exploration (DEX):}
    If the rate of unique molecules $r_\mathrm{u}$ per generated batch drops below 30\%, the agent is suffering from mode collapse.
    We respond by increasing the rank-sampling coefficient, which flattens the sampling distribution and increases the diversity of sampled molecules.
\end{itemize}
\vspace{-1.33ex}Once the validity and uniqueness rates return to normal, the hyperparameters are reset and normal training resumes.

\subsection{Full Pipeline}
Our full pipeline combines all the components described above into a coherent framework and an overview is given in \cref{fig:overview}.
First, we pretrain the model on our procedurally generated synthetic dataset filtered for chemical stability.
Next, we initialize the agent with the pretraining weights and enter the optimization phase.
In each iteration, the agent generates a batch of molecules, which are first filtered according to \cref{sec:benchmark_protocol} and then evaluated by the oracle.
High-reward molecules are stored in a replay buffer.
From these, a GA proposes refined candidates, which are also filtered, then evaluated by the oracle, and added to the buffer.
Using the molecules from the replay buffer, we perform a training step on the agent, combining the trajectory balance loss $\mathcal{L}_\text{TB}$ with the descriptor loss $\mathcal{L}_{\text{desc}}$.
This loop continues until the oracle budget is exhausted. \\
In summary, our key modifications are:\vspace{-1ex}
\begin{itemize}[leftmargin=*,itemsep=0pt,parsep=0pt,topsep=0pt,partopsep=0pt]
    \item \textbf{Unbiased Pretraining:} We replace the pharmaceutical pretraining dataset with a synthetic dataset that we screen with lightweight heuristic filters. This eliminates domain bias, while still enabling the model to learn the syntax of valid molecules. 
    \item \textbf{GGS Representation:} We substitute SMILES with Graph Group SELFIES. By restricting the vocabulary to standard fragments, we guarantee that generated molecules are easily synthesizable. The encoding naturally models application-specific constraints like molecule-electrode binding and enables domain-specific genetic operations.
    \item \textbf{Architecture \& Stability:} The original network architecture is a GRU \cite{cho2014gru}, which we replace with a Transformer \cite{vaswani2017attention}. To stabilize training in the rugged energy landscapes of physical oracles, we introduce stability mechanisms (DCD and DEX) and remove the KL. Finally, we introduce an auxiliary descriptor prediction task to inject basic chemical knowledge.
\end{itemize}

\subsection{Methodological Details: Action Space}
\label{sec:action_space}
The agent iteratively samples actions from the action space to construct a molecular candidate.
Each action is associated with an index, and at each step $t$, the agent samples an index from the action space. 
Fundamentally, every decision process begins with a start state and terminates with the \texttt{[end]} token. 
The action space differs between SMILES and GGS encodings.

In the SMILES representation, the action space consists of all tokens within the SMILES vocabulary. 
This includes atomic symbols (e.g., C, O, N, Cl, Br), bond symbols (e.g., =, \#), and ring-closure symbols (e.g., 1, 2, 3, ..., 9). 
The vocabulary is fixed and remains unchanged, thereby defining the chemical space from the outset. 
Assuming an exemplary action space with three actions and the corresponding index-to-action mapping
\(0 \rightarrow \mathrm{C}\), \(1 \rightarrow =\), and \(2 \rightarrow \texttt{[end]}\),
the SMILES string \texttt{C=C} would be constructed by sampling the action
sequence \([0, 1, 0, 2]\).

For the SMILES studies, we primarily adopt the vocabulary established in \cite{kim2024geneticgflow}.
In this case, the action space comprises 53 distinct tokens. 
In cases where the random dataset is translated into SMILES (see Row 2 in \cref{tab:auc_table} and column 2 in \cref{tab:pmo_results}), tokens not present in the vocabulary needed to encode molecules from the dataset are added. 
During pretraining, the agent must learn to construct valid SMILES strings and, specifically, to assemble chemically valid molecules.

Conversely, when utilizing the GGS encoding, the action space comprises three parts (see \cref{sec:ggs}): fragments $\mathrm{frag}_n$, coupling-in $S_\mathrm{in}$, and coupling-out $S_\mathrm{out}$ (see Table \ref{tab:token_index_mapping}), complemented by \texttt{pop} and \texttt{end} tokens (see \cref{fig:encoding_overview}). 
The action space is automatically generated based on the selection of fragments. 
The number of coupling-in and coupling-out options is set to the maximum number of attachment points in the fragment library.
For the full selection of fragments in \cref{fig:building_blocks}, the action space consists of 109 actions.
Leaving out every fragment containing sulfur (as done for physics-based tasks) reduces the action space to~91 actions.
During pretraining in the GGS case, the agent learns to construct valid GGS strings. 
In contrast to the SMILES approach, the agent does not need to learn chemical validity rules, as a valid GGS string automatically generates chemically valid molecules by construction.

\subsection{Extended Evaluation}
\label{sec:extended_eval}
In this section, we provide additional experimental details on the developed machine learning techniques presented in this work.
We discuss the pretraining on the synthetic dataset, analyze the interplay between agent sampling and genetic search, and discuss the effect of our stability mechanisms.
\subsubsection{Pretraining}
\label{sec:pretraining_ana}
During the pretraining phase, the agent must learn the syntax of valid molecules. 
For the SMILES encoding used in the original Genetic GFN method, this implicitly includes learning the underlying chemistry, which is a challenging task. 
In contrast, the GGS encoding employed in our method is designed to represent only valid molecules. 
Thus, the agent need only learn the syntax of GGS strings that can be translated into valid GGS graphs, which is considerably easier.

\begin{figure}
    \centering
    \includegraphics[width=0.75\textwidth]{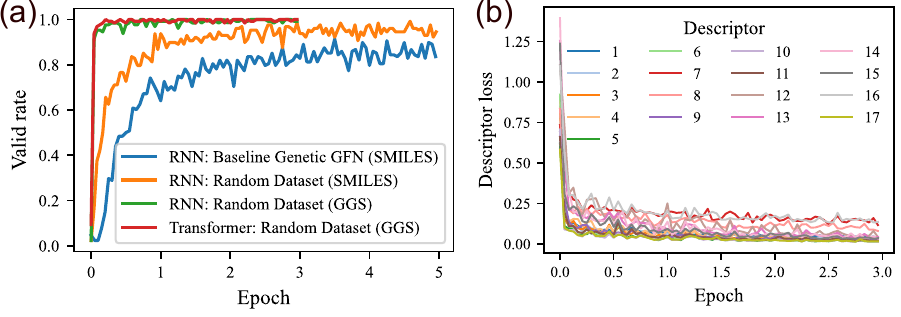}
    \caption{(a) Valid rate of the models and employed encoding according to the legend versus pretraining epochs. (b) Normalized descriptor loss of the predicted descriptors for the transformer model during pretraining. The numbering of the descriptors corresponds to section \ref{sec:descriptors}.}
    \label{fig:pretraining_ana}
\end{figure}

The valid rate during pretraining, defined as the proportion of sampled candidates that correspond to valid molecules, is shown in Figure \ref{fig:pretraining_ana}(a). 
The original Genetic GFN with an RNN model (more specifically, multiple GRU cells) and SMILES encoding achieves the lowest valid rate after 5 epochs of pretraining. 
Switching to the random dataset increases both the learning speed and valid rate, as the model is repeatedly exposed to similar building blocks present in the dataset which was translated from the GGS random dataset to SMILES.
Switching to GGS encoding with an RNN model significantly improves the valid rate. 
In the first epoch, the valid rate reaches nearly 100\%. 
Consequently, we terminate training after 3 epochs to prevent overfitting. 
Changing to the transformer model further increases the learning speed slightly.

We also incorporated a non-interfering regression head that predicts $K = 17$ cheaply computable molecular descriptors (listed in Section \ref{sec:descriptors}) into the transformer model to provide the model with chemical intuition. 
The normalized descriptor loss during pretraining is shown in Figure \ref{fig:pretraining_ana}(b). 
The loss decreases rapidly in the first pretraining epoch for all descriptors and subsequently saturates with a slightly decreasing trend. 
This indicates that the model learns to predict molecular features and acquires chemical intuition about the available building blocks.

\subsubsection{Sampling and Genetic Algorithm}
Within the Genetic GFN framework, molecules are generated alternately through agent sampling and genetic refinement. 
Here, we demonstrate how this process operates and analyze its impact on the results. 
We present the behavior for all oracles using the best-performing seed for both the original Genetic GFN with SMILES encoding and our method (last entry in Table \ref{tab:auc_table}).

\begin{figure}
    \centering
    \includegraphics[width=0.99\textwidth]{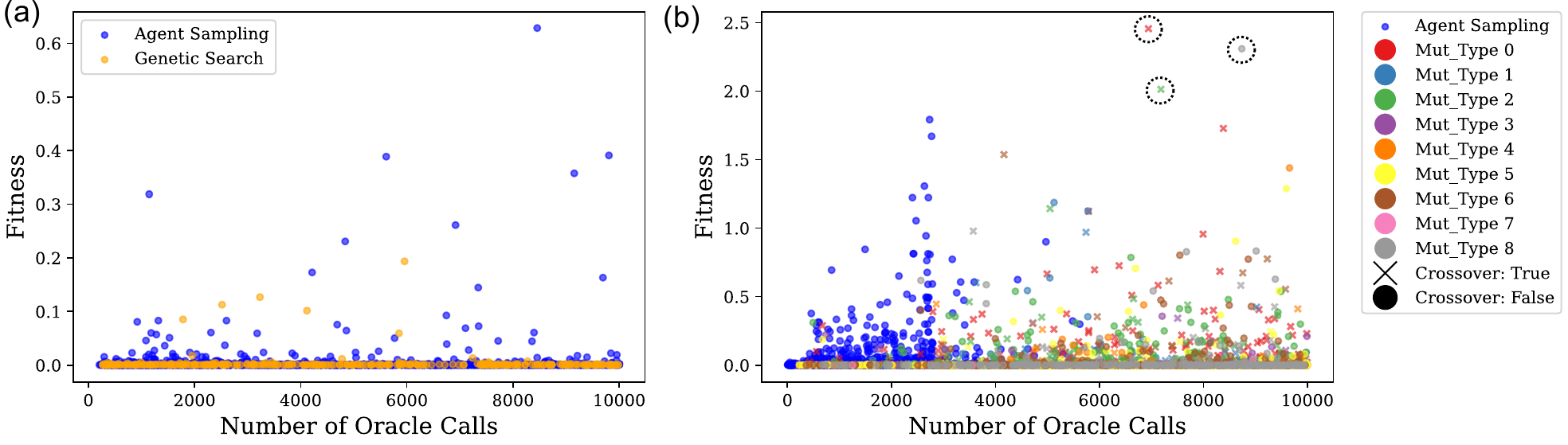}
    \caption{Sampling statistics for the thermoelectric oracle. (a) Fitness plotted during training versus oracle calls. Only fitness values $>0$ are shown. Blue dots correspond to molecules sampled from the agent and orange dots to molecules generated via the genetic algorithm. (b) Same statistics for our method. The statistics of the genetic algorithm are broken down into the specific genetic operations, with mutations listed in Section \ref{sec:genetic_operations_search} shown in different colors. Mutation type 0 indicates no mutation was applied. In both cases, only molecules that are newly discovered are shown. Dashed circles mark candidates discussed in the text.}
    \label{fig:ZT_sampling}
\end{figure}

For the thermoelectric oracle, the sampling statistics are shown in Figure \ref{fig:ZT_sampling}. 
Panel (a) shows that only a fraction of the $10000$ tested molecules can satisfy the hard constraints in Equation \eqref{eq:fitness_struc}. 
An even smaller portion of these molecules achieves a fitness significantly greater than $0$.
The genetic search fails to identify top-performing candidates. 
The GA in the original Genetic GFN operates on SMILES strings, which proves ineffective.
For our method in panel (b), significantly more candidates with fitness $> 0$ are observed. 
Initially, training is driven by agent sampling. 
After approximately 3000 oracle evaluations, the genetic algorithm begins to discover higher-performing molecules. 
Particularly important here are pure crossover (red cross), crossover combined with the bond mutation (green cross), and the pure insert branch mutation (gray point), all of which are marked with dashed circles.
However, this does not indicate that the agent stops learning. 
The sampling statistics in \cref{fig:ZT_sampling} show only newly discovered molecules. 
When the agent samples a molecule encoding that has already been evaluated during training, this is not visible in the statistics, as the molecule is not re-evaluated.

\begin{figure}
    \centering
    \includegraphics[width=0.99\textwidth]{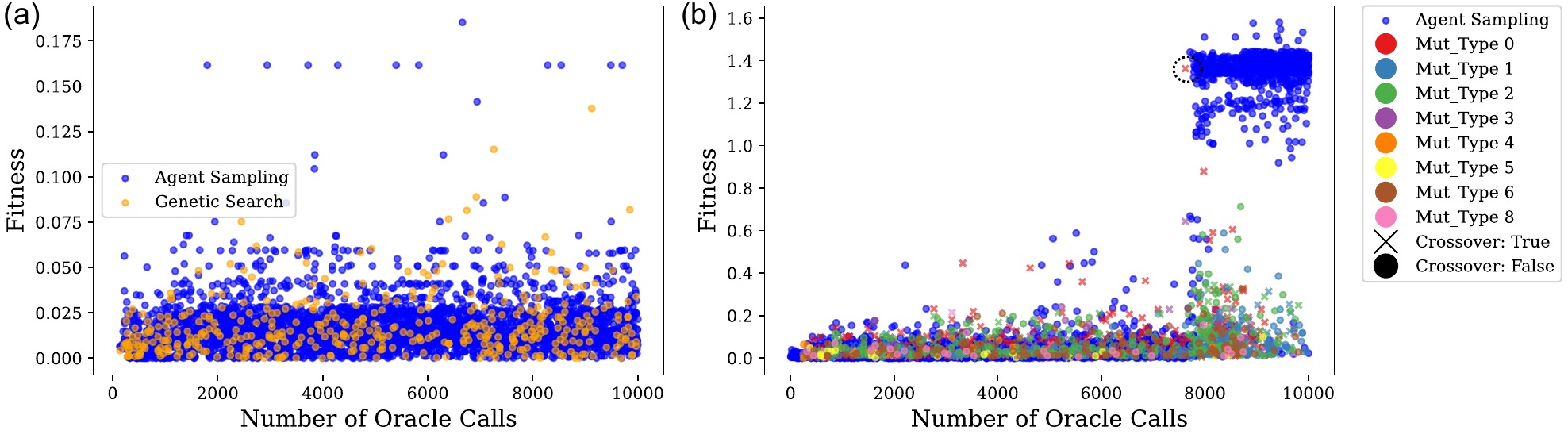}
    \caption{Same as in Figure \ref{fig:ZT_sampling} but for the phonon oracle.}
    \label{fig:phonon_sampling}
\end{figure}

Figure \ref{fig:phonon_sampling} shows the sampling statistics for the phonon transport oracle.
Panel (a) shows similar behavior to the thermoelectric oracle, with only a few candidates satisfying the hard constraints in Equation \eqref{eq:fitness_struc}. 
The genetic search on SMILES strings also proves ineffective in this case. 
For our method in panel (b), numerous candidates with fitness $>0$ are observed.
Initially, high-performing candidates are discovered through agent sampling and the GA equally. 
Around oracle call $8000$, the GA discovers a high-performing candidate through crossover (marked with a dashed circle).
Subsequently, the agent continues to explore this region and discovers even better candidates.

\begin{figure}
    \centering
    \includegraphics[width=0.99\textwidth]{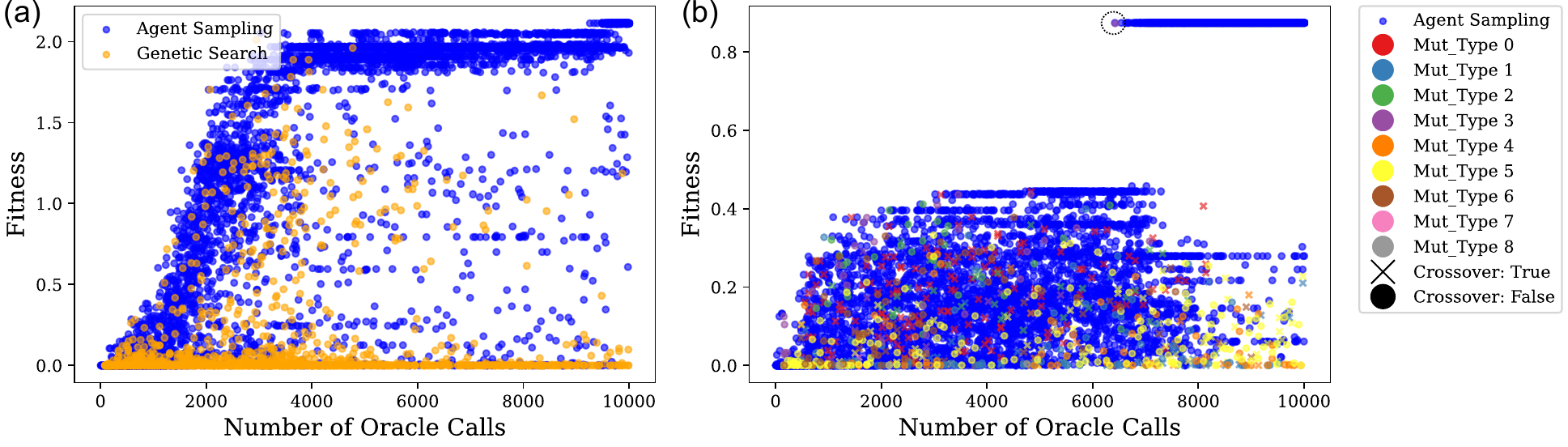}
    \caption{Same as in Figure \ref{fig:ZT_sampling} but for the optomechanical oracle.}
    \label{fig:upconversion_sampling}
\end{figure}

The third oracle in our benchmark is the optomechanical oracle. 
The sampling behavior is shown in Figure~\ref{fig:upconversion_sampling}. 
Panel (a) again shows that in this case a large portion of the sampled molecules have fitness $>0$, as the hard constraints in Equation~\eqref{eq:fitness_struc}are relatively easy to satisfy. 
The string-based genetic search works better here because the replay buffer is filled with a larger number of valid molecules. 
For our method in panel (b), both agent sampling and genetic search are important. 
Here again, our graph-based genetic operators appear to perform better. 
The best candidate is discovered through agent sampling (marked with the dashed circle).
Note that a Mutation type 3 (purple dot) is also found in this high fitness region, but the agent sampling finds it first. 
The fitness appears to remain constant afterwards.
This occurs because GGS strings are not unique, and the agent samples different encodings that lead to the same molecule.

The analysis shows that for all oracles, the hybrid approach combining agent sampling and genetic refinement is key. 
Furthermore, the genetic search in our method performs significantly better. 
Through the graph-based approach and our GGS encoding, the genetic operators always yield valid molecules. 
Note that molecules generated by the GA may still have fitness values of 0 due to the hard constraints in all fitness functions.

\subsubsection{Stability via Adaptive Hyperparameters}
\begin{figure}[th]
    \centering
    \includegraphics[width=0.9\textwidth]{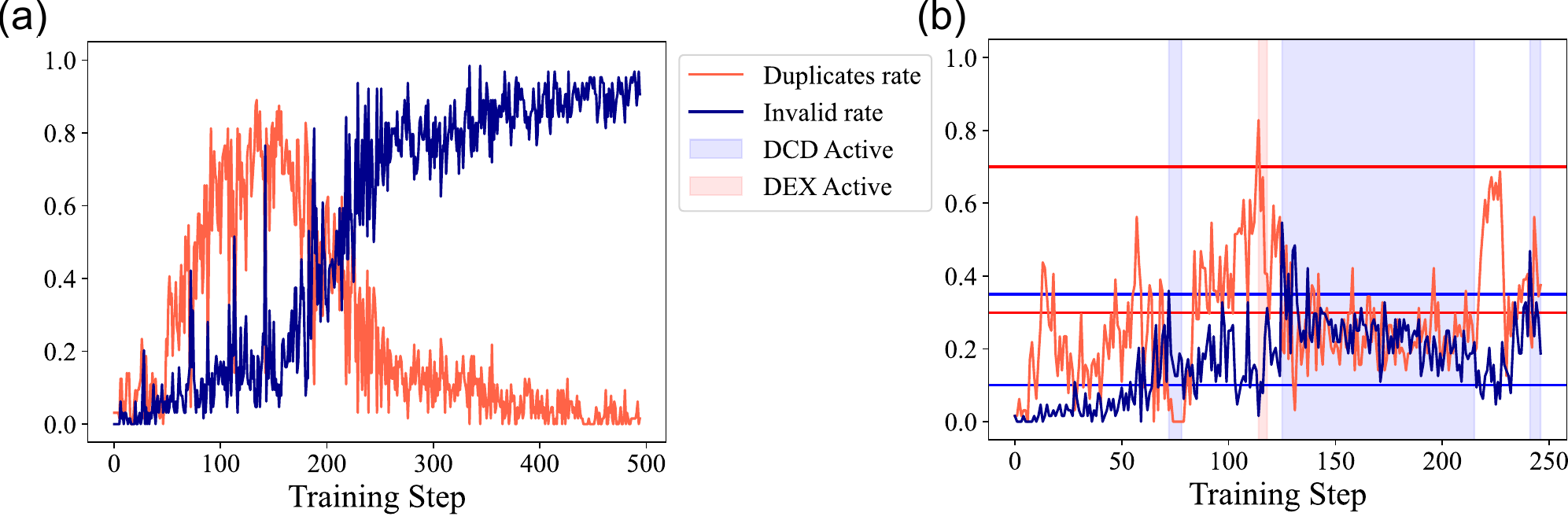}
    \caption{Training stability analysis for the thermoelectric oracle using the transformer model with KL loss and GGS encoding. (a) Evolution of the invalid rate and duplicate rate during training without adaptive mechanisms (Row 4 in Table \ref{tab:auc_table}). (b) Same evolution with Dynamic Cooldown (DCD) and Dynamic Exploration (DEX) enabled (Row 5 in Table \ref{tab:auc_table}). The horizontal dashed lines indicate the activation and deactivation thresholds for DCD and DEX.}
    \label{fig:stability_ana}
\end{figure}

In \cref{sec:adaptive_stability}, we introduced Dynamic Cooldown (DCD) and Dynamic Exploration (DEX) as two adaptive mechanisms to stabilize training. 
These mechanisms prove particularly effective for the thermoelectric oracle, as demonstrated in Table \ref{tab:auc_table}, where switching to DCD and DEX halves the invalid rate.
Here we analyze the dynamics of these mechanisms during training. 
Figure \ref{fig:stability_ana}(a) shows the evolution of the invalid rate (the fraction of invalid molecules sampled by the agent) and the duplicate rate within a sample for the thermoelectric oracle using the transformer model with KL loss and GGS encoding. 
Early in training, the duplicate rate rises, peaking at nearly 80\% around step 100, before dropping sharply as the agent begins to unlearn valid molecular syntax. 
Consequently, the invalid rate approaches 100\% by the end of training, indicating complete training collapse.
This demonstrates that the KL loss is insufficient to maintain training stability in this challenging optimization task.
The adaptive mechanisms DCD and DEX effectively stabilize this training process, completely without KL loss. 
Figure \ref{fig:stability_ana}(b) shows the evolution of both rates for the same configuration with DCD and DEX enabled. 
Around step 75, the invalid rate exceeds the DCD activation threshold of 0.35. Within a few training steps, DCD successfully reduces the invalid rate below the deactivation threshold of 0.1. 
Around step 120, the duplicate rate rises above the DEX activation threshold of 0.7, and again, DEX rapidly reduces it below the deactivation threshold of 0.3 within a few steps. 
Shortly thereafter, the invalid rate rises again, and DCD is reactivated. 
In this instance, stabilization requires approximately 80 training steps. 
Near the end of training, DCD is briefly activated once more to suppress the invalid rate.
The comparison in Figure \ref{fig:stability_ana} clearly demonstrates how DCD and DEX improve training stability.
It should also be noted that the adaptive mechanisms lead to the oracle budget being exhausted in fewer optimization steps, as fewer invalid molecules are generated (the configuration in \cref{fig:stability_ana}(a) requires approximately 500 steps while \cref{fig:stability_ana}(b) requires approximately 250 steps to reach 10,000 oracle calls).
This further enhances the training efficiency.
These adaptive mechanisms are particularly valuable in rugged fitness landscapes with computationally expensive oracles, such as in our NMO benchmark, where classical hyperparameter tuning is infeasible due to the high computational cost.

\subsection{Ablation Studies}\,

\subsubsection{Genetic GFN Framework Components}
\begin{table}[h]
\centering
\setlength{\extrarowheight}{0pt}
\addtolength{\extrarowheight}{\aboverulesep}
\addtolength{\extrarowheight}{\belowrulesep}
\setlength{\aboverulesep}{0pt}
\setlength{\belowrulesep}{0pt}
\caption{\textbf{Full ablation study for our baseline method.}}
\label{tab:auc_table_suppl}
\resizebox{\linewidth}{!}{%
\begin{tabular}{l|ccccc|ccccc|ccccc} 
\toprule
\multirow{2}{*}{\textbf{Variant}}                                                     & \multicolumn{5}{c|}{\textbf{TE Task}}                                                                                                                                                                                                                                                                                                                                                                                                                                        & \multicolumn{5}{c|}{\textbf{PH Task}}                                                                                                                                                                                                                                                                                                                                                                                                                                                & \multicolumn{5}{c}{\textbf{MO Task}}                                                                                                                                                                                                                                                                                                                                                                                                                                   \\ 
\cmidrule(l){2-16}
                                                                                      & \textbf{AUC} $\left.\kern-\nulldelimiterspace\right\uparrow$ & \begin{tabular}[c]{@{}c@{}}\textbf{Mean $f_\mathrm{TE}$}\\\textbf{Top 10} $\left.\kern-\nulldelimiterspace\right\uparrow$\end{tabular} & \begin{tabular}[c]{@{}c@{}}\textbf{Mean SA}\\\textbf{Top 10} $\left.\kern-\nulldelimiterspace\right\downarrow$\end{tabular}  &  \begin{tabular}[c]{@{}c@{}}\textbf{Invalid}\\\textbf{Rate} $\left.\kern-\nulldelimiterspace\right\downarrow$\end{tabular} &  \textbf{RI}         & \textbf{AUC} $\left.\kern-\nulldelimiterspace\right\uparrow$ & \begin{tabular}[c]{@{}c@{}}\textbf{Mean $f_\mathrm{PH}$}\\\textbf{Top 10} $\left.\kern-\nulldelimiterspace\right\uparrow$\end{tabular} & \begin{tabular}[c]{@{}c@{}}\textbf{Mean SA}\\\textbf{Top 10} $\left.\kern-\nulldelimiterspace\right\downarrow$\end{tabular} & \begin{tabular}[c]{@{}c@{}}\textbf{Invalid}\\\textbf{Rate} $\left.\kern-\nulldelimiterspace\right\downarrow$\end{tabular} &  \textbf{RI}        & \textbf{AUC} $\left.\kern-\nulldelimiterspace\right\uparrow$ & \begin{tabular}[c]{@{}c@{}}\textbf{Mean $f_\mathrm{MO}$}\\\textbf{Top 10} $\left.\kern-\nulldelimiterspace\right\uparrow$\end{tabular} & \begin{tabular}[c]{@{}c@{}}\textbf{Mean SA}\\\textbf{Top 10} $\left.\kern-\nulldelimiterspace\right\downarrow$\end{tabular} & \begin{tabular}[c]{@{}c@{}}\textbf{Invalid}\\\textbf{Rate} $\left.\kern-\nulldelimiterspace\right\downarrow$\end{tabular} & \textbf{RI}          \\ 
\midrule\begin{tabular}[c]{@{}l@{}}Original Genetic GFN \\ (SMILES \& ZINC)\end{tabular}         & 0.14 $\pm$ 0.06                                              & 0.23 $\pm$ 0.10                                                                                                                       & \textcolor{red}{4.98}$\pm$ 0.56                                                                                             & 0.37 $\pm$ 0.26                                                                                                           & \ding{55}(0/5)                       & 0.08 $\pm$ 0.03                                              & 0.10 $\pm$ 0.05                                                                                                                        & \textcolor{red}{4.85} $\pm$ 0.31                                                                                            & 0.29 $\pm$ 0.12                                                                                                           & \ding{55}(1/5)                      & 1.29 $\pm$ 0.32                                              & 1.71 $\pm$ 0.42                                                                                                                         & \textcolor{red}{4.55} $\pm$ 0.39                                                                                           & 0.13 $\pm$ 0.06                                                                                                           & \ding{51}(5/5)                    \\
(1): + Synth. Dataset                                                                 & 0.22 $\pm$ 0.07                                              & 0.36 $\pm$ 0.09                                                                                                                       & \textcolor{red}{4.89} $\pm$ 0.23                                                                                            & 0.32 $\pm$ 0.17                                                                                                           & \ding{55}(0/5)                       & 0.10 $\pm$ 0.04                                              & 0.20 $\pm$ 0.10                                                                                                                        & \textcolor{red}{5.16} $\pm$ 0.13                                                                                            & 0.21 $\pm$ 0.08                                                                                                           & \ding{55}(1/5)                      & 0.62 $\pm$ 0.31                                              & 0.85 $\pm$ 0.32                                                                                                                         & \textcolor{red}{5.15} $\pm$ 0.39                                                                                           & 0.49 $\pm$ 0.32                                                                                                                      & \ding{51}(5/5)                       \\
(2): + Switch to GGS                                                                  & 0.63 $\pm$ 0.22                                              & 0.91 $\pm$ 0.18                                                                                                                       & 3.59 $\pm$ 0.33                                                                                                             & 0.46 $\pm$ 0.33                                                                                                           & \ding{51}(5/5)                      & 0.52 $\pm$ 0.22                                              & 0.75 $\pm$ 0.25                                                                                                                        & 3.35 $\pm$ 0.23                                                                                                             & 0.12 $\pm$ 0.13                                                                                                           & \ding{51}(4/5)                      & 0.55 $\pm$ 0.10                                              & 0.79 $\pm$ 0.13                                                                                                                         & 4.17 $\pm$ 0.30                                                                                                             & 0.14 $ \pm$0.15                                                                                                                      & \ding{51}(5/5)                    \\
(3): + Transformer arch.                                                              & 0.65 $\pm$ 0.13                                              & 0.97 $\pm$ 0.35                                                                                                                       & 3.55 $\pm$ 0.42                                                                                                             & 0.53 $\pm$ 0.36                                                                                                           & \ding{51}(5/5)                      & 0.31 $\pm$ 0.15                                              & 0.50 $\pm$ 0.20                                                                                                                        & 3.36 $\pm$ 0.21                                                                                                             & 0.15 $\pm$ 0.11                                                                                                           & \ding{55}(3/5)                      & 0.69 $\pm$ 0.29                                              & 1.20 $\pm$ 0.49                                                                                                                         & 4.38 $\pm$ 0.30                                                                                                             & 0.23 $\pm$ 0.14                                                                                                                      & \ding{51}(5/5)                   \\
(4): + Stab. Mechanisms ($-$KL)                                                       & 0.60 $\pm$ 0.11                                              & 0.78 $\pm$ 0.17                                                                                                                       & 3.94 $\pm$ 0.31                                                                                                             & 0.19 $\pm$ 0.07                                                                                                           & \ding{51}(5/5)                      & 0.32 $\pm$ 0.18                                              & 0.56 $\pm$ 0.31                                                                                                                        & 3.38 $\pm$ 0.25                                                                                                             & 0.14 $\pm$ 0.09                                                                                                           & \ding{55}(3/5)                      & 0.74 $\pm$ 0.13                                              & 1.04 $\pm$ 0.13                                                                                                                         & \textcolor{red}{4.56} $\pm$ 0.27                                                                                            & 0.21 $\pm$ 0.08                                                                                                                      & \ding{51}(5/5)                     \\
\rowcolor[rgb]{0.949,0.949,0.949}(5): + Descriptors                                   & 0.78 $\pm$ 0.23                                              & 1.19 $\pm$ 0.42                                                                                                                       & 4.06 $\pm$ 0.30                                                                                                             & 0.21 $\pm$ 0.13                                                                                                           & \ding{51}(5/5)                    & 0.33 $\pm$ 0.16                                              & 0.75 $\pm$ 0.48                                                                                                                        & 3.37 $\pm$ 0.42                                                                                                             & 0.32 $\pm$ 0.22                                                                                                           & \ding{51}(4/5)                      & 0.42 $\pm$ 0.08                                              & 0.59 $\pm$ 0.16                                                                                                                         & 4.40 $\pm$ 0.37                                                                                                               & 0.18  $\pm$ 0.05                                                                                                                    & \ding{51}(5/5)                       \\
\bottomrule
\end{tabular}
}
\end{table}

We provide a comprehensive ablation study of our genetic GFN framework extensions in \cref{tab:auc_table_suppl}, systematically evaluating the impact of each component on performance metrics and the ability to discover relevant molecules.
Because the genetic GFN approach relies on a sampling step not present in the other methods, it faces unique stability challenges. Therefore, we evaluate its invalid rate, defined as the percentage of invalid samples in the final step. An invalid molecule can be for example a string that does not correspond to the scheme of GGS written in \cref{sec:ggs}.

\uline{TE:} \,
The SMILES variants (including (1)) fail to find high-performing candidates, as this encoding cannot natively model the two-sided gold binding required for MJs.
Switching to GGS (2) resolves this topology mismatch and substantially increases AUC and fitness, including robust RI discovery.
The transformer architecture (3) leaves TE performance unchanged within noise.
A persistent issue across (2) and (3) is the high invalid rate, reflecting the rugged loss landscape.
Adaptive Stability Control (4) cuts this rate by more than half, with a small drop in fitness.
Finally, auxiliary molecular descriptors (5) provide the agent with structure-property intuition, yielding the best mean AUC and fitness, though with high seed-to-seed variance. \\[0.33ex]
\uline{PH:} \,
SMILES-based variants (including (1)) again struggle with the dual binding topology.
GGS (2) is the decisive change here, achieving the highest AUC and fitness on PH and robustly finding relevant candidates.
The subsequent additions (3)-(5) trade some AUC and fitness but still demonstrate the ability to discover physically relevant molecules.
The full model (5) robustly finds relevant candidates, despite no clear advantage on the proxy metrics.
The lowest thermal conductance found in (5) is $0.10$ pW/K, which is slightly lower than the best candidate from (2) with $0.13$ pW/K.
\\[0.33ex]
\uline{MO:} \,
MO requires only one-sided binding and is the only task where the distributional bias from pharmaceutical pretraining appears to help.
The original setup (SMILES + ZINC) performs competitively on AUC and fitness, but its mean SA score lies above the synthesizability threshold.
Removing the ZINC pretraining (1) drops performance and worsens SA.
Switching to GGS (2) brings SA below threshold, which is crucial for practical applications, at the cost of some AUC and fitness.
The transformer (3) and Adaptive Stability Control (4) recover AUC progressively.
In contrast to the other oracles, auxiliary descriptors (5) measurably degrade MO performance, suggesting the auxiliary objective conflicts with this task's optimization. \\[0.33ex]

\subsubsection{Fragment Library}
\label{sec:library_ablation}
The choice of fragments in the GGS encoding can impact the performance of our method.
We have carefully selected the fragments incorporating domain knowledge, as explained in \cref{sec:building_blocks_selection}.
The discussion of the top-performing candidates in \cref{fig:top_molecules} shows that the acetylene group is a critical motif for all tasks.
For the TE and PH tasks (which both benefit from low phononic transport), this is expected, as specialized literature has identified this motif as a key design principle for low phononic transport~\cite{blaschke2025revealing}.
In general, acetylene groups have been reported to show beneficial properties in these molecular junctions~\cite{stefani2018large, yan2024substituents}.
Our approach rediscovers this, but for scientific rigor, we explicitly remove the acetylene group from the fragment library and run all three tasks again.

The comparison of our baseline method with and without the acetylene group is shown in \cref{tab:auc_table_fragment_ablation}.
For the TE task, removing the acetylene group leads to a drop in AUC and mean fitness, while the SA score of the top candidates improves.
The invalid rate stays unchanged compared to the baseline with the full fragment library.
The method still discovers robustly candidates that surpass the RI threshold.

The PH task is most affected by the removal of the acetylene group, with a significant drop in AUC and mean fitness.
The SA score stays at the same level, and the invalid rate decreases slightly.
In summary, no candidate passed the RI threshold without the acetylene group.
This can be explained by the fact that the acetylene group is a key motif for low phononic transport, and its absence makes it more difficult for the agent to discover high-performing candidates.
The RI threshold is taken from specialized literature optimizing for low phononic transport~\cite{blaschke2025revealing}, making it very strict in the first place.
Top-performing candidates for the PH task in that study also contain acetylene groups, and the underlying physics is explained in detail.
Therefore, the significant performance drop when removing the acetylene group is not an artifact of our method, but rather a reflection of the underlying physics of the problem.

For the MO task, it was not initially clear that the acetylene group is beneficial for performance.
Removing the acetylene group leads to a drop in AUC and mean fitness, while the SA score of the top candidates improves.
The invalid rate stays unchanged compared to the baseline with the full fragment library.
Notably, the molecules still show the same motifs (conjugated chains and aromatic rings), but without the acetylene group itself.

In summary, the fragment library naturally contains some bias (as all fragment-based approaches do).
However, we discuss the enabling components of our benchmark in detail, allowing the community to develop methods that can perform well.
Most interesting for future work would be approaches with an open vocabulary that discover similar important motifs, such as the acetylene group, without being explicitly given them as building blocks.

\begin{table}[h]
\centering
\setlength{\extrarowheight}{0pt}
\addtolength{\extrarowheight}{\aboverulesep}
\addtolength{\extrarowheight}{\belowrulesep}
\setlength{\aboverulesep}{0pt}
\setlength{\belowrulesep}{0pt}
\caption{\textbf{Full ablation comparing baseline method with and without acetylene group.}}
\label{tab:auc_table_fragment_ablation}
\resizebox{\linewidth}{!}{%
\begin{tabular}{l|ccccc|ccccc|ccccc} 
\toprule
\multirow{2}{*}{\textbf{Variant}}                                                     & \multicolumn{5}{c|}{\textbf{TE Task}}                                                                                                                                                                                                                                                                                                                                                                                                                                        & \multicolumn{5}{c|}{\textbf{PH Task}}                                                                                                                                                                                                                                                                                                                                                                                                                                                & \multicolumn{5}{c}{\textbf{MO Task}}                                                                                                                                                                                                                                                                                                                                                                                                                                   \\ 
\cmidrule(l){2-16}
                                                                                      & \textbf{AUC} $\left.\kern-\nulldelimiterspace\right\uparrow$ & \begin{tabular}[c]{@{}c@{}}\textbf{Mean $f_\mathrm{TE}$}\\\textbf{Top 10} $\left.\kern-\nulldelimiterspace\right\uparrow$\end{tabular} & \begin{tabular}[c]{@{}c@{}}\textbf{Mean SA}\\\textbf{Top 10} $\left.\kern-\nulldelimiterspace\right\downarrow$\end{tabular}  &  \begin{tabular}[c]{@{}c@{}}\textbf{Invalid}\\\textbf{Rate} $\left.\kern-\nulldelimiterspace\right\downarrow$\end{tabular} &  \textbf{RI}         & \textbf{AUC} $\left.\kern-\nulldelimiterspace\right\uparrow$ & \begin{tabular}[c]{@{}c@{}}\textbf{Mean $f_\mathrm{PH}$}\\\textbf{Top 10} $\left.\kern-\nulldelimiterspace\right\uparrow$\end{tabular} & \begin{tabular}[c]{@{}c@{}}\textbf{Mean SA}\\\textbf{Top 10} $\left.\kern-\nulldelimiterspace\right\downarrow$\end{tabular} & \begin{tabular}[c]{@{}c@{}}\textbf{Invalid}\\\textbf{Rate} $\left.\kern-\nulldelimiterspace\right\downarrow$\end{tabular} &  \textbf{RI}        & \textbf{AUC} $\left.\kern-\nulldelimiterspace\right\uparrow$ & \begin{tabular}[c]{@{}c@{}}\textbf{Mean $f_\mathrm{MO}$}\\\textbf{Top 10} $\left.\kern-\nulldelimiterspace\right\uparrow$\end{tabular} & \begin{tabular}[c]{@{}c@{}}\textbf{Mean SA}\\\textbf{Top 10} $\left.\kern-\nulldelimiterspace\right\downarrow$\end{tabular} & \begin{tabular}[c]{@{}c@{}}\textbf{Invalid}\\\textbf{Rate} $\left.\kern-\nulldelimiterspace\right\downarrow$\end{tabular} & \textbf{RI}          \\ 
\rowcolor[rgb]{1.0,1.0,1.0}Baseline                                   & 0.78 $\pm$ 0.23                                              & 1.19 $\pm$ 0.42                                                                                                                       & 4.06 $\pm$ 0.30                                                                                                             & 0.21 $\pm$ 0.13                                                                                                           & \ding{51}(5/5)               & 0.33 $\pm$ 0.16                                              & 0.75 $\pm$ 0.48                                                                                                                        & 3.37 $\pm$ 0.42                                                                                                             & 0.32 $\pm$ 0.22                                                                                                           & \ding{51}(4/5)                      & 0.42 $\pm$ 0.08                                              & 0.59 $\pm$ 0.16                                                                                                                         & 4.40 $\pm$ 0.37                                                                                                               & 0.18  $\pm$ 0.05                                                                                                                    & \ding{51}(5/5)                       \\
\rowcolor[rgb]{1.0,1.0,1.0}Baseline without acetylene                 & 0.53 $\pm$ 0.05                                              & 0.74 $\pm$ 0.12                                                                                                                       & 3.50 $\pm$ 0.18                                                                                                             & 0.26 $\pm$ 0.08                                                                                                           & \ding{51}(5/5)                    & 0.11 $\pm$ 0.01                                              & 0.15 $\pm$ 0.03                                                                                                                        & 3.20 $\pm$ 0.24                                                                                                             & 0.12 $\pm$ 0.07                                                                                                           & \ding{55}(0/5)                      & 0.28 $\pm$ 0.09                                              & 0.39 $\pm$ 0.09                                                                                                                         & 3.58 $\pm$ 0.46                                                                                                               & 0.16  $\pm$ 0.08                                                                                                                    & \ding{51}(5/5)                       \\

\bottomrule
\end{tabular}
}
\end{table}

\subsection{Analysis of the Best Performing Candidates}
\label{sec:extended_physics_ana}
In this section, we take a closer look at the best-performing candidates identified by our method.
We provide in-depth analyses of their physical properties and theoretically ground the reasons for their outstanding performance.
In addition to the presented selection, we provide all candidates proposed by our baseline that exceed the RI threshold (see \cref{sec:code_and_data_avail}).
\subsubsection{Thermoelectric oracle}

\begin{figure}
    \centering
    \includegraphics[width=0.8\textwidth]{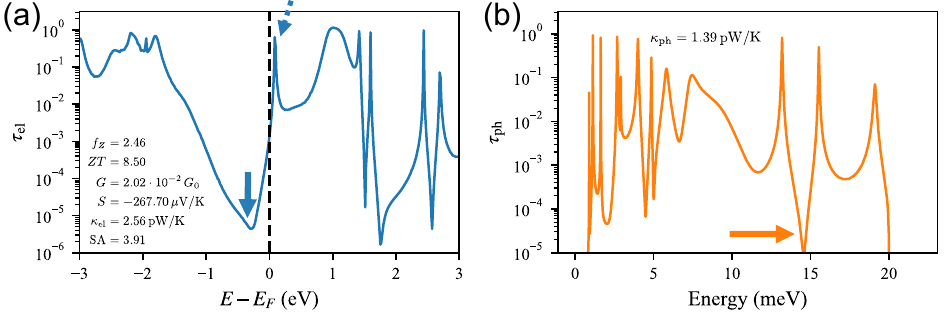}
    \caption{Quantum transport properties of the best-performing candidate for the thermoelectric oracle shown in \cref{fig:top_molecules}(a). (a) Electronic transport properties and transport coefficients given in the plot. The solid blue arrow indicates a destructive quantum interference feature in the transmission. The dashed blue arrow marks a resonance close to the Fermi energy. (b) Phononic transport properties. The calculated thermal conductance is indicated in the plot. The orange arrow indicates a phononic destructive quantum interference.}
    \label{fig:TE_ana_detail}
\end{figure}

\begin{figure}
    \centering
    \includegraphics[width=0.8\textwidth]{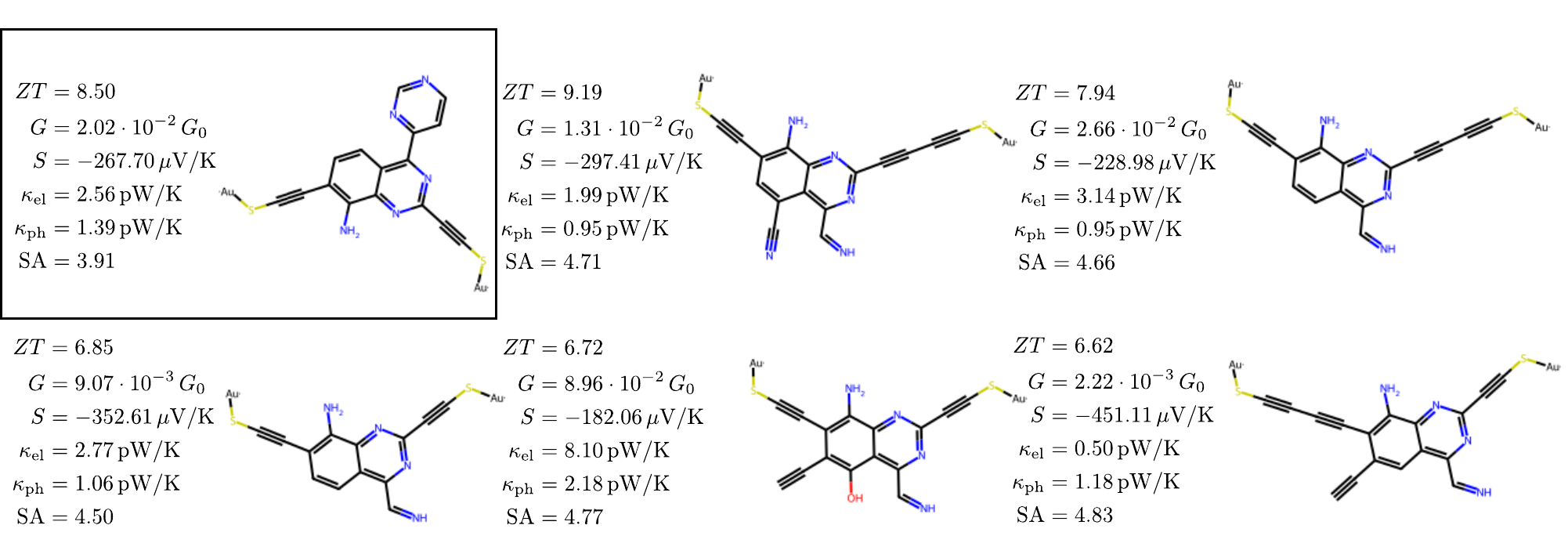}
    \caption{Structural formulas for the molecule shown in \cref{fig:top_molecules}(a) (highlighted by a black box) and additional unique top-performing candidates from the same optimization run. Transport coefficients are provided next to each molecule. The electrodes are attached at the gold-thiol groups on both sides.}
    \label{fig:more_te_candidates} 
\end{figure}

In this section, we analyze the transport properties of the top-performing candidate identified by the thermoelectric oracle, shown in Figure~\ref{fig:top_molecules}(a).
In addition to the three-dimensional representation in Figure~\ref{fig:top_molecules}(a), we provide a two-dimensional structural formula in the top left panel of Figure~\ref{fig:more_te_candidates} for clarity.
The dimensionless thermoelectric figure of merit $ZT$ reaches a value of $8.50$ at room temperature ($T=300~\mathrm{K}$).
This value is exceptionally high and lies significantly above the proposed threshold of $ZT>3$ for technologically relevant thermoelectrics~\cite{gemma2021roadmap}.
The outstanding performance is a combination of electronic and phononic transport effects.
The electronic transport shown in Figure~\ref{fig:TE_ana_detail}(a) is characterized by a combination of destructive quantum interference and a sharp resonance close to the Fermi energy. 
This results in a large absolute Seebeck coefficient and simultaneously high electrical conductance. 
The resonance state originates from the amino group ($\mathrm{NH_2}$, also marked in \cref{fig:top_molecules}(a)) attached to the central ring, which reduces the HOMO-LUMO gap size~\cite{li2015effects}. 
The destructive interference arises from the overall meta-type configuration of the molecule \cite{reznikova2021substitution, yan2024substituents}.
Here, the meta configuration refers to a kinked molecular geometry, as illustrated by the bent line in \cref{fig:top_molecules}(a).

The phononic transport is also strongly suppressed, which is crucial for good thermoelectric performance. 
First, the molecule features acetylene groups near the anchoring sites, which have been identified as critical in previous literature~\cite{blaschke2025revealing}. 
Additionally, the phononic transmission exhibits destructive quantum interference, typically induced by side groups such as the amino group or the pyrimidine ring \cite{klockner2017tuning}. 
Both characteristics responsible for the suppressed phononic transport are highlighted in \cref{fig:top_molecules}(a).

In general, our algorithm combines molecular motifs that have been individually studied in the literature to solve the high-dimensional multi-objective optimization problem. 
It is remarkable that these features are found without any explicit prior knowledge within a very limited number of oracle calls. 

In experimental conditions, the $ZT$ value would likely be somewhat lower (see limitations in~\cref{sec:te_details}). 
However, the identified molecule is highly promising compared to theoretical predictions for other molecules at room temperature using similar modeling approaches, which achieved a maximum of $ZT=2.4$ \cite{ZT2}.
The combination of robust transport features, reasonable synthetic accessibility, and standard chemical building blocks makes our candidate a promising route toward achieving high $ZT$ values.
Thus, the proposed molecule represents a high-value candidate for future research, which, to the best of our knowledge, has not previously been reported in the literature.
Our algorithm discovers many more such candidates, as indicated by the binary metric of Table~\ref{tab:auc_table}.
We give an overview of additional high-performing candidates from the same optimization run in Figure~\ref{fig:more_te_candidates}.
A detailed analysis of all candidates is beyond the scope of this work. 
However, the code is made available to the scientific community to facilitate further investigation of these candidates.

\subsubsection{Phonon Transport Oracle}

\begin{figure}
    \centering
    \includegraphics[width=0.9\textwidth]{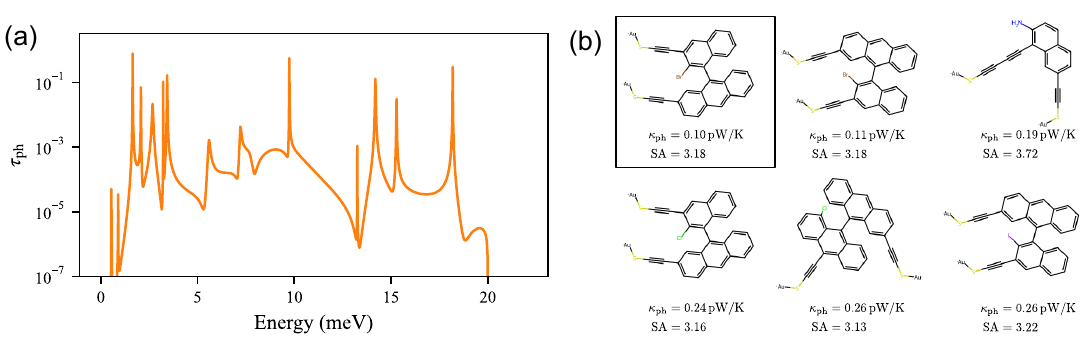}
    \caption{(a) Phononic transmission for the candidate shown in Figure~\ref{fig:top_molecules}(b), along with the calculated thermal conductance at room temperature. (b) Structural formulas for the candidate under discussion (highlighted by a black box) and additional unique high-performing candidates from the same optimization run. Phononic thermal conductance values at room temperature are indicated below each molecule. Electrodes are attached at the gold-thiol groups on both sides.}
    \label{fig:best_phonon_ana}
\end{figure}

This section provides a detailed analysis of the best-performing candidate identified by the phonon transport oracle, shown in Figure~\ref{fig:top_molecules}(b). 
In addition to the three-dimensional representation in the main text, we provide a two-dimensional structural formula in the top left of Figure~\ref{fig:best_phonon_ana}(b) for clarity.
The phonon transmission $\tau_\mathrm{ph}$ is depicted in Figure~\ref{fig:best_phonon_ana}(a) along with the calculated thermal conductance at room temperature ($T=300~\mathrm{K}$). 
The thermal conductance is $\kappa_\mathrm{ph} = 0.0990\,\mathrm{pW/K}$, rounded to $\kappa_\mathrm{ph} = 0.10\,\mathrm{pW/K}$, which is extremely low for a single-molecule junction. 
This low thermal conductance can be attributed to the acetylene linkers positioned near the anchoring sites at both ends of the molecule (encircled in \cref{fig:top_molecules}(b)), which have been identified as crucial for suppressing phononic transport in previous literature~\cite{blaschke2025revealing}.
It was shown that these acetylene groups act as phonon filters, effectively blocking phonons from propagating through the junction. 
This is evident from the generally low phononic transmission, which hardly reaches a value of 1 across the entire energy range.
An additional effect also discussed in \citet{blaschke2025revealing} is the suppression of thermal conductance through the dihedral angle between the anthracene and naphthalene rings, which is locked by steric repulsion from the bromine atom.
The twist angle is also highlighted in \cref{fig:top_molecules}(b).
The structure of our molecule is fundamentally different from those previously studied, suggesting that the transport-suppressing effects identified in the literature are robust and manifest in diverse geometries.
Previous work on low phononic transmission has primarily investigated the so-called para and meta configurations of molecules, which describe how rings are interconnected (see $p$OPE3 and $m$OPE3 in Figure~\ref{fig:thermoelectric_junction}(c)). 
Here, our algorithm selects a completely novel substitution pattern by connecting the anthracene rings at a central position rather than at the typical para or meta positions.

The cited study serves as our primary reference for evaluating phononic transport suppression. 
That work, focused exclusively on optimizing phonon transport, identifies a molecule with $\kappa_\mathrm{ph} = 0.07\,\mathrm{pW/K}$ but with an SA score of $4.4$. 
All other molecules in the cited study exhibit higher thermal conductance than our candidate while maintaining SA scores around $4.4$.
Our candidate achieves a lower (better) SA score of $3.18$.
We therefore conclude that we have identified a significantly better-performing candidate. 
As experimental measurement techniques for such systems are only now being developed \cite{luan2026tuning, yelishala2025phonon, mosso:NanoLett2019, cui2019thermal}, with the first high-resolution results emerging, our work contributes to this emerging field by proposing promising candidates. 
Additional candidates from the same optimization run that produced the presented molecule are shown in Figure~\ref{fig:best_phonon_ana}(b). 
They share similar structural motifs, such as acetylene linkers near the anchoring sites and side groups on the central rings that induce a dihedral angle.
A detailed analysis of all these candidates is beyond the scope of this work.
However, the code will be made available to the scientific community to facilitate further investigation of these candidates.

\subsubsection{Molecular Optomechanics Oracle}
\label{sec:thz_results}
This section contains the discussion of best-performing molecules proposed by the MO task. 
In contrast to the other applications, we performed additional validation of the proposed top-performing molecules with more accurate, DFT-level simulations of their target property $P$, to investigate the reliability of the PTB method.  
The molecule depicted in Figure~\ref{fig:top_molecules}c proposed by our full baseline method was selected based on a combination of properties: it has a high $P$ value according to both PTB and DFT calculations (9.91 and 8.31, respectively), good synthetic accessibility score (4.35) and a molecular shape promoting SAM and nanocavity formation (small $S$ and moderate $g$).   
While other molecules achieved significantly higher fitness scores, those molecules tend to have a very high error in the $P$ value predicted by PTB compared to DFT results (see discussion below), and/or a high SA score (above 4.5).
The selected molecule consists of a conjugated chain including double and triple bonds and an aromatic head group, and its most intensive vibration for frequency upconversion is a combination of in-plane bending motions of the chain and the aromatic end group. 
The high performance of this molecule can be rationalized by its increased polarizability along the molecular backbone due to extended conjugation, leading to high Raman intensities, while the bending motions also induce changes in the dipole moment, providing a good THz absorption capability. 
This type of in-plane bending modes of conjugated chains has not been identified for this application before. 
Previous studies to identify good candidates for THz detection were either limited to commercially available drug-like molecules \cite{koczor2021molecular}, or relied on training databases containing experimentally reported crystal forming molecules \cite{koczor2025generative}, and completely missed similar chain motifs.
Conjugated molecular wires, such as oligoyne \cite{bryce2021review} and oligo(arylene ethynylene) derivatives \cite{o2021review} have been used extensively in molecular electronics applications, and for the latter, stable and reproducible SAMs have been demonstrated \cite{valkenier2011formation}. 
The stability and SAM forming capabilities of the current molecule will need to be studied experimentally to verify the applicability of this molecule for THz detection. 
Many of the top-performing molecules share a similar structural motif, consisting of a thiol anchor, a conjugated chain including various numbers of double and triple bonds, and a heteroaromatic and/or functionalized aromatic end group. 

DFT validation of 180 top-performing molecules across different optimization runs shows that many of the proposed molecules indeed have an exceptional upconversion capability, with 78 having $P$ values above the $P=7.88$ value of the best candidate identified in \citet{koczor2025generative}.
 However, we also found that $P$ values are severely overestimated by PTB for some molecules. 
In a significant portion of the molecules generated by the original model, the PTB determined $P$ values can be as high as 26-29, whereas DFT gives $P$ values between 5-12.
This indicates that these molecules are outside the chemical space for which PTB was developed, likely due to extended delocalization of electrons.
However, we note that $P$ is measured on a logarithmic scale relative to prior work (see Equation \eqref{eq:P_value_def}).
Thus, every candidate with $P > 0$ is still an improvement over the mean $\mu$ of this reference.
Analyzing the shortfall of the PTB method for these molecules, and introducing improved predictors for the spectroscopic performance of molecules is a challenging task, which is outside the scope of the current study.
A recently published xTB version \cite{froitzheim2025g} may improve accuracy for these molecules.
 
\begin{figure}
    \centering
    \includegraphics[width=\textwidth]{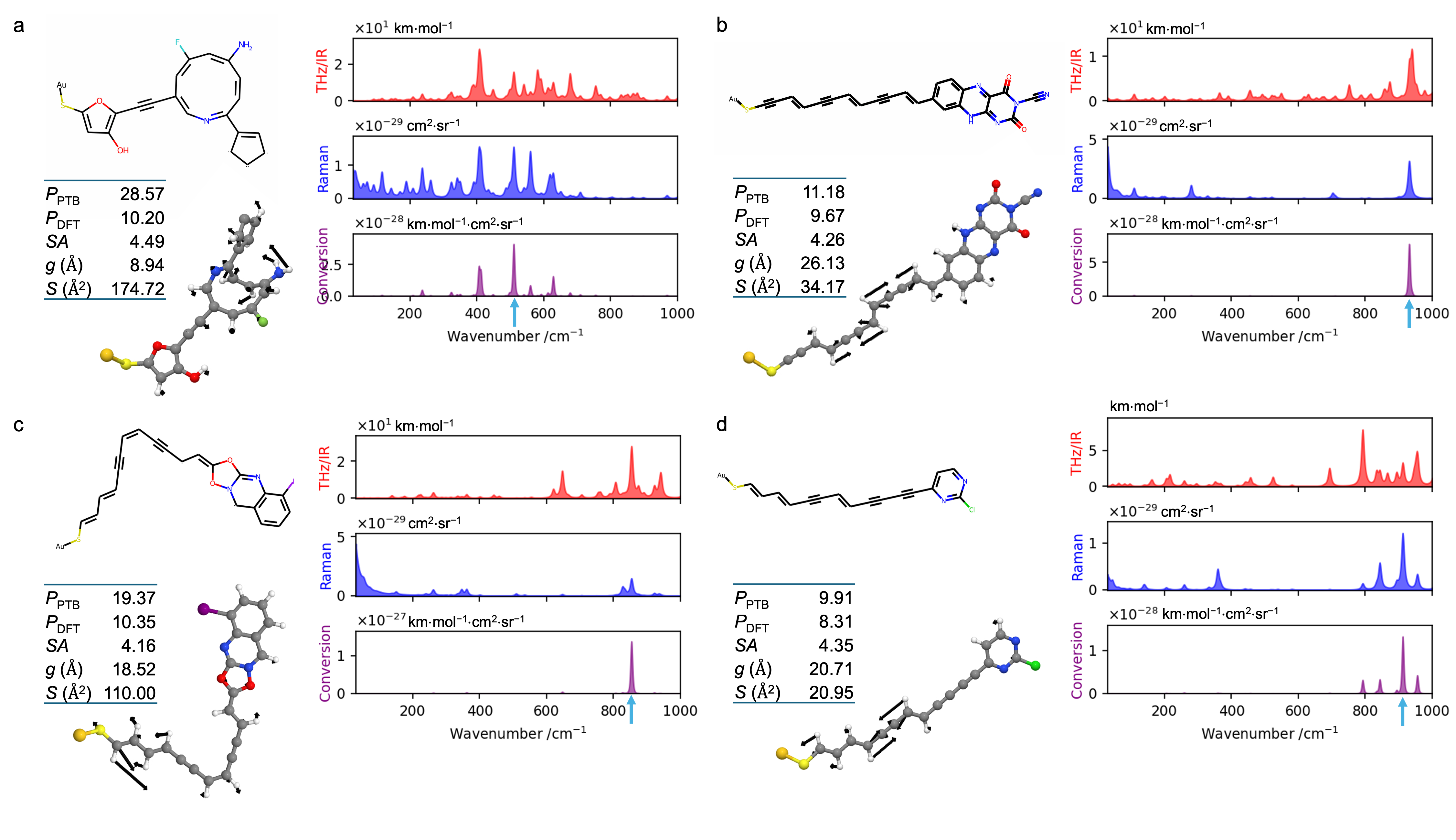}
    \caption{Selected top-performing candidates for THz upconversion and their relevant properties, along with their simulated absorption, Raman scattering and upconversion spectra. Vibrational spectra were calculated at the DFT level.  The atomic displacement vectors are also depicted for the most intensive vibrational mode for upconversion (marked by a blue arrow under the upconversion spectrum) of each molecule. }
    \label{fig:thz_mols} 
\end{figure}

Figure~\ref{fig:thz_mols} showcases different structural motifs and vibrational properties uncovered by our generative approach. The original model identified molecules containing a 10-membered azecine ring (as in Figure~\ref{fig:thz_mols}(a)), which is a structural motif present in some complex natural products and is of interest for drug design \cite{listratova2024advances}, but the isolated ring is highly unstable. 
This highlights problems with relying on drug discovery training datasets and also shows some limitations of the SA score.
The molecule depicted in Figure~\ref{fig:thz_mols}(a) has multiple highly active modes for frequency upconversion, which typically involve the out-of-plane bending motion of an amino (NH$_2$) group.
This bending motion of the amino group has already been identified in previous studies \cite{koczor2021molecular,koczor2025generative}, where most of the top performing molecules contain this functional group. 
The original Genetic GFN model successfully rediscovered this structural motif, and proposed candidates with significantly higher $P$ values. 
In contrast to the original model, the other Genetic GFN variants rely on the curated building block library (\cref{fig:building_blocks}), and they all tend to propose the already discussed conjugated chain and aromatic end group motifs for the highest-performing molecules (see, for example, Figure~\ref{fig:thz_mols}(b-d).
The most intensive vibrations of these molecules are in-plane bending modes. 
Molecules (b) and (d) have a planar, rod-like structure which results in a small footprint and thus higher predicted molecular density in SAMs, while for molecule (c) the automatic structure optimisation procedure arrived at a bent (cis isomer) structure. 
Cis-trans isomerization of the conjugated chains is expected to affect the position and intensity of the high-intensity vibrational peaks, which may be harnessed in photoswitching applications.

\subsection{Training Details}
\label{sec:training_details}
In this section, further details on the model architectures, hyperparameters, and training procedures are provided.
Furthermore, compute resources for the oracle evaluations and trainings are discussed.

\paragraph{Model Architectures}
The agent is parameterized by an autoregressive causal decoder-only Transformer architecture \cite{vaswani2017attention}.
We utilize Rotary Positional Embeddings (RoPE) \cite{jianlin2024roformer} to inject relative position information at each attention layer.
The network consists of $3$ layers with $8$ query attention heads, $2$ key/value attention heads, and an embedding dimension of $512$.
The final embedding is processed by two separate linear heads: A Policy Head for the action logits, predicting the next action to build the molecule, and a Descriptor Head, which predicts the $17$ scalar molecular descriptors used as auxiliary guidance during training and pretraining.

The original Genetic GFN \cite{kim2024geneticgflow} builds on a Gated Recurrent Unit (GRU) architecture \cite{cho2014gru}, which processes the sequence sequentially, maintaining a hidden state that acts as a compressed memory of the trajectory.
The model has a hidden size of $512$ and consists of $3$ layers.

\paragraph{Hyperparameters}
The model is pretrained on a procedurally generated dataset of $\num{300000}$ valid molecules in GGS representation.
We pretrain for 3 (GGS encoding) and 5 (SMILES encoding) epochs using the Adam optimizer \cite{diederik2015adam} with a learning rate of $1 \times 10^{-3}$ and a batch size of $128$.
The loss function is a weighted sum of the Negative Log-Likelihood (NLL) for sequence generation and the Mean Squared Error (MSE) for descriptor prediction.

For the optimization phase, we utilize the Trajectory Balance (TB) objective with a learned scalar partition function $Z_\mathrm{p}$.
The reward scaling coefficient is set to $\beta=30$, such that the reward is $R(x) = \exp(30 \cdot f_i(x))$ with the fitness function $f_i(x)$ from the oracles.
We note that the scaling constants in the fitness functions (Equation~\eqref{eq:fitness_struc}) have the same effect as changing $\beta$. 
We chose these constants to achieve fitness ranges similar to those in previous literature \cite{kim2024geneticgflow} and adopted their $\beta$ values.
The Adam optimizer is used with decoupled learning rates: $5 \times 10^{-4}$ for the agent parameters $\theta$ and $0.1$ for optimizing the scalar $\log Z_\mathrm{p}$.
Gradients are clipped at a norm of $10$ to prevent instability.
The base batch size is $64$.
The replay buffer has a maximum size of $1024$ and utilizes rank-based sampling with a base coefficient of $c=0.01$.
The replay training has $8$ iterations per training step.
The genetic search has an initial mating pool size of $64$ candidates, from which $8$ offspring molecules are generated per generation. 
We use two generations per training step with a crossover and mutation probability of $0.5$ each. 
The maximum token length is set to $140$ for SMILES. 
For GGS, we use a limit of $30$ tokens for the molecular optomechanics and thermoelectrics tasks, and $18$ tokens for phonon transport. 
Note that SMILES uses atomic tokens only, while in GGS each token can represent an entire fragment comprising multiple atoms.
The stricter limit for phonon transport prevents the trivial solution of generating large molecules, which, despite effectively suppressing phononic conductance, result in prohibitive computation times and poor SA scores.

\subsubsection{Compute Resources}
\label{sec:comput_res_baseline}
Our approach requires relatively modest resources for pretraining. 
Pretraining in the GGS approach with the transformer model, including the generation of the random dataset, takes only approximately 10 minutes with 16 cores on an AMD EPYC 7713 in combination with an NVIDIA A40 GPU. 
In Chapter \ref{sec:pretraining_ana}, we have also shown that only a few epochs are necessary for pretraining.
The transformer model has approximately 8.5 million parameters, while the GRU model used in the original Genetic GFN has approximately 4.3 million parameters.

During the main training phase, the extensive quantum chemistry calculations represent the bottleneck. 
One seed for the phonon oracle requires approximately 50 hours on 36 cores of an Intel Xeon Platinum 8360Y processor. 
The upconversion oracle is somewhat faster and requires approximately 40 hours on the same hardware. 
The thermoelectric oracle requires approximately 24 hours with the same setting. 
The exact duration depends significantly on the sampling performance, which is discussed in \cref{sec:adaptive_stability}, and the size of the molecules. 
If the agent generates molecules inefficiently, the training takes considerably longer, as more steps must be executed to exhaust the oracle budget.

\subsection{Experiments on the PMO Benchmark}
\label{sec:PMO_benchmark}
While we explicitly focus on our Nanotechnology Molecular Optimization (NMO) benchmark suite and its specific challenges, we can also apply our baseline method to the well-established Practical Molecular Optimization (PMO) benchmark \cite{gao2022pmo} from drug discovery.
This benchmark consists of 23 tasks that evaluate the ability of generative models to optimize various drug-like properties.
The results for the PMO benchmark are summarized in Table \ref{tab:pmo_results}, including comparisons to literature results for methods shown in \cref{tab:auc_table}. 

f-RAG and GenMol use task specific vocabularies and hyperparameters, for each of the 23 tasks.
The custom vocabularies are built by heavily querying the oracle prior to optimization, which effectively undermines the imposed limit of 10,000 allowed oracle calls \cite{kaech2025invirtuogen}.
Under this approach, each optimization is initialized near the optimal solution, preventing the benchmark from meaningfully assessing actual optimization performance for these methods.
Within this setting f-RAG and GenMol achieve the highest AUC scores on the PMO benchmark.

We compare the original Genetic GFN \cite{kim2024geneticgflow} with their pretraining dataset to all our ablation steps from Table \ref{tab:auc_table}. 
The fifth column in Table \ref{tab:pmo_results} shows the results for the original Genetic GFN with the original pretraining based on the ZINC dataset.
We note that the values differ slightly from the original publication \cite{kim2024geneticgflow}, most likely due to differences in random seeds.
The performance drops significantly without the ZINC-based pretraining dataset (sixth column), indicating that Genetic GFN also benefits from pretraining bias. 
Switching to our GGS encoding (seventh column) achieves performance between the original Genetic GFN and the random dataset version. 
Columns six and seven employ the same model architecture and pretraining dataset, differing only in the molecular encoding.
Based on these results, GGS encoding outperforms SMILES when dataset bias from pretraining is removed.
Changing to the transformer architecture (eighth column) slightly decreases performance. 
While this considerably larger transformer model proves beneficial for our complex NMO benchmark, it appears to overfit on the PMO benchmark with its relatively simple tasks. 
Adding DCD and DEX (ninth column) improves performance again. 
Notably, DCD and DEX are only activated for some seeds in the $\texttt{isomers\_c7h8n2o2}$ oracle, as training is already stable without DCD and DEX.
The main difference arises from the absence of KL loss in this configuration.
Finally, adding the descriptors in pretraining and training slightly decreases performance. 
In the simple energy landscape of the PMO benchmark without hard constraints, these descriptors appear to provide limited benefit.

Compared to molGA and REINVENT, the Genetic GFN family in \cref{tab:pmo_results} performs on par across the ablation variants, confirming that the underlying framework remains competitive on PMO across encodings and architectural choices.

\begin{sidewaystable}[ht]
\centering
\caption{Combined PMO Benchmark Results. GenMol, f-RAG, molGA and REINVENT taken from literature \cite{lee2025fragfm} (there reported without standard deviations). Mean and standard deviation of AUC top-10 are reported over 5 random seeds. The model variants are according to the ablation study in Table \ref{tab:auc_table_suppl}.}
\label{tab:pmo_results}
\resizebox{0.9\textheight}{!}{%
\begin{tabular}{lcccccccccc}
\toprule
\textbf{Oracle} &
\textbf{GenMol} &
\textbf{f-RAG} &
\textbf{molGA} &
\textbf{REINVENT} &
\textbf{\shortstack{Original Genetic GFN \\ {\small (SMILES \& ZINC)}}} &
\textbf{\shortstack{$+$ Switch to \\ {\small (Random Dataset)}}} &
\textbf{\shortstack{$+$ Switch to  \\ {GGS encoding}}}    &
\textbf{\shortstack{$+$ Transformer \\ {architecture}}} &
\textbf{\shortstack{$+$ DCD, DEX \\ {$-$KL}}} &
\textbf{$+$ Descriptors } \\
\midrule
albuterol\_similarity & 0.94 & 0.98 & 0.90 & 0.88 & $0.94 \pm 0.02$ & $0.92 \pm 0.02$ & $0.89 \pm 0.02$ & $0.84 \pm 0.10$ & $0.87 \pm 0.11$ & $0.90 \pm 0.01$ \\
amlodipine\_mpo & 0.81 & 0.75 & 0.69 & 0.64 &  $0.67 \pm 0.03$ & $0.68 \pm 0.05$ & $0.60 \pm 0.02$ & $0.58 \pm 0.04$ & $0.58 \pm 0.01$ & $0.56 \pm 0.02$ \\
celecoxib\_rediscovery & 0.83 & 0.78 & 0.57 & 0.71 &  $0.77 \pm 0.11$ & $0.58 \pm 0.04$ & $0.59 \pm 0.08$ & $0.60 \pm 0.07$ & $0.71 \pm 0.11$ & $0.68 \pm 0.06$ \\
deco\_hop & 0.96 & 0.94 & 0.65 & 0.67 &  $0.68 \pm 0.08$ & $0.63 \pm 0.01$ & $0.78 \pm 0.08$ & $0.65 \pm 0.01$ & $0.70 \pm 0.10$ & $0.65 \pm 0.01$ \\
drd2 & 1.0 & 0.99 & 0.94 & 0.95 &  $0.97 \pm 0.01$ & $0.97 \pm 0.01$ & $0.96 \pm 0.02$ & $0.96 \pm 0.01$ & $0.96 \pm 0.01$ & $0.96 \pm 0.00$ \\
fexofenadine\_mpo & 0.90 & 0.86 & 0.83 & 0.78 &  $0.83 \pm 0.05$ & $0.80 \pm 0.01$ & $0.77 \pm 0.01$ & $0.78 \pm 0.02$ & $0.79 \pm 0.02$ & $0.79 \pm 0.01$ \\
gsk3b & 0.99 & 0.97 & 0.84 & 0.87 & $0.88 \pm 0.04$ & $0.79 \pm 0.04$ & $0.86 \pm 0.08$ & $0.83 \pm 0.05$ & $0.83 \pm 0.08$ & $0.80 \pm 0.04$ \\
isomers\_c7h8n2o2 & 0.94 & 0.96 & 0.88 & 0.85 &  $0.97 \pm 0.01$ & $0.98 \pm 0.00$ & $0.98 \pm 0.00$ & $0.98 \pm 0.01$ & $0.97 \pm 0.01$ & $0.98 \pm 0.01$ \\
isomers\_c9h10n2o2pf2cl & 0.83 & 0.85 & 0.87 & 0.64 &  $0.89 \pm 0.02$ & $0.89 \pm 0.05$ & $0.81 \pm 0.03$ & $0.82 \pm 0.02$ & $0.78 \pm 0.06$ & $0.83 \pm 0.04$ \\
jnk3 & 0.91 & 0.90 & 0.70 & 0.78 &  $0.71 \pm 0.16$ & $0.57 \pm 0.10$ & $0.66 \pm 0.03$ & $0.65 \pm 0.05$ & $0.66 \pm 0.04$ & $0.62 \pm 0.04$ \\
median1 & 0.40 & 0.34 & 0.26 & 0.36 &  $0.35 \pm 0.01$ & $0.34 \pm 0.02$ & $0.32 \pm 0.03$ & $0.30 \pm 0.03$ & $0.28 \pm 0.01$ & $0.28 \pm 0.04$ \\
median2 & 0.40 & 0.32 & 0.30 & 0.28 &  $0.27 \pm 0.01$ & $0.27 \pm 0.02$ & $0.33 \pm 0.01$ & $0.33 \pm 0.01$ & $0.32 \pm 0.03$ & $0.32 \pm 0.02$ \\
mestranol\_similarity & 0.98 & 0.67 & 0.60 & 0.62 &  $0.76 \pm 0.08$ & $0.75 \pm 0.17$ & $0.85 \pm 0.02$ & $0.85 \pm 0.02$ & $0.86 \pm 0.02$ & $0.84 \pm 0.05$ \\
osimertinib\_mpo & 0.88 & 0.87 & 0.84 & 0.84 &  $0.85 \pm 0.01$ & $0.84 \pm 0.01$ & $0.87 \pm 0.01$ & $0.86 \pm 0.01$ & $0.86 \pm 0.01$ & $0.85 \pm 0.01$ \\
perindopril\_mpo & 0.72 & 0.68 & 0.55 & 0.54 &  $0.59 \pm 0.02$ & $0.57 \pm 0.03$ & $0.58 \pm 0.04$ & $0.56 \pm 0.03$ & $0.55 \pm 0.01$ & $0.56 \pm 0.04$ \\
qed & 0.94 & 0.94 & 0.94 & 0.94 &  $0.95 \pm 0.00$ & $0.94 \pm 0.00$ & $0.94 \pm 0.00$ & $0.94 \pm 0.00$ & $0.94 \pm 0.00$ & $0.94 \pm 0.00$ \\
ranolazine\_mpo & 0.82 & 0.82 & 0.80 & 0.76 &  $0.78 \pm 0.02$ & $0.77 \pm 0.02$ & $0.77 \pm 0.01$ & $0.79 \pm 0.01$ & $0.79 \pm 0.01$ & $0.79 \pm 0.01$ \\
scaffold\_hop & 0.63 & 0.58 & 0.53 & 0.56 &  $0.56 \pm 0.03$ & $0.60 \pm 0.15$ & $0.58 \pm 0.02$ & $0.59 \pm 0.11$ & $0.59 \pm 0.11$ & $0.56 \pm 0.03$ \\
sitagliptin\_mpo & 0.59 & 0.60 & 0.58 & 0.02 &  $0.64 \pm 0.05$ & $0.42 \pm 0.17$ & $0.48 \pm 0.09$ & $0.46 \pm 0.08$ & $0.45 \pm 0.08$ & $0.50 \pm 0.11$ \\
thiothixene\_rediscovery & 0.70 & 0.58 & 0.52 & 0.53 &  $0.58 \pm 0.04$ & $0.65 \pm 0.11$ & $0.45 \pm 0.04$ & $0.44 \pm 0.03$ & $0.44 \pm 0.04$ & $0.40 \pm 0.02$ \\
troglitazone\_rediscovery & 0.87 & 0.45 & 0.43 & 0.44 &  $0.51 \pm 0.04$ & $0.53 \pm 0.07$ & $0.42 \pm 0.02$ & $0.39 \pm 0.05$ & $0.39 \pm 0.04$ & $0.41 \pm 0.04$ \\
valsartan\_smarts & 0.82 & 0.63 & 0.00 & 0.18 &  $0.12 \pm 0.23$ & $0.00 \pm 0.00$ & $0.10 \pm 0.23$ & $0.03 \pm 0.08$ & $0.05 \pm 0.11$ & $0.00 \pm 0.00$ \\
zaleplon\_mpo & 0.58 & 0.49 & 0.52 & 0.36 &  $0.54 \pm 0.03$ & $0.49 \pm 0.03$ & $0.53 \pm 0.03$ & $0.51 \pm 0.02$ & $0.50 \pm 0.02$ & $0.49 \pm 0.01$ \\
\midrule
\textbf{Total} & 18.36 & 16.93 & 14.71 & 14.20 &  $\mathbf{15.81 \pm 0.34}$ & $\mathbf{14.97 \pm 0.34}$ & $\mathbf{15.11 \pm 0.30}$ & $\mathbf{14.75 \pm 0.23}$ & $\mathbf{14.90 \pm 0.28}$ & $\mathbf{14.70 \pm 0.18}$ \\
\bottomrule

\end{tabular}
}
\end{sidewaystable}


\end{document}